\documentclass[10pt,twocolumn,letterpaper]{article}

\usepackage{iccv}
\usepackage{times}
\usepackage{epsfig}
\usepackage{graphicx}
\usepackage{amsmath}
\usepackage{amssymb}
\usepackage{subfig}
\usepackage{booktabs}
\usepackage{float}
\usepackage{caption}
\usepackage{balance}

\usepackage[pagebackref=true,breaklinks=true,letterpaper=true,colorlinks,bookmarks=false,citecolor=blue,linkcolor=blue]{hyperref}
\iccvfinalcopy 

\def\iccvPaperID{4475} 

\ificcvfinal\fi
\begin{document}

\title{Fast Universal Style Transfer for Artistic and Photorealistic Rendering
}

\author{Jie An\thanks{Equal contribution.}\ \thanks{This work was done when Jie An worked as an Intern at Baidu Inc..}\\
School of Mathematical Sciences\\
Peking University\\
{\tt\small jie.an@pku.edu.cn}
\and
Haoyi Xiong\footnotemark[1]\\
Big Data Lab\\
Baidu Research\\
{\tt\small xionghaoyi@baidu.com}
\and
Jiebo Luo\\
Department of Computer Science\\
University of Rochester \\
{\tt\small jluo@cs.rochester.edu}
\and
Jun Huan\thanks{Corresponding author.}\\
Big Data Lab\\
Baidu Research\\
{\tt\small huanjun@baidu.com}
\and
Jinwen Ma\footnotemark[3]\\
School of Mathematical Sciences \\
Peking University \\
{\tt\small jwma@math.pku.edu.cn}
}

\maketitle

\begin{figure*}[t]    
\label{fig:maincomp}
\vspace{-5mm}
 \centering
    \subfloat[Content~/~Reference images (Photorealistic)]{\includegraphics[width=0.31\textwidth]{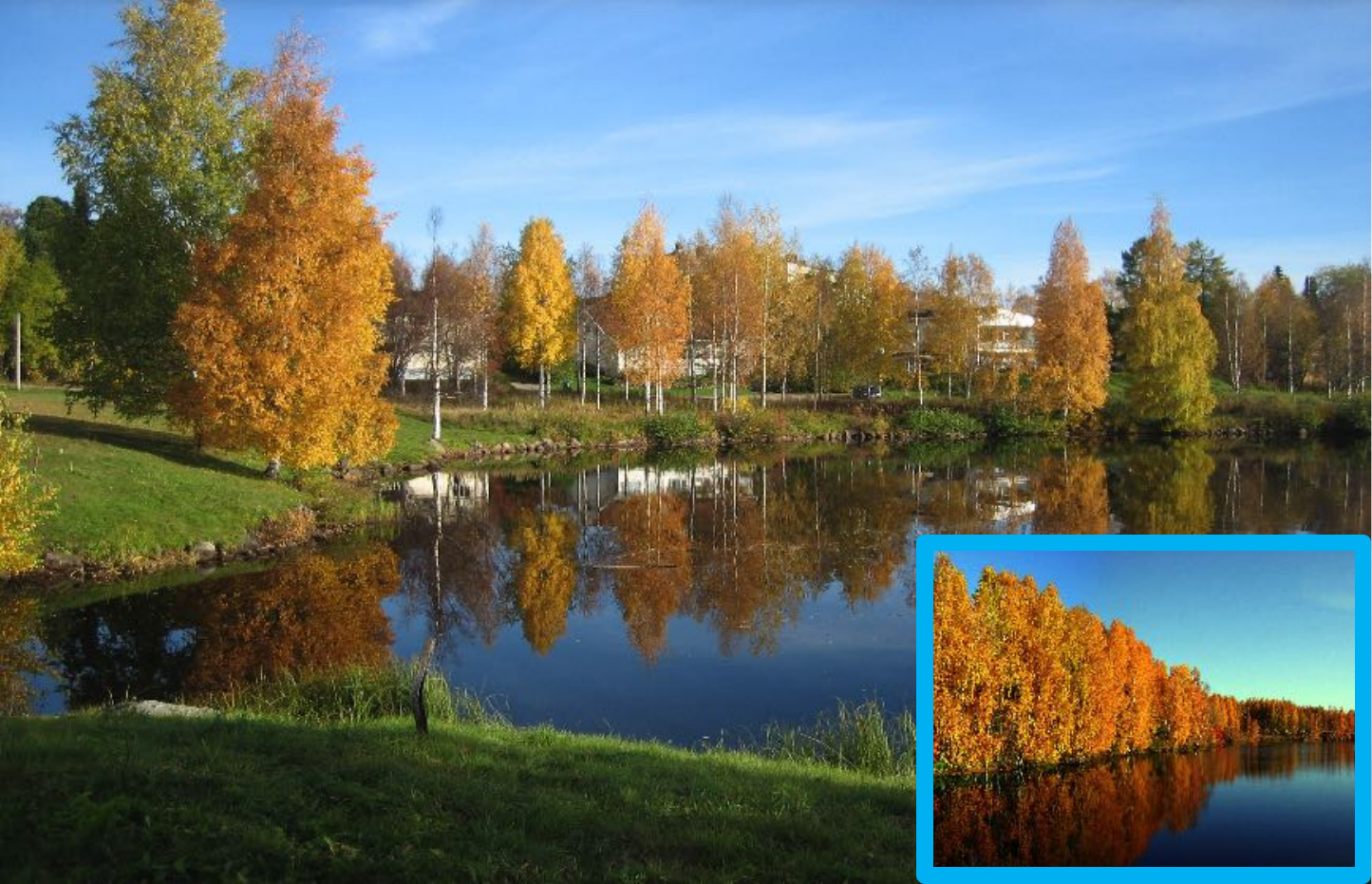}}\hspace{1mm}
    \subfloat[PhotoWCT~\cite{li2018closed}~/~ Computing time: 53.05s]{\includegraphics[width=0.31\textwidth]{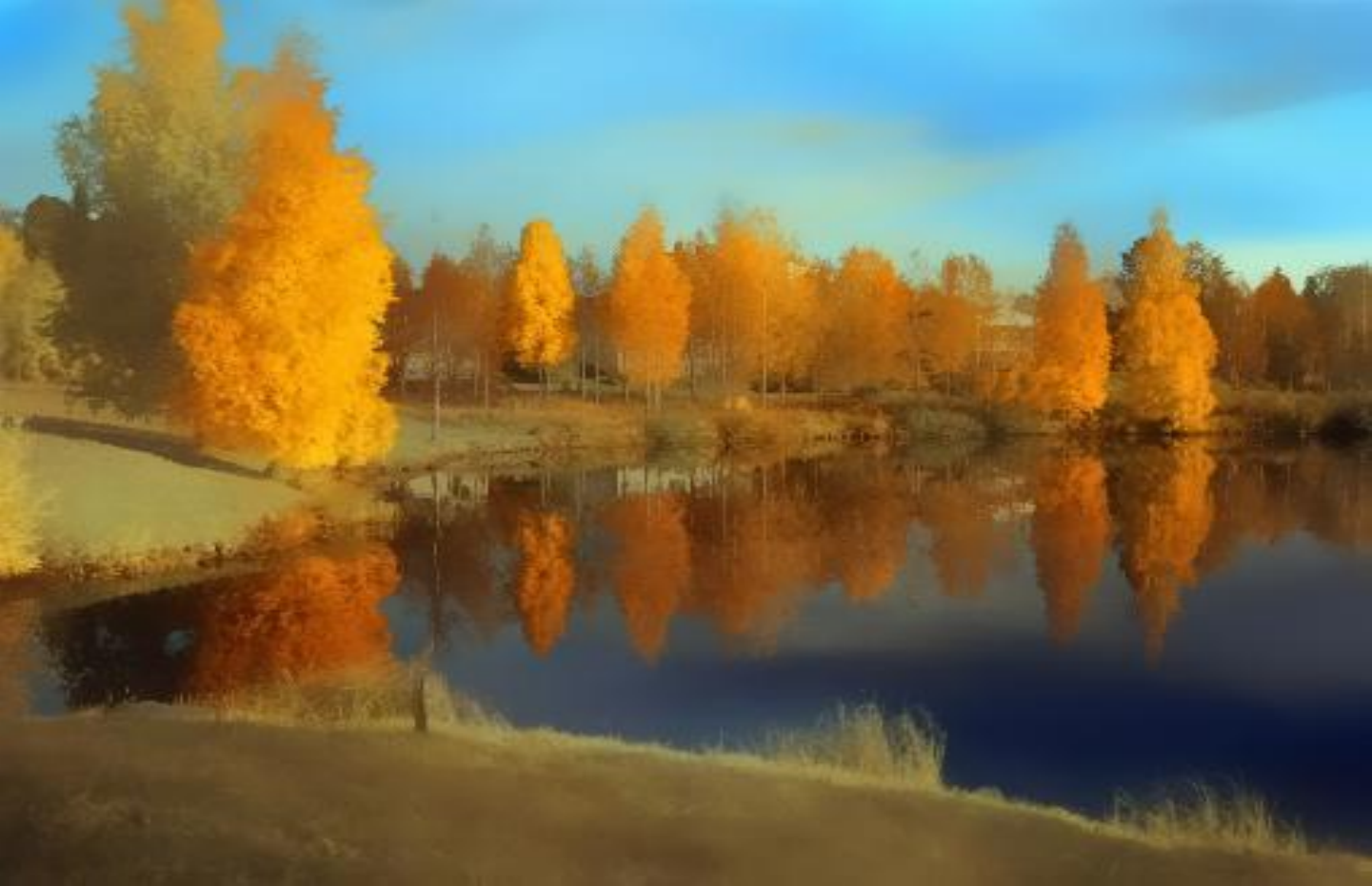}}\hspace{1mm}
    \subfloat[\textbf{PhotoNet(WCT)~/~Computing time: 0.95s}]{\includegraphics[width=0.31\textwidth]{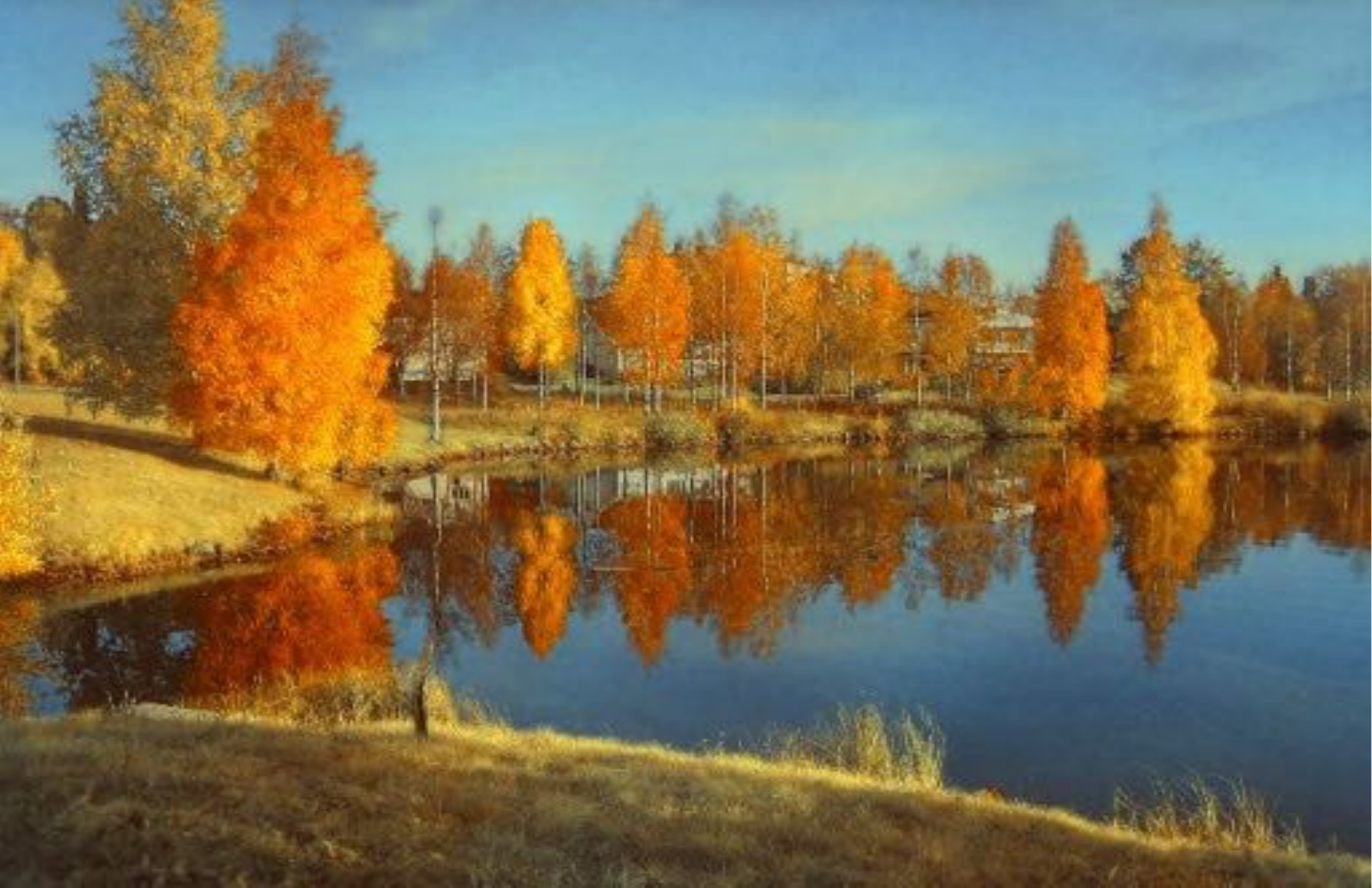}}\\
    \subfloat[Content~/~Reference images (Artistic)]{\includegraphics[width=0.31\textwidth]{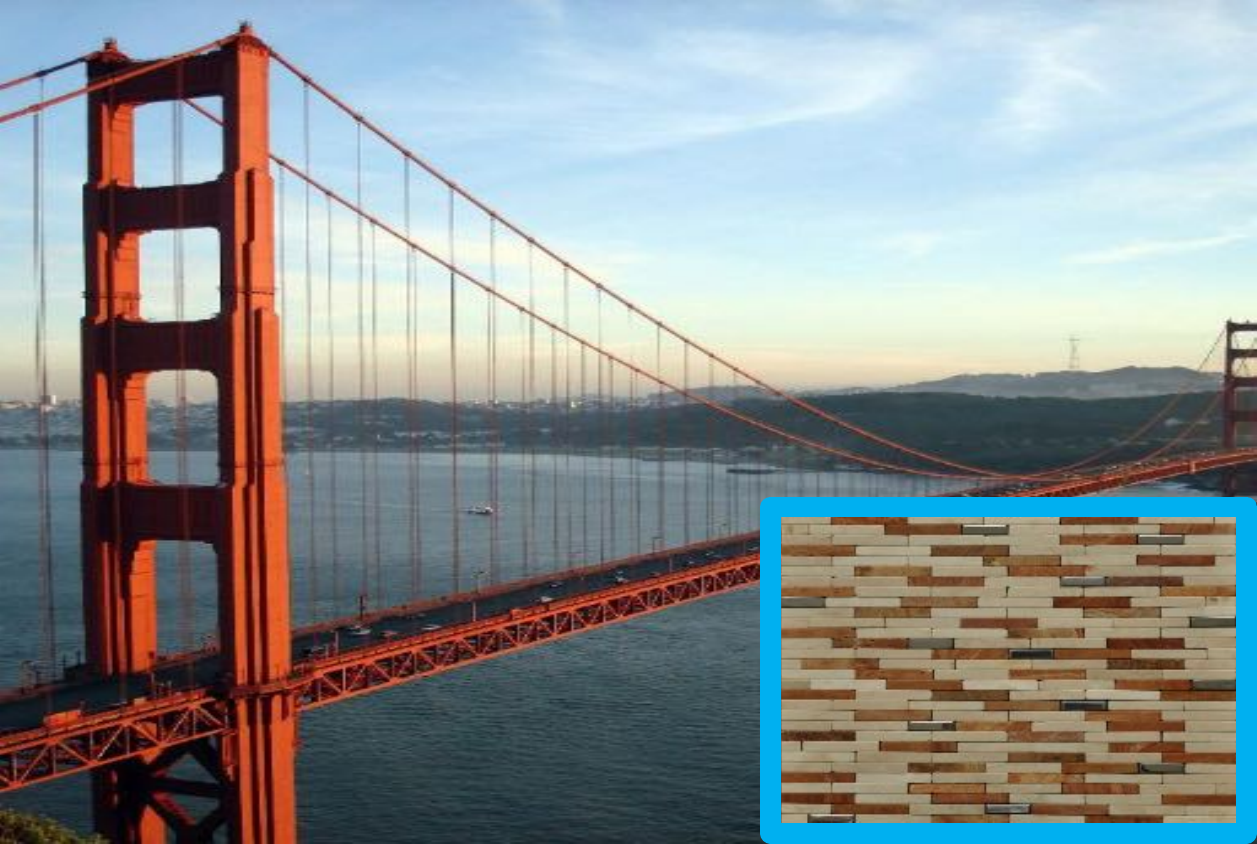}}\hspace{1mm}
    \subfloat[WCT~\cite{li2017universal}~/~Computing time: 1.20s]{\includegraphics[width=0.31\textwidth]{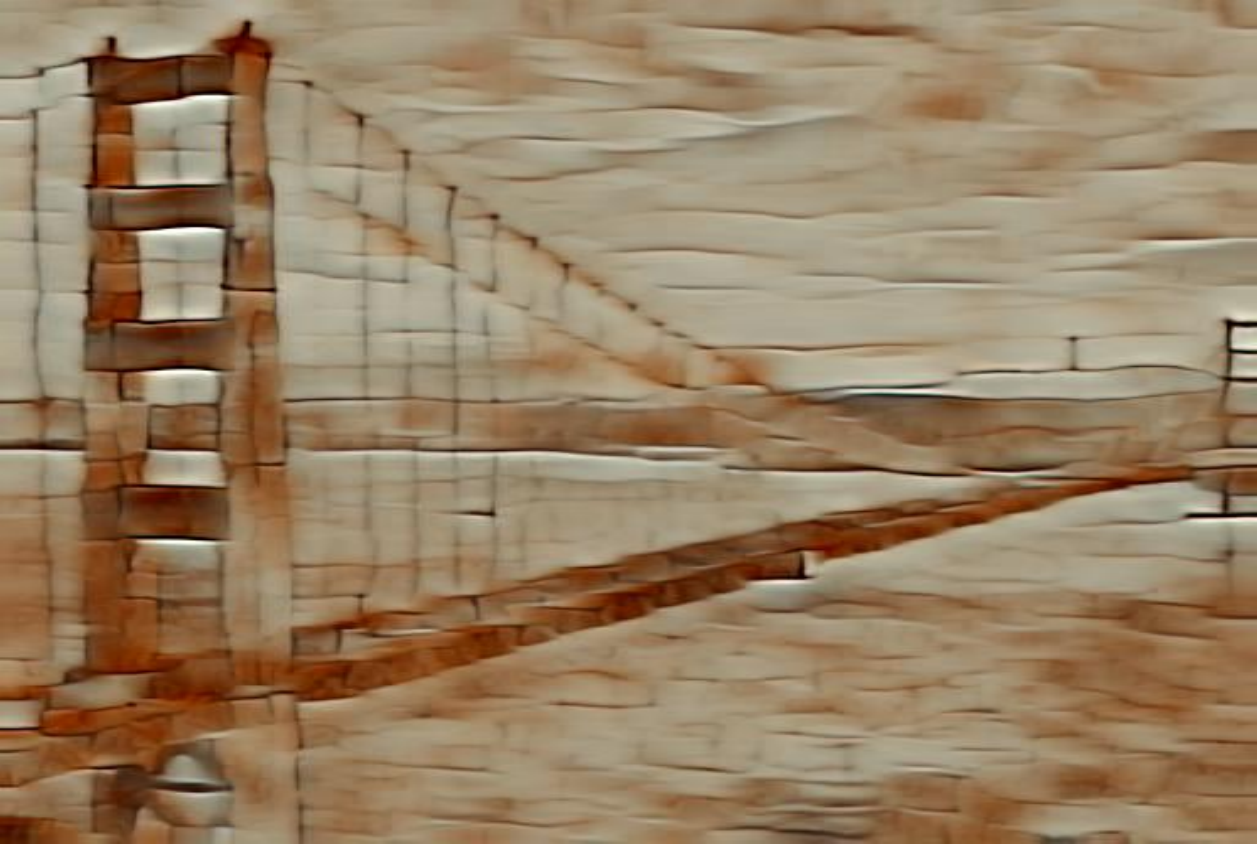}}\hspace{1mm}
    \subfloat[\textbf{ArtNet(WCT)~/~Computing time: 0.45s}]{\includegraphics[width=0.31\textwidth]{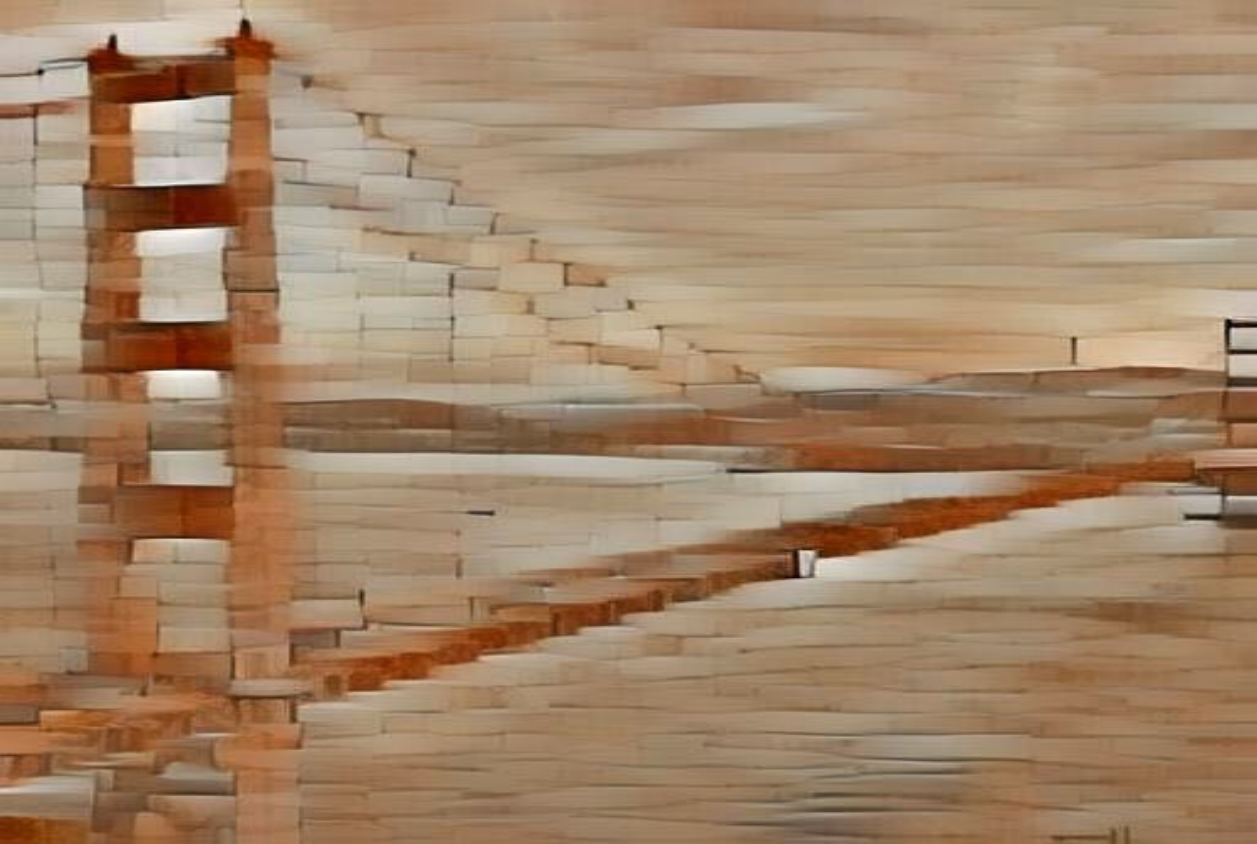}}
    \vspace{-3mm}
    \caption{\textbf{Visual comparison of  photorealistic and artistic style transfer}. Content images are (a) and (d); Reference style images are shown in the bottom-right corners of (a) and (d). For photorealistic stylization, PhotoWCT~\cite{li2018closed} consumes significant computing time while producing an overly smooth image shown in (b). Our proposed PhotoNet generates the image shown in (c) of rich details with only 1/50th of the computational time. 
    Similarly, our proposed ArtNet produces the result in (f) with reduced artifacts and distortion in comparison with WCT~\cite{li2017universal} in (e) for artistic style transfer.}
    \vspace{-3mm}
\end{figure*}

\begin{abstract}
    Universal style transfer is an image editing task that renders an input content image using the visual style of arbitrary reference images, including both artistic and photorealistic stylization.  
    Given a pair of images as the source of content and the reference of style, existing solutions usually first train an auto-encoder (AE) to reconstruct the image using deep features and then embeds pre-defined style transfer modules into the AE reconstruction procedure to transfer the style of the reconstructed image through modifying the deep features.
    %
    %
    While existing methods typically need multiple rounds of time-consuming AE reconstruction for better stylization, our work intends to design novel neural network architectures on top of AE for fast style transfer with fewer artifacts and distortions \emph{all in one pass of end-to-end inference}.
   To this end, we propose two network architectures named \textbf{ArtNet} and \textbf{PhotoNet} to improve artistic and photo-realistic stylization,  respectively.
   Extensive experiments demonstrate that ArtNet generates images with fewer artifacts and distortions against the state-of-the-art artistic transfer algorithms, while PhotoNet improves the photorealistic stylization results by creating sharp images faithfully preserving rich details of the input content.
    Moreover, ArtNet and PhotoNet can achieve  3$\times$ to 100$\times$ speed-up over the state-of-the-art algorithms, which is a major advantage for large content images. 
\end{abstract}

\section{Introduction}
Universal style transfer is an image editing task that renders an input content image using the visual styles of arbitrary reference images. 
Among a wide range of stylization tasks, two common tasks of style transfer are \emph{artistic style transfer} and \emph{photorealistic stylization}.
More specifically, given a pair of images for content and style reference,  respectively,  \emph{artistic style transfer} aims to generate an artistic rendition of the given content using the textures, colors, and shapes of the style reference, while  \emph{photorealistic stylization} creates ``photographs'' of the given content as if they were captured in the same settings of the style references. 

%

Many research efforts~\cite{gatys2015neural, Gatys2016, johnson2016perceptual, ulyanov2016texture, li2017universal, huang2017arbitrary} have been dedicated to universal style transfer. A pioneer work is presented in \cite{gatys2015neural, Gatys2016} where Gatys.~\etal~ make the first attempt to connect the style representation to the Gram matrices of deep features. Their work shows that Gram matrices of deep features have an exceptional ability to encode the styles of images. Following this line of research, several recent works aiming to minimize a Gram matrices based loss function~\cite{gatys2015neural, Gatys2016, risser2017stable, luan2017deep} have been proposed. While these algorithms can produce high-quality stylization results, they all suffer from a high computational cost even with acceleration techniques  ~\cite{johnson2016perceptual, ulyanov2016texture, ulyanov2017improved, dumoulin2017learned, li2017diversified}. Moreover, all these algorithms usually can work well only on a limited number of image styles. 

%

\begin{figure*}[tb]
    \centering
    \hspace{10mm}
     \subfloat[The AE used by WCT~\cite{li2017universal} and AdaIN~\cite{huang2017arbitrary}.]{\includegraphics[width=0.3\linewidth]{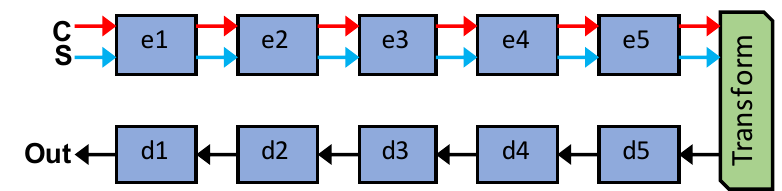}}\hspace{22mm}
     \subfloat[Multi-level stylization with multi-round computation used by WCT~\cite{li2017universal}.]{\includegraphics[width=0.50\linewidth]{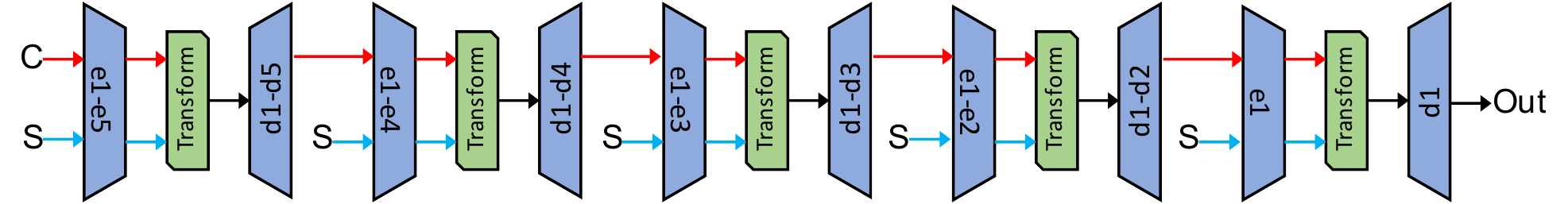}}\\
    \subfloat[ArtNet architecture.]{\includegraphics[width=0.43\linewidth]{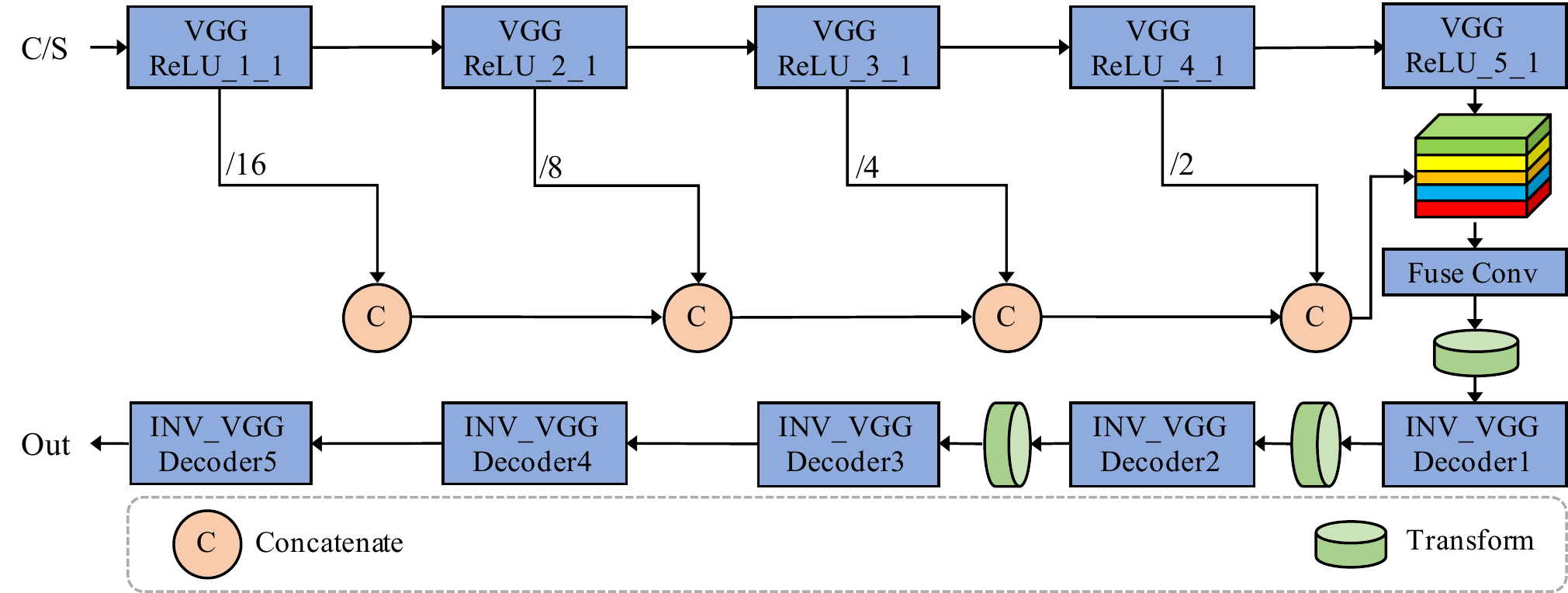}}\hspace{7mm}
    \subfloat[PhotoNet architecture.]{\includegraphics[width=0.52\linewidth]{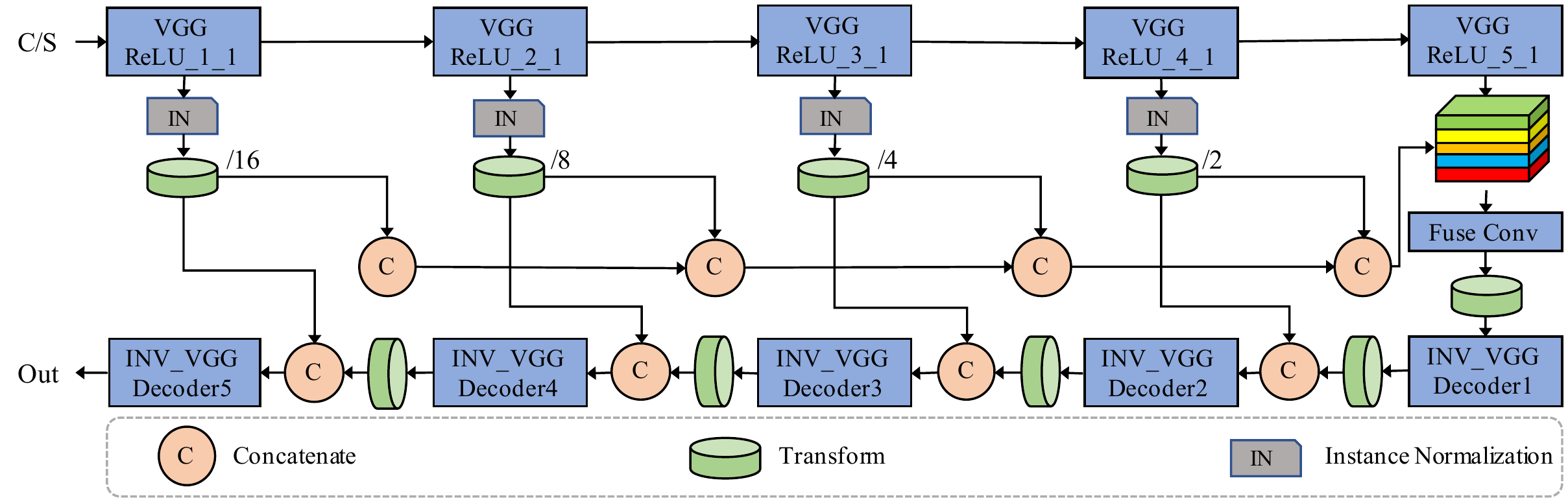}}
    \vspace{-3mm}
     \caption{\textbf{Comparison of architectures.} The multi-level stylization scheme first trains an auto-encoder (AE) shown in (a) with an image reconstruction loss, then runs multi-round of AE with the  WCT~\cite{li2017universal} transform for style transfer (shown in (b)). Our proposed ArtNet (c) introduces deep feature aggregation and multi-stage stylization on the decoder to better stylize images, while PhotoNet (d) utilizes additional normalized skip connections to preserve more details in style transfer. Please note that networks shown in (c) and (d) are inference-time architectures of ArtNet and PhotoNet, respectively. For training, the transform operations (green cylinders) are excluded while the networks are trained as common AEs using the reconstruction loss.}
    \label{fig:architecture}
    \vspace{-3mm}
\end{figure*}

Recently, universal style transfer methods~\cite{chen2016fast, li2017universal, huang2017arbitrary, li2018closed, sheng2018avatar, gu2018arbitrary} have been proposed for image transfer with respect to \emph{arbitrary} styles and contents. 
For example, multi-level stylization~\cite{li2017universal, li2018closed} or iterative EM process~\cite{gu2018arbitrary} have been proposed recently. Multi-level stylization algorithms first extract multi-level features using a multi-layer auto-encoder (AE) in depth, where each layer of AE refers to a level of features,  then render styles using all the features from high-level to low-level iteratively with style transfer modules. In Figs.~\ref{fig:architecture} (a) and (b), we show the architectures of auto-encoder and multi-level stylization  used in popular algorithms. In summary, multi-level stylization processes the images over trained AEs and transfer modules multiple times~\cite{li2017universal, li2018closed} by utilizing features from high to low levels to improve universal style transfer results. 

The improved quality does come with a significant drawback: multi-level stylization has a significantly large computation cost. There is an option to ``turn off'' the multi-level settings, at the expense of imperfect  textures~\cite{chen2016fast}, artifacts~\cite{huang2017arbitrary, sheng2018avatar}, and distortions~\cite{li2017universal}. In addition to multi-round stylization, post-processing for photo-realistic stylization~\cite{li2018closed} is yet another performance bottle-neck in style transfer. 
It is desirable to design novel approaches to utilize multi-level features of images for better image quality (\ie, fewer artifacts, less distortion, and higher  sharpness) while reducing the computational cost for both artistic and photorealistic rendering. 

%
%
%


%

In this work, we propose a novel neural network architecture on top of common multi-layer AEs for fast universal style transfer using multi-level features with pre-defined style transfer modules (e.g., AdaIN~\cite{huang2017arbitrary}, WCT~\cite{li2017universal}, and PhotoWCT~\cite{li2018closed}), while avoiding the use of multi-round stylization and post-processing.
%
%
%
%
%
%
%
We also design novel auto-encoder architectures with superior reconstruction capacity that can 
%
alleviate the artifacts and distortions with better stylization performance. With such AE and pre-defined transfer modules, we can obtain high-quality style transferred images using multi-level features but with \emph{single-round} computation for much reduced computation.

Specifically for artistic stylization tasks,  we first introduce a feed-forward network named \textbf{ArtNet} based on feature aggregation~\cite{yu2018deep} to  achieve visually pleasing artistic stylization results by eliminating the artifacts and distortions of images.
With a pre-trained encoder, the proposed method first extracts and aggregates features ranging from low-level and high-level of encoder parts in AE to better preserve the details of the images (see also in Fig.~\ref{fig:architecture}(c)). Next, to obtain better stylization, ArtNet replaces the time-consuming multi-round stylization with a single-round multi-stage decoder with transfer modules embedded, where a \emph{``sandwich structure''} that segments every two stages of the decoder with a transfer module is adopted. Please refer to Figs.~\ref{fig:architecture} (a), (b) and (c) for the detailed comparisons of the two architectures.

For photorealistic image transfer tasks, we propose \textbf{PhotoNet} that further extends ArtNet with more sophisticated connections to produce remarkable quality improvement for photorealistic stylization while avoiding the use of any  post-processing.
%
%
%
In particular, PhotoNet incorporates the additional skip connections coupled with  instance normalization~\cite{ulyanov1607instance}, passing the low-level features (which may be diminished after multi-layer decoding) directly to the decoder stages respectively so as to achieve exact image reconstruction. In this way with pre-defined transfer modules, PhotoNet makes the transferred images sharper with more details from the content image preserved. Figs.~\ref{fig:architecture} (a), (b) and (d) offered more details. 

Our extensive experiments based on subjective and objective metrics show that ArtNet outperforms the artistic stylization results of AdaIN~\cite{huang2017arbitrary} and WCT~\cite{li2017universal} when using these two as transfer modules.
For photo-realistic stylization, PhotoNet with WCT~\cite{li2018closed} as transfer modules has a unique capability of rendering semantically consistent, sharp images while avoiding the artifacts or too much smoothness.
In addition to superior performance against the state-of-the-art methods, PhotoNet achieves more than 50$\times$ speed-up in terms of computational time. 

Our main contributions are summarized as follows:
\begin{itemize}
\vspace{-0.1in}
    \item[-] We propose ArtNet for artistic style transfer based on deep feature aggregation and multi-stage stylization on the decoder. ArtNet remarkably outperforms  AdaIN~\cite{huang2017arbitrary} with significantly fewer artifacts. ArtNet with the WCT module achieves visually more pleasing results in less distortion while cutting down the time-consumption of the WCT~\cite{li2017universal} to a third.
    \vspace{-0.1in}
   \item[-] We further propose PhotoNet with additional skip connections coupled with instance normalization. PhotoNet produces rich-detailed, locally-consistent, accurately-stylized photorealistic images with 1/50th of the time-consumption against the state-of-the-art algorithms~\cite{luan2017deep, li2018closed}.
\end{itemize}

\section{Related Work}
We first review the most relevant work to our study and discuss the contribution made by our work. 

\textbf{Artistic style transfer.~} 
%
Prior to the adoption of deep neural networks, several classical models based on stroke rendering~\cite{hertzmann1998painterly}, image analogy~\cite{hertzmann2001image,shih2013data,shih2014style,frigo2016split,liao2017visual}, or image filtering~\cite{winnemoller2006real} have been proposed to make a trade-off between quality, generalization, and efficiency for artistic style transfer.
%
%
%
In addition to the work already mentioned in the introduction, numerous Gram based algorithms have been recently developed inspired by Gatys~\etal~\cite{gatys2015neural, Gatys2016}. Such methods can be classified into one style per model~\cite{li2016combining, ulyanov1607instance, johnson2016perceptual, ulyanov2016texture, ulyanov2017improved, wang2017multimodal, risser2017stable, li2017laplacian}, multi-style per model~\cite{dumoulin2017learned, chen2017stylebank}, and universal stylization methods~\cite{chen2016fast, huang2017arbitrary, li2017universal, gu2018arbitrary} with respect to the generalization ability.
%
%

\textbf{Photo-realistic style transfer.~} 
%
In terms of methodologies, existing photo-realistic stylization methods~\cite{luan2017deep, li2018closed} either introduce smoothness-based loss term~\cite{luan2017deep} or utilize post-processing to smooth the transferred images~\cite{li2018closed}, which inevitably decreases the sharpness of images and increases the time-consumption significantly.
%
%
%
%
%
In addition to style transfer, photo-realistic stylization has also been studied in image-to-image translation~\cite{isola2017image, wang2018high, liu2016coupled, taigman2016unsupervised, shrivastava2017learning, liu2017unsupervised, zhu2017unpaired, huang2018multimodal}.
The major difference between photo-realistic style transfer and image-to-image translation is that photo-realistic style transfer does not require paired training data (i.e., pre-transfer and post-transfer images). Of course, image-to-image translation can solve even more complicated task such as the  man-to-woman and cat-to-dog adaption problems. 

\textbf{Discussion.}
The work most relevant to our study includes WCT~\cite{li2017universal}, AdaIN~\cite{huang2017arbitrary} and PhotoWCT~\cite{li2018closed}, while the first two have been used for artistic stylization and the last one is for photo-realistic stylization. Specifically, ArtNet consists of a new network architecture that can incorporate the transfer modules of WCT~\cite{li2017universal} and AdaIN~\cite{huang2017arbitrary}. The proposed method can avoid multi-round stylization while ensuring the effectiveness of style transfer. PhotoNet also incorporates the transfer module of PhotoWCT~\cite{li2018closed} in our newly-proposed architecture. It does not require  time-consuming multi-round stylization and post-processing. The images produced by PhotoNet have considerably higher sharpness, reduced distortion and significant reduction of computational cost.

%
%
%

\begin{figure*}[tb]
    \centering
    \subfloat[Content]{\includegraphics[width=0.16\linewidth]{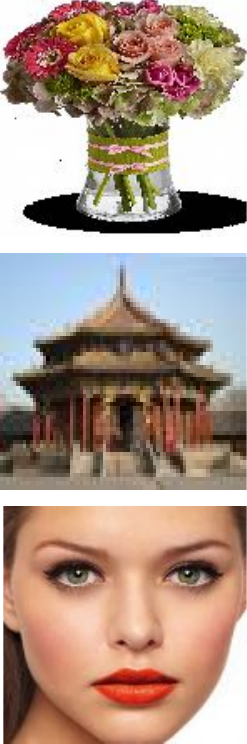}}\hspace{1mm}
    \subfloat[Style]{\includegraphics[width=0.16\linewidth]{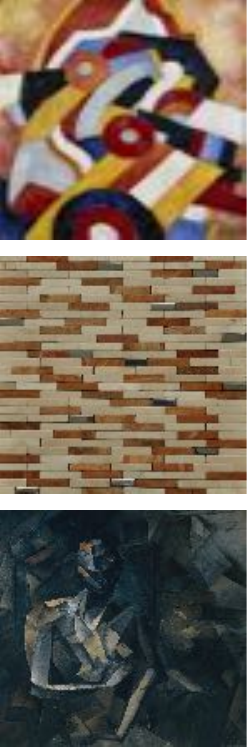}\hspace{1mm}}
    \subfloat[AdaIN~\cite{huang2017arbitrary}.]{\includegraphics[width=0.16\linewidth]{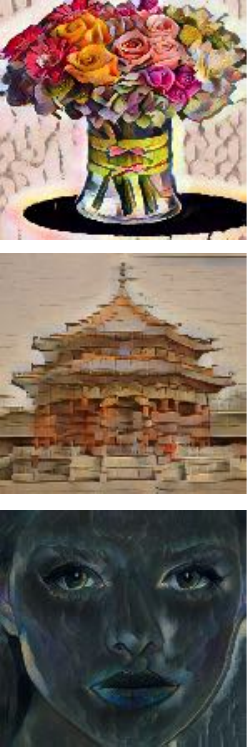}\hspace{1mm}}
    \subfloat[ArtNet(AdaIN)]{\includegraphics[width=0.16\linewidth]{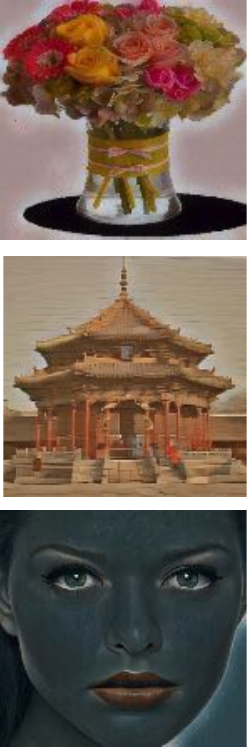}\hspace{1mm}}
    \subfloat[WCT~\cite{li2017universal}.]{\includegraphics[width=0.16\linewidth]{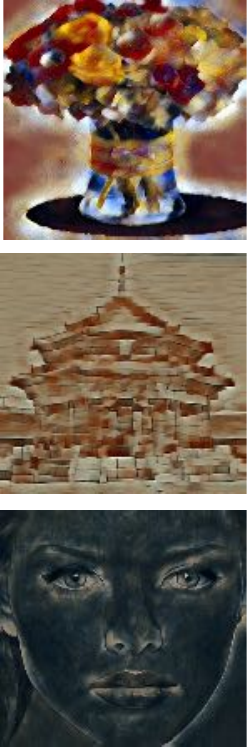}\hspace{1mm}}
    \subfloat[ArtNet(WCT)]{\includegraphics[width=0.16\linewidth]{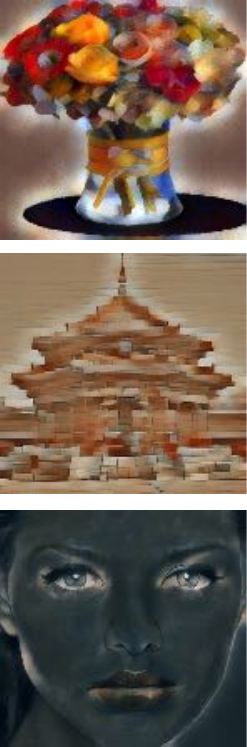}}
       \vspace{-3mm}
 \caption{\textbf{Results of the contrast experiments against baseline artistisc style transfer methods.}}
     \vspace{-5mm}
   \label{fig:comp_artistic}
\end{figure*}

\begin{figure*}[tb]
    \centering
    \subfloat[Content]{\includegraphics[width=0.19\linewidth]{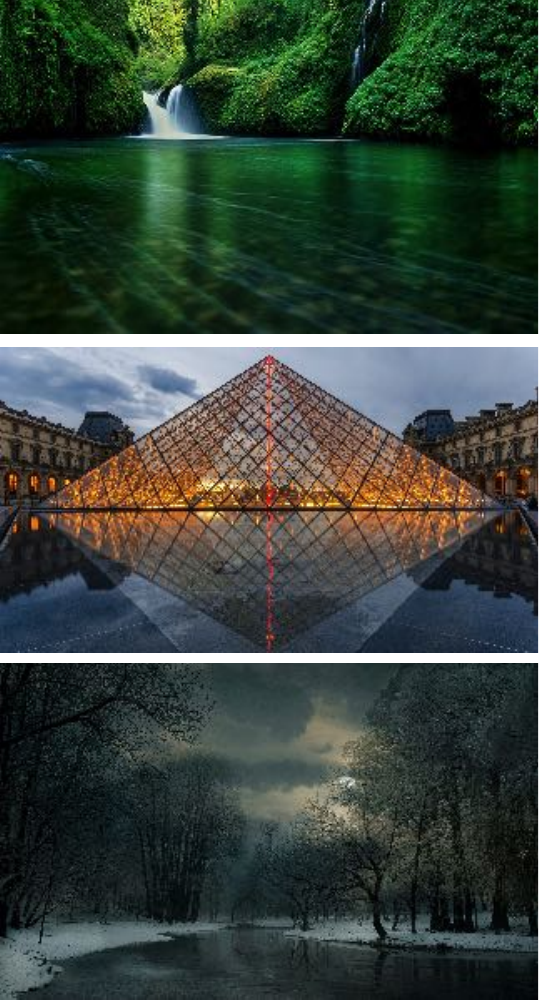}}\hspace{1mm}
    \subfloat[Style]{\includegraphics[width=0.19\linewidth]{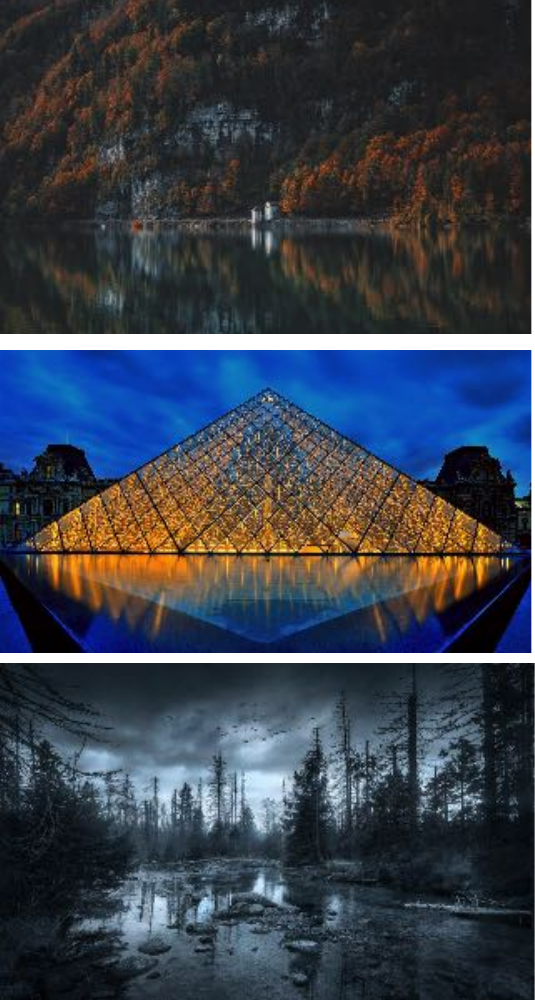}}\hspace{1mm}
    \subfloat[PhotoWCT~\cite{li2018closed}.]{\includegraphics[width=0.19\linewidth]{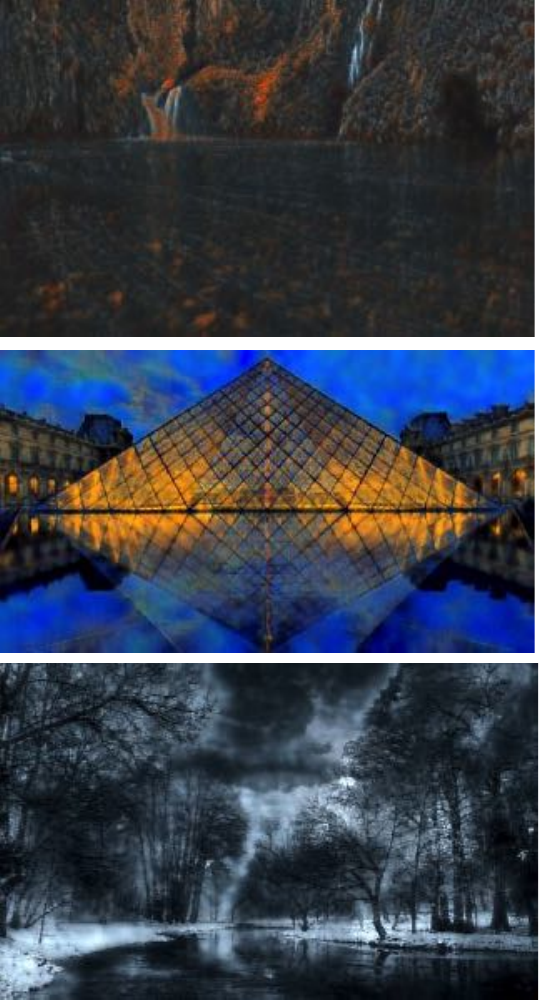}}\hspace{1mm}
    \subfloat[PhotoWCT+Smooth~\cite{li2018closed}.]{\includegraphics[width=0.19\linewidth]{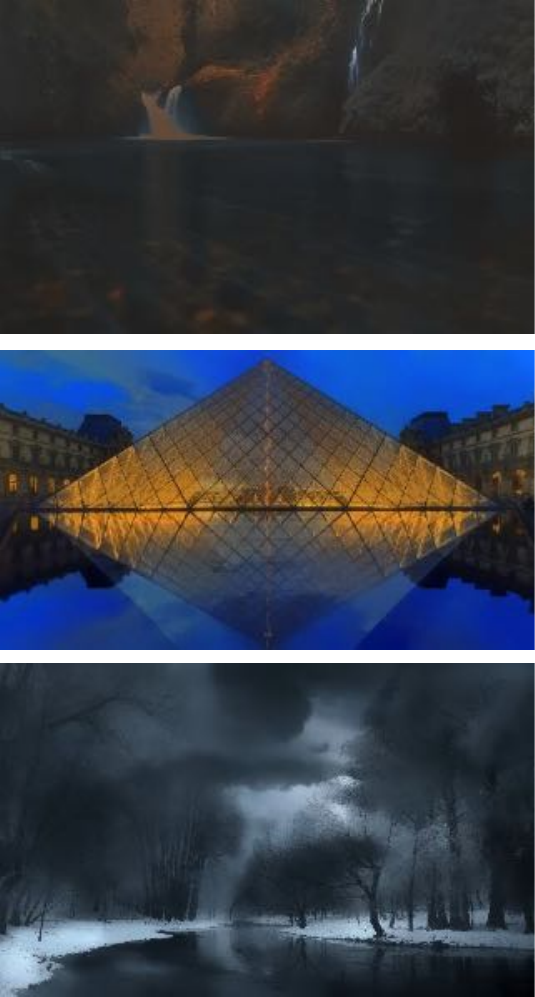}}\hspace{1mm}
    \subfloat[PhotoNet(WCT)]{\includegraphics[width=0.19\linewidth]{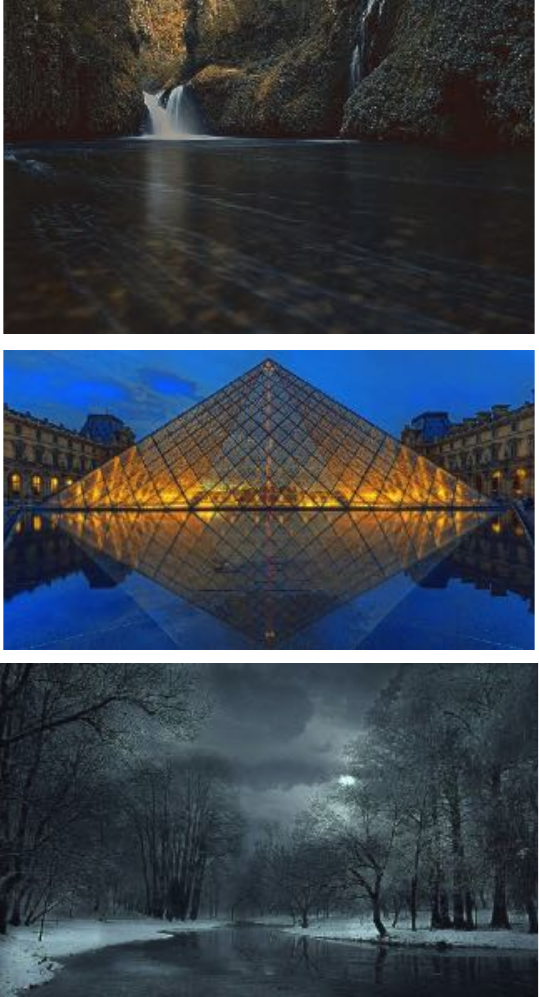}}
        \vspace{-3mm}
\caption{\textbf{Results of the contrast experiments against baseline photorealistc style transfer method.}}
    \vspace{-3mm}
    \label{fig:comp_photorealistic}
\end{figure*}
\begin{table*}[t]
        \vspace{-3mm}
\caption{\textbf{Quantitative evaluation results for stylization methods.} The FID and TV scores are only applicable to the photo-realistic methods. A lower FID score means the evaluated method creates an image with a more similar style to the reference style image. A higher total variation score indicates that the measured image has more details.}
        \vspace{-3mm}
\centering
    \small
    \begin{tabular}{lcc|c}
        \toprule
        Method & AdaIN~\cite{huang2017arbitrary} \vs \textbf{ArtNet(AdaIN)} & WCT~\cite{li2017universal} \vs \textbf{ArtNet(WCT)} & PhotoWCT~\cite{li2018closed} \vs \textbf{PhotoNet(WCT)} \\
        \midrule
        Preference~$\uparrow$ & 14.58\%~/~\textbf{85.42\%} & 3.13\%~/~\textbf{96.87\%} & 6.25\%~/~\textbf{93.75\%}\\
        TV score~$\uparrow$ & - & - & 5.15~/~\textbf{7.11} \\
        FID score~$\downarrow$ & - & - & 169.22~/~\textbf{167.06}\\
        \bottomrule
    \end{tabular}
    \label{tab:evaluation}
\end{table*}
\begin{table*}[t]
        \vspace{-3mm}
    \caption{\textbf{Computing-time comparison for the artistic and photo-realistic style transfer methods.}}
            \vspace{-3mm}
    \small
    \centering
    \begin{tabular}{lccc|ccc}
        \toprule
        Method & Gatys~\etal~\cite{Gatys2016} & WCT~\cite{li2017universal} & \textbf{ArtNet(WCT)} & Luan~\etal~\cite{luan2017deep} & PhotoWCT~\cite{li2018closed} & \textbf{PhotoNet(WCT)} \\
        \midrule
        $256\times128$ & 6.01 & 0.36 & \textbf{0.34} & 114.11 & 4.07 & \textbf{0.76}\\
        $512\times256$ & 17.84 & 0.71 & \textbf{0.42} & 293.28 & 20.72 & \textbf{0.86}\\
        $768\times384$ & 38.18 & 1.20 & \textbf{0.45} & 628.24 & 53.05 & \textbf{0.95}\\
        $1024\times512$ & 66.24 & 1.88 & \textbf{0.52} & 947.61 & 133.90 & \textbf{1.06}\\
        \bottomrule
    \end{tabular}
        \vspace{-3mm}
\label{tab:efficiency}
\end{table*}
\section{ArtNet and PhotoNet: Architectural Approaches for Fast Universal Style Transfer}
%
The network architectural design of ArtNet and PhotoNet consists of three elements or operations:  (i)~Deep Feature Aggregation, (ii)~Multi-stage Stylization, and (iii)~Normalized Skip Connection,  as detailed below.  

\textbf{Deep Feature Aggregation.} Inspired by the PSPNet~\cite{zhao2017pspnet} and DLA~\cite{yu2018deep} for semantic segmentation, we introduce the deep feature aggregation operation to concatenate the multi-level features and improve the image reconstruction quality of the auto-encoder.

\textbf{Multi-stage Stylization.} We utilize the transfer module at different stages of the decoder besides in the middle of the auto-encoder to improve style transfer effects and as a replacement of the multi-level stylization used by WCT~\cite{li2017universal} and PhotoWCT~\cite{li2018closed} to speed up the algorithm.

\textbf{Normalized Skip Connection.} We employ skip connections as used in ~\cite{ronneberger2015u} and coupled with instance normalization~\cite{ulyanov1607instance} to encourage the decoder to preserve more details when reconstructing images.
%


\subsection{ArtNet architectural design}\vspace{-1mm}
%
%
%
%
In order to utilize the multi-level features and avoid the usage of the time-consuming multi-level stylization, we introduce two strategies to ArtNet to improve the image reconstruction quality and thereby enhance the artistic stylization performance.

As Fig.~\ref{fig:architecture} (c) shows, the proposed ArtNet uses the pre-trained VGG-19 as the encoder and utilizes a structurally symmetric decoder to invert deep features back to images.
We apply deep feature aggregation to concatenate and fuse multi-level features.
Moreover, two additional transfer modules are placed at the end of the first two stages of the decoder to improve the stylization performance and as a replacement of the multi-level stylization strategy in WCT~\cite{li2017universal}.
Fig.~\ref{fig:reconstruction} (c) demonstrates that ArtNet outperforms the auto-encoder used by AdaIN~\cite{huang2017arbitrary} and WCT~\cite{li2017universal} regarding the quality of image reconstruction.

\vspace{-1.5mm}
\subsection{PhotoNet architectural design}\vspace{-1mm}
\label{sec:method_photonet}
%
The vanilla auto-encoder and ArtNet are not sufficient to reconstruct large amounts of fine details due to the distortion of lines and shapes in the inverted images (Fig.~\ref{fig:reconstruction}).
Although such distortions make the generated images look more like art creations in the artistic stylization task, they seriously harm the photorealistic stylization effects due to the decrease of the visual authenticity of images.
%
%
%
%

On the basis of ArtNet, our PhotoNet additionally adopts normalized skip connections to straightforwardly introduce low-level information from the encoder to the corresponding decoder stages to improve the image reconstruction effects.
Moreover, we place transfer modules at normalized skip connections and every stage of the decoder to improve the stylization quality as shown in Fig.~\ref{fig:architecture} (d).
As demonstrated in Fig.~\ref{fig:reconstruction} (e), the proposed PhotoNet outperforms the auto-encoder used by AdaIN~\cite{huang2017arbitrary} and PhotoWCT~\cite{li2018closed} in faithfully inverting deep features back to images.
With PhotoNet coupled with the ZPA transform in WCT~\cite{li2017universal}/PhotoWCT~\cite{li2018closed} as the feature stylization module, PhotoNet avoids the use of post-processing as well as time-consuming optimization and achieves visually pleasing transfer results with rich details and is more than 600$\times$ faster than Luan~\etal~\cite{luan2017deep} and 50$\times$ faster than Li~\etal~\cite{li2018closed} on large images.

\vspace{-1.5mm}
\subsection{Decoder training}\vspace{-1mm}
We train the decoder of ArtNet and PhotoNet to invert deep features back to images with the Frobenius norm of the original and inverted image as the reconstruction loss,
\begin{equation}
    \mathcal{L}_{recon} = \|I_{in} - I_{out}\|_F, 
    \label{eq:loss_recon}
\end{equation}
where $I_{in}$ denotes the input image, $I_{out}$ is reconstructed image, and $\|\cdot\|_F$ represents the Frobenius norm.
Inspired by Li~\etal~\cite{li2017universal}, we introduce the perceptual loss term~\cite{johnson2016perceptual} to improve the reconstruction quality of the decoder,
\begin{equation}
    \mathcal{L}_{percep} = \sum\limits_{i=1}^5 \|\Phi_i\left( I_{in} \right) - \Phi_i\left( I_{out} \right)\|_F,
    \label{eq:loss_percep}
\end{equation}
where $\Phi_i\left( \cdot \right)$ denotes the output of the $i^{th}$ stage of the ImageNet~\cite{deng2009imagenet} pre-trained VGG-19~\cite{simonyan2014very}.
The overall loss function is,
\begin{equation}
    \mathcal{L} = \alpha \mathcal{L}_{recon} + \left( 1 - \alpha \right)\mathcal{L}_{percep},
    \label{eq:loss}
\end{equation}
where the $\alpha$ balances two loss terms.
During training, all transfer modules in ArtNet and PhotoNet are skipped and the whole framework is trained in an image reconstruction manner.

\begin{figure*}[t]
    \centering
    \subfloat[Content]{\includegraphics[width=0.13\linewidth]{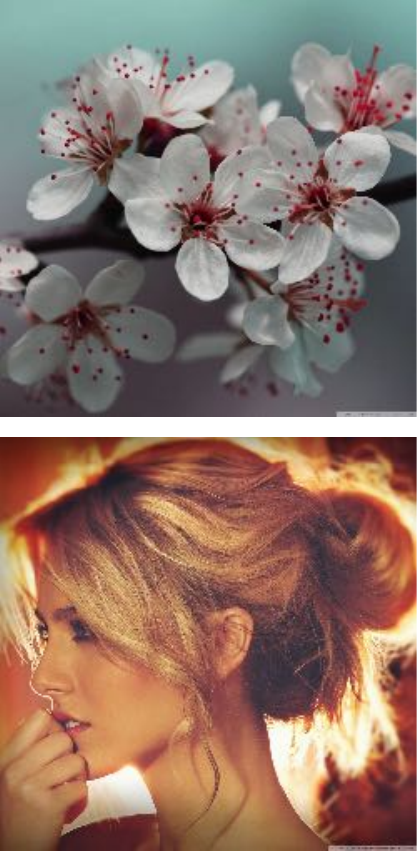}}\hspace{1mm}
    \subfloat[Style]{\includegraphics[width=0.13\linewidth]{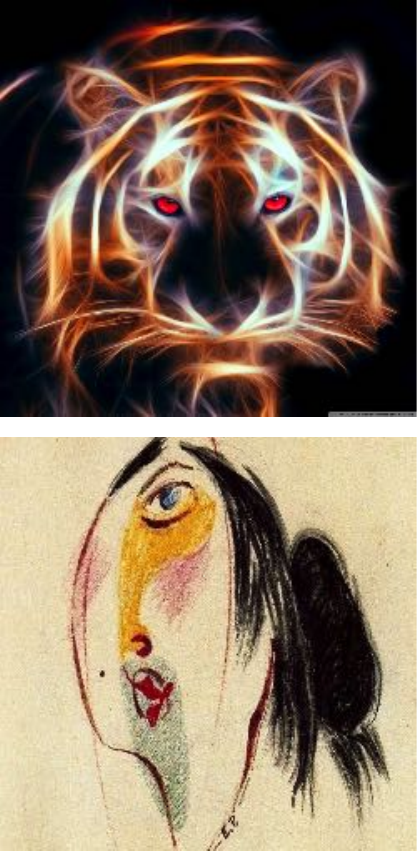}}\hspace{1mm}
    \subfloat[Gatys~\etal~\cite{Gatys2016}.]{\includegraphics[width=0.13\linewidth]{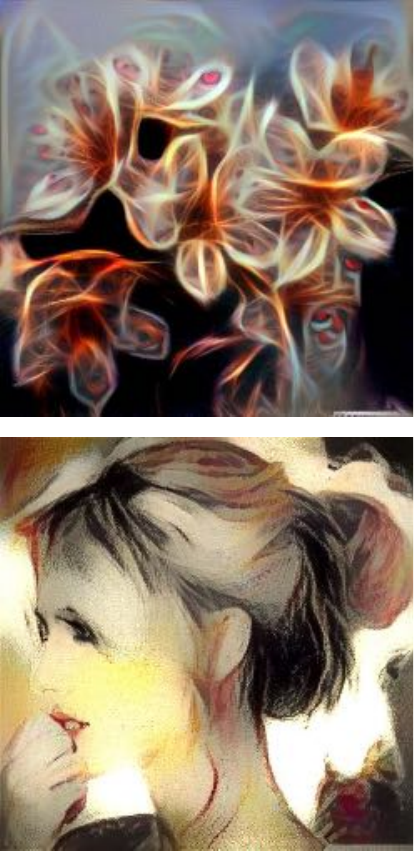}}\hspace{1mm}
    \subfloat[AdaIN~\cite{huang2017arbitrary}.]{\includegraphics[width=0.13\linewidth]{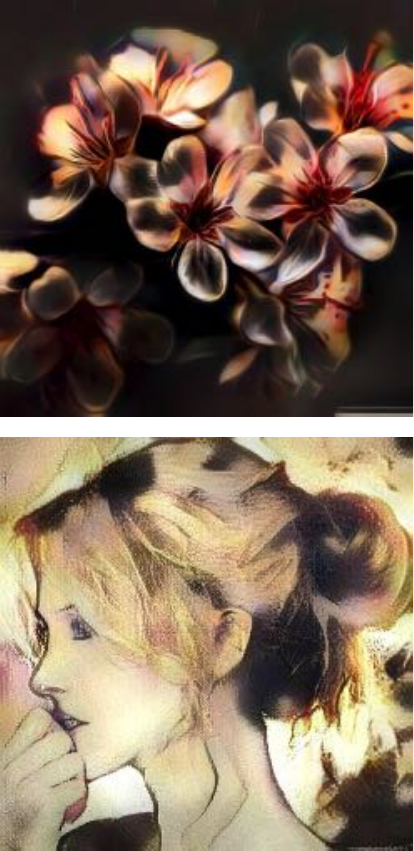}}\hspace{1mm}
    \subfloat[WCT~\cite{li2017universal}.]{\includegraphics[width=0.13\linewidth]{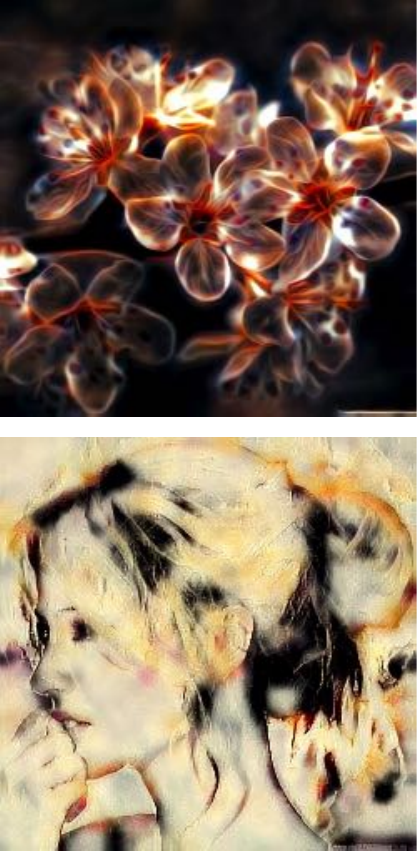}}\hspace{1mm}
    \subfloat[Avatar-Net~\cite{sheng2018avatar}.]{\includegraphics[width=0.13\linewidth]{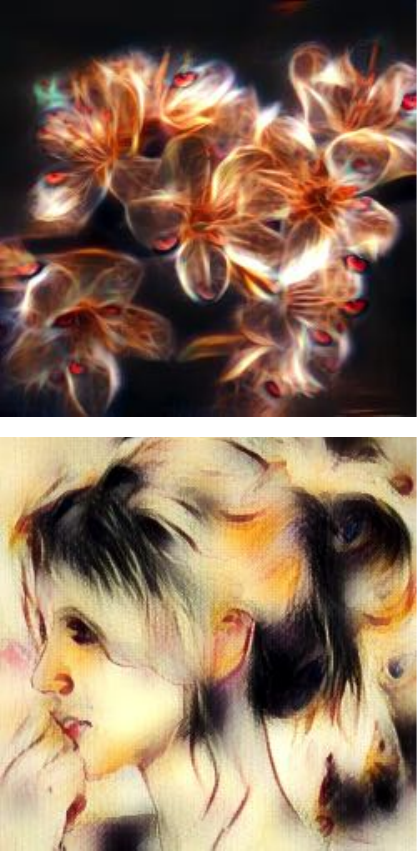}}\hspace{1mm}
    \subfloat[ArtNet (WCT)]{\includegraphics[width=0.13\linewidth]{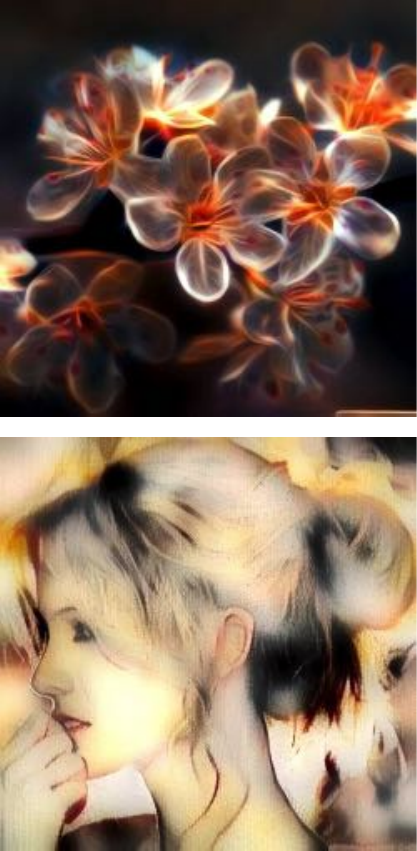}}
        \vspace{-3mm}
\caption{\textbf{Artistic stylization results.}}
        \vspace{-5mm}
\label{fig:artistic_stylization}
\end{figure*}

\begin{figure*}[t]
    \centering
    \subfloat[Content]{\includegraphics[width=0.133\linewidth]{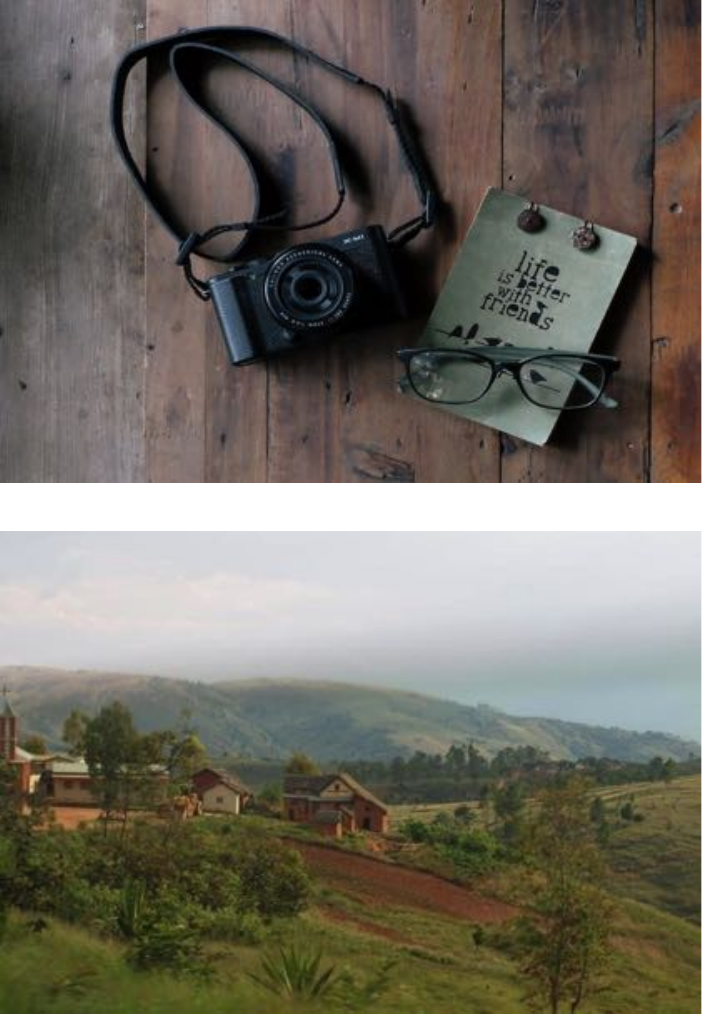}}\hspace{1mm}
    \subfloat[Style]{\includegraphics[width=0.133\linewidth]{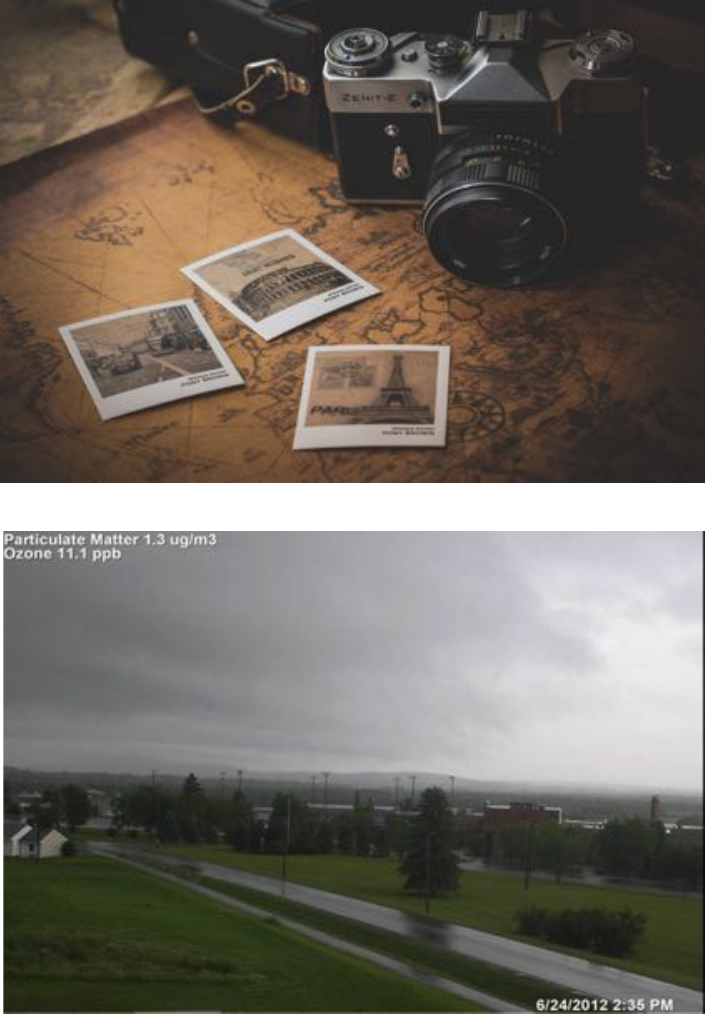}}\hspace{1mm}
    \subfloat[IDT~\cite{pitie2005n}.]{\includegraphics[width=0.17\linewidth]{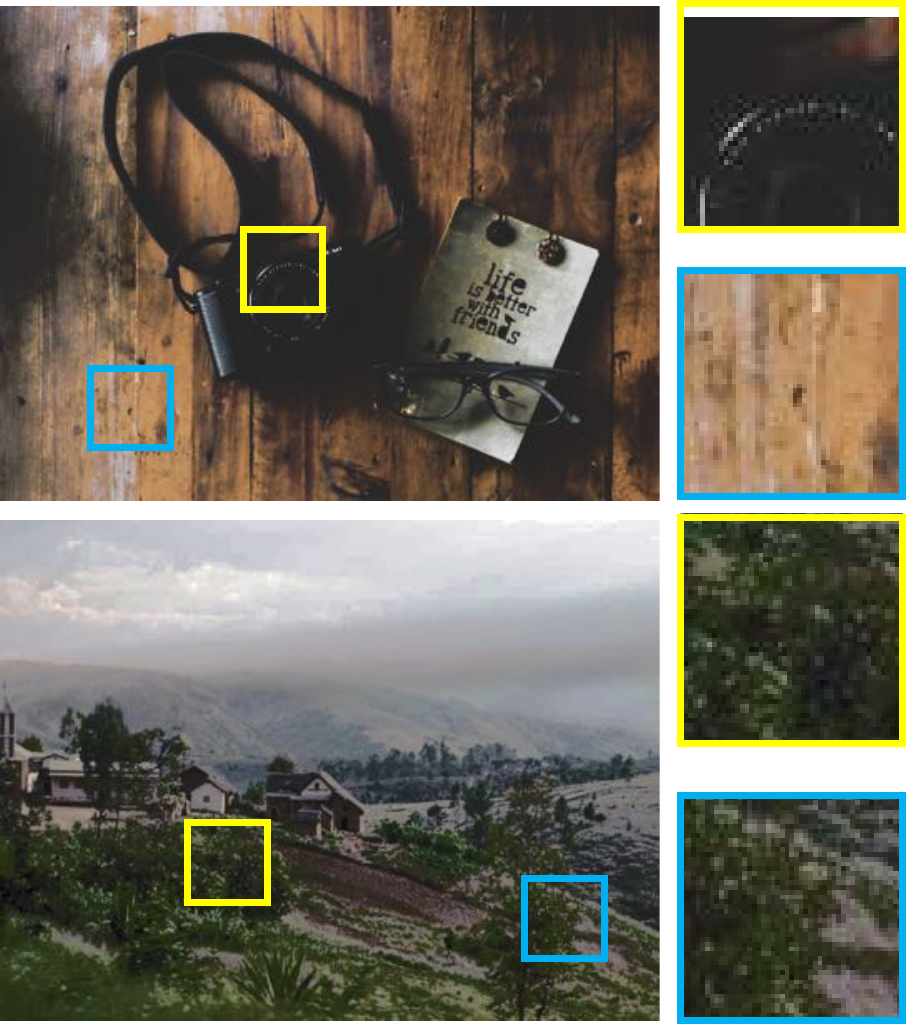}}\hspace{1mm}
    \subfloat[Luan~\etal~\cite{luan2017deep}.]{\includegraphics[width=0.17\linewidth]{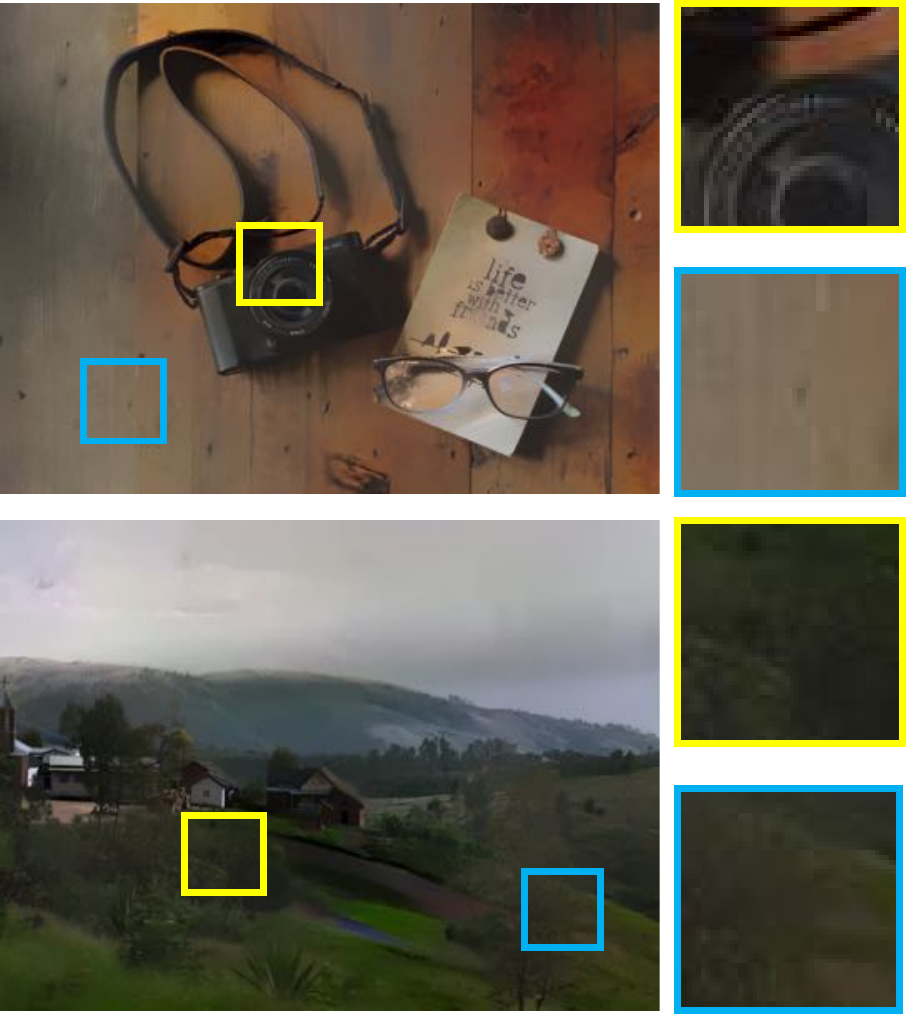}}\hspace{1mm}
    \subfloat[PhotoWCT~\cite{li2018closed}.]{\includegraphics[width=0.17\linewidth]{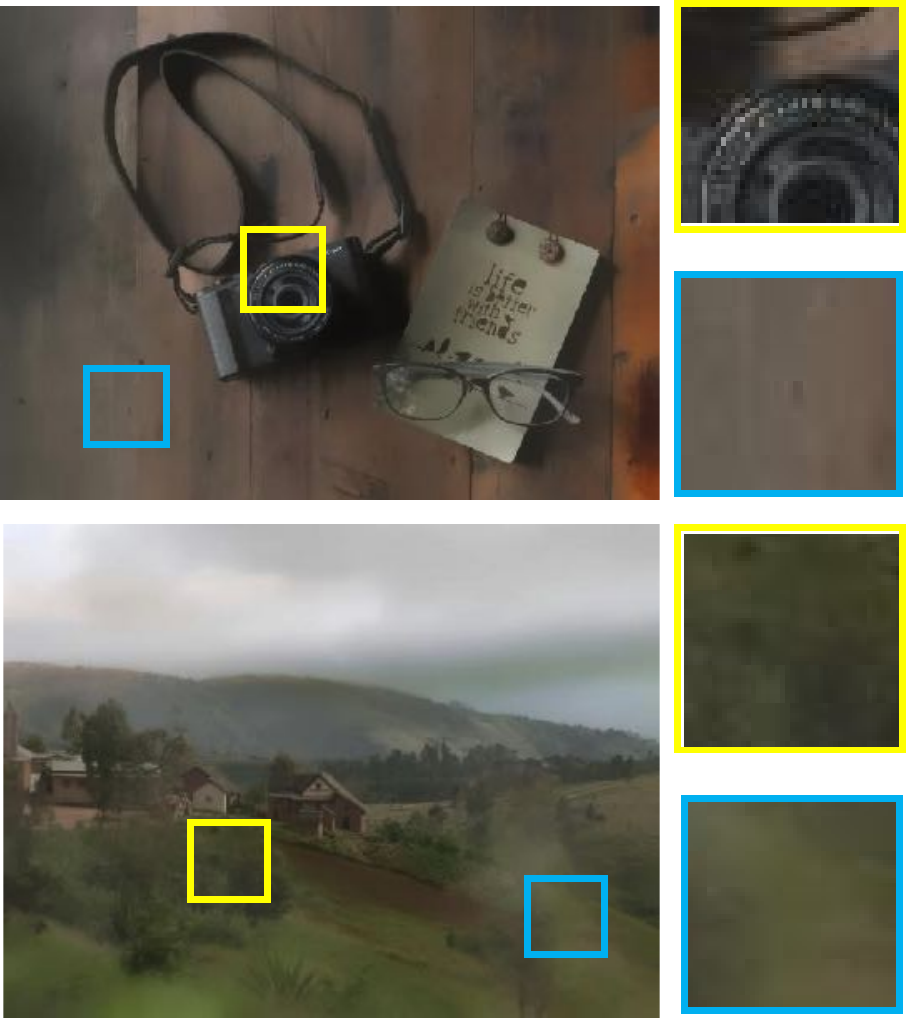}}\hspace{1mm}
    \subfloat[PhotoNet (WCT)]{\includegraphics[width=0.17\linewidth]{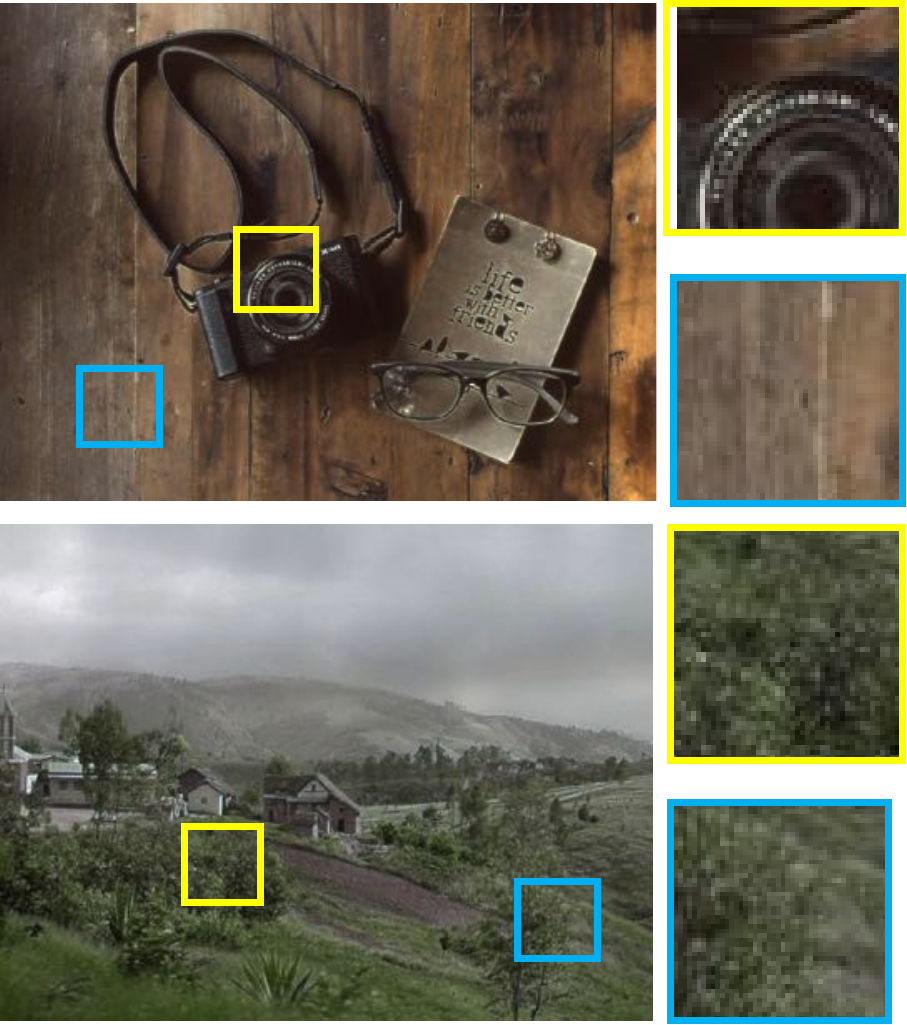}}
        \vspace{-3mm}
\caption{\textbf{Photo-realistic stylization results.} Note that the method of Luan~\etal~\cite{luan2017deep} requires additional segmentation masks for stylization while other compared algorithms do not.}
        \vspace{-5mm}
\label{fig:photorealistic_stylization}
\end{figure*}
\begin{figure*}[t]
    \centering
    \subfloat[Input (Artistic)]{\includegraphics[width=0.15\linewidth]{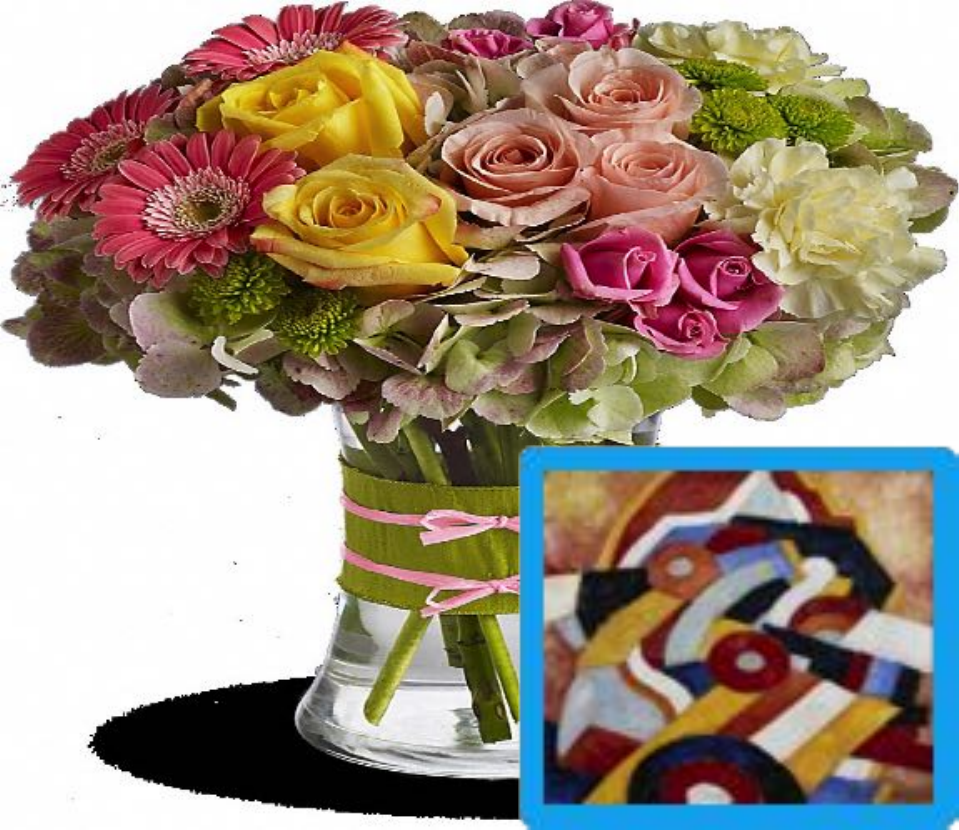}}\hspace{0.5mm}
    \subfloat[Base]{\includegraphics[width=0.15\linewidth]{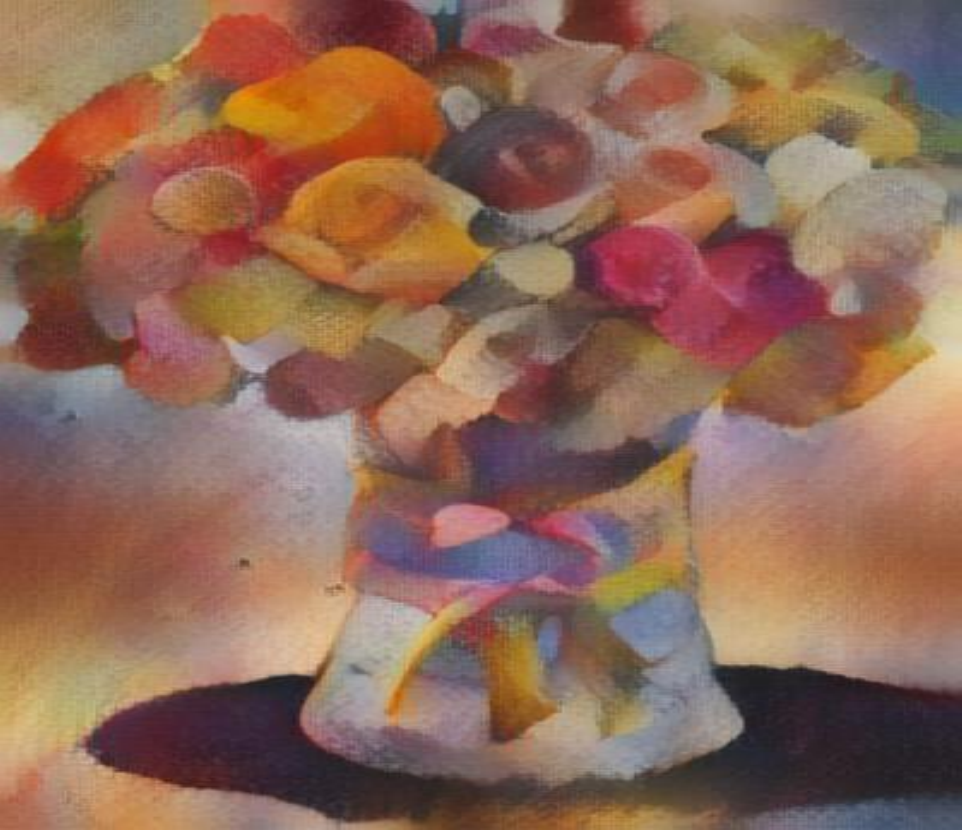}}\hspace{0.5mm}
    \subfloat[Base+FA]{\includegraphics[width=0.15\linewidth]{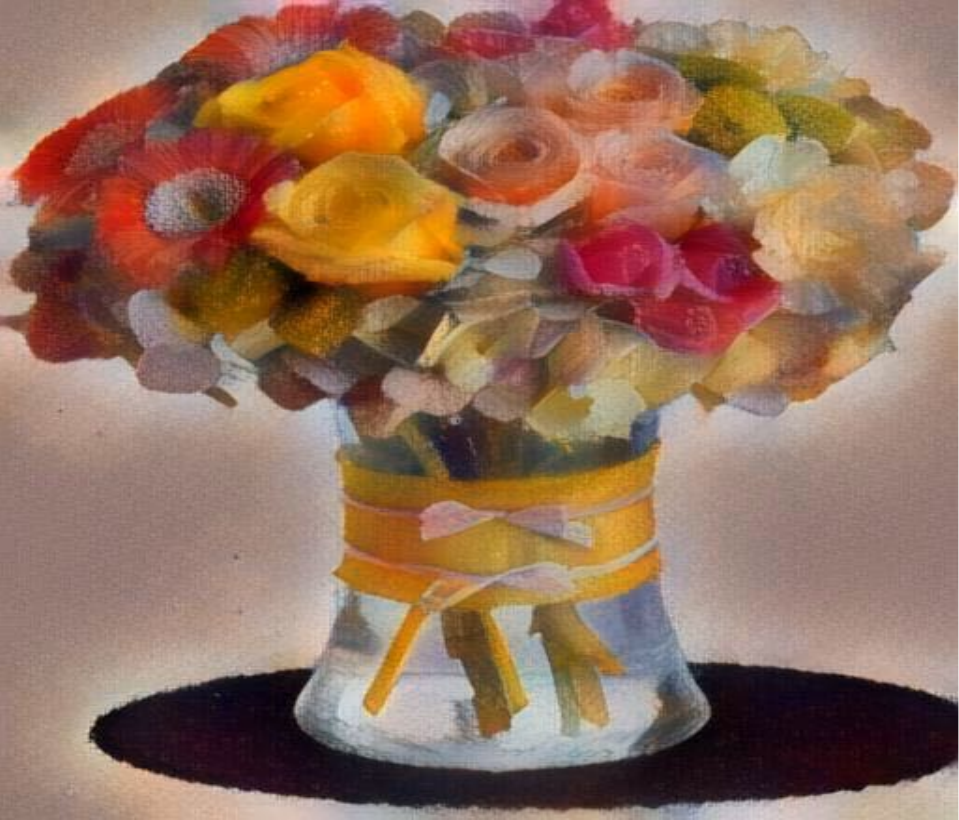}}\hspace{0.5mm}
    \subfloat[Base+MST-3]{\includegraphics[width=0.15\linewidth]{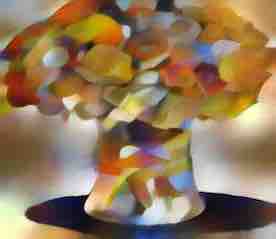}}\hspace{0.5mm}
    \subfloat[Base+FA+MST-3]{\includegraphics[width=0.15\linewidth]{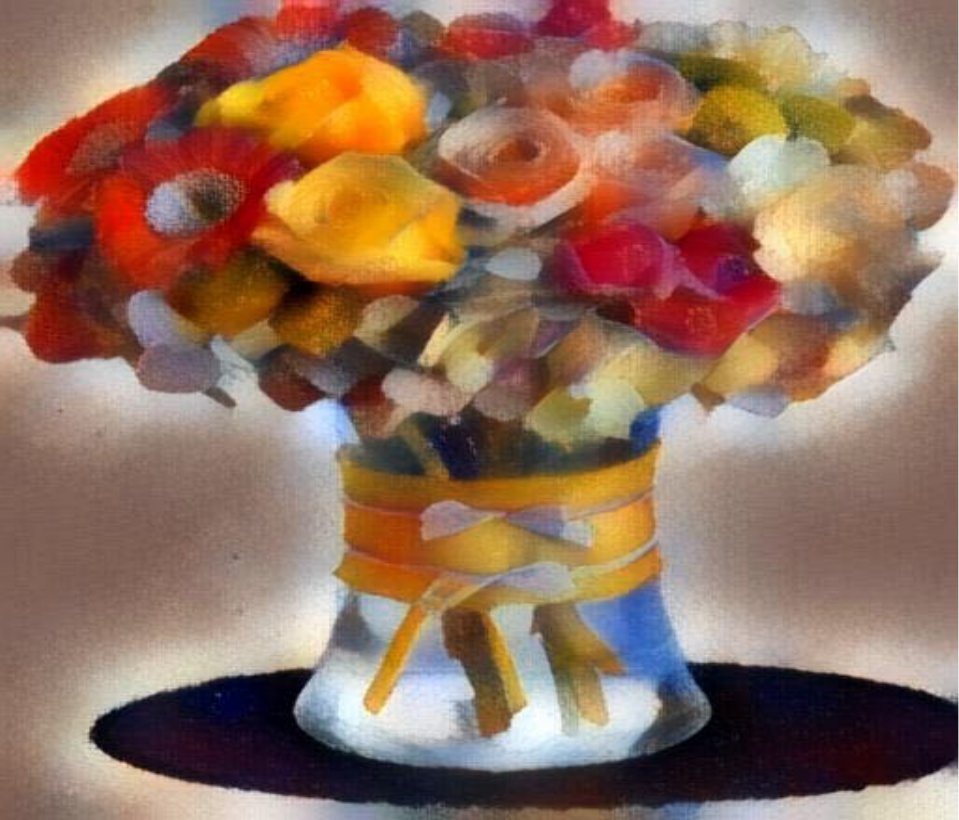}}\hspace{0.5mm}
    \subfloat[Base+FA+MST-5]{\includegraphics[width=0.15\linewidth]{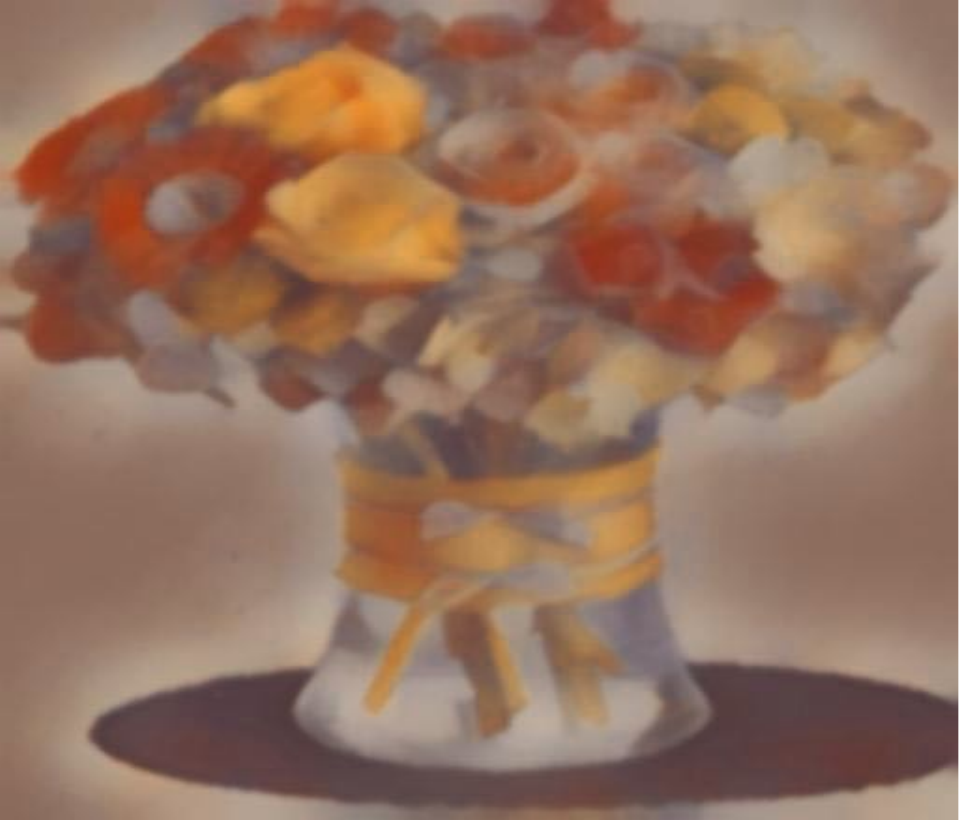}}\\
    \subfloat[Input (Photorealistic)]{\includegraphics[width=0.18\linewidth]{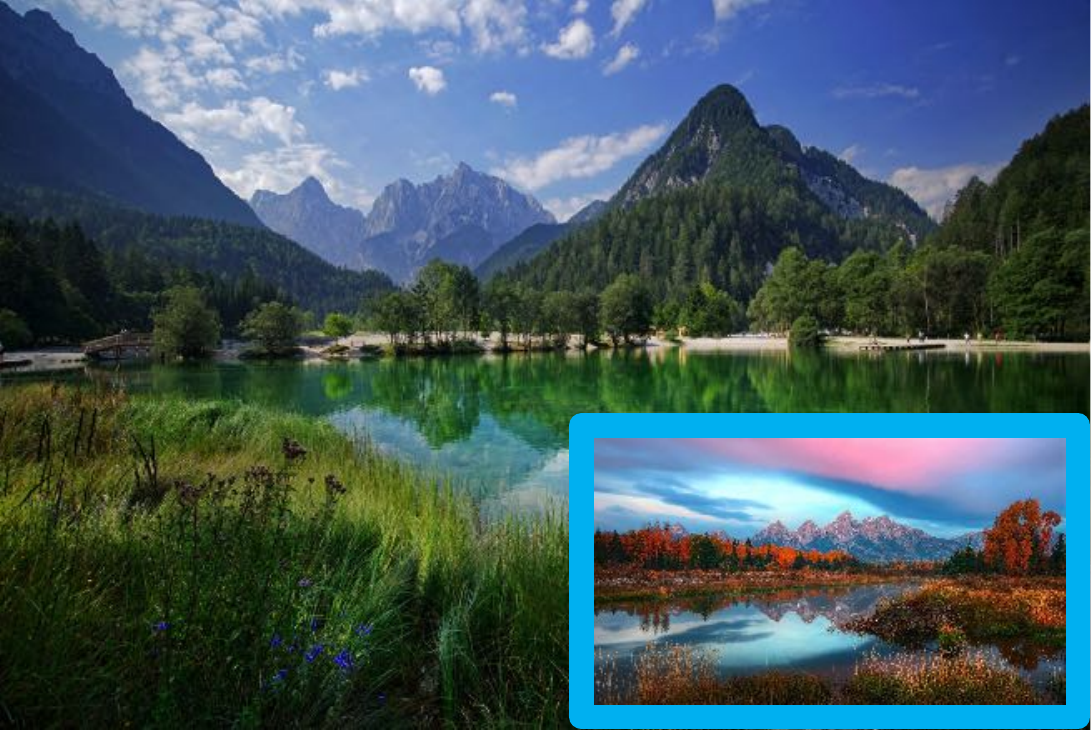}}\hspace{0.5mm}
    \subfloat[Base+FA]{\includegraphics[width=0.18\linewidth]{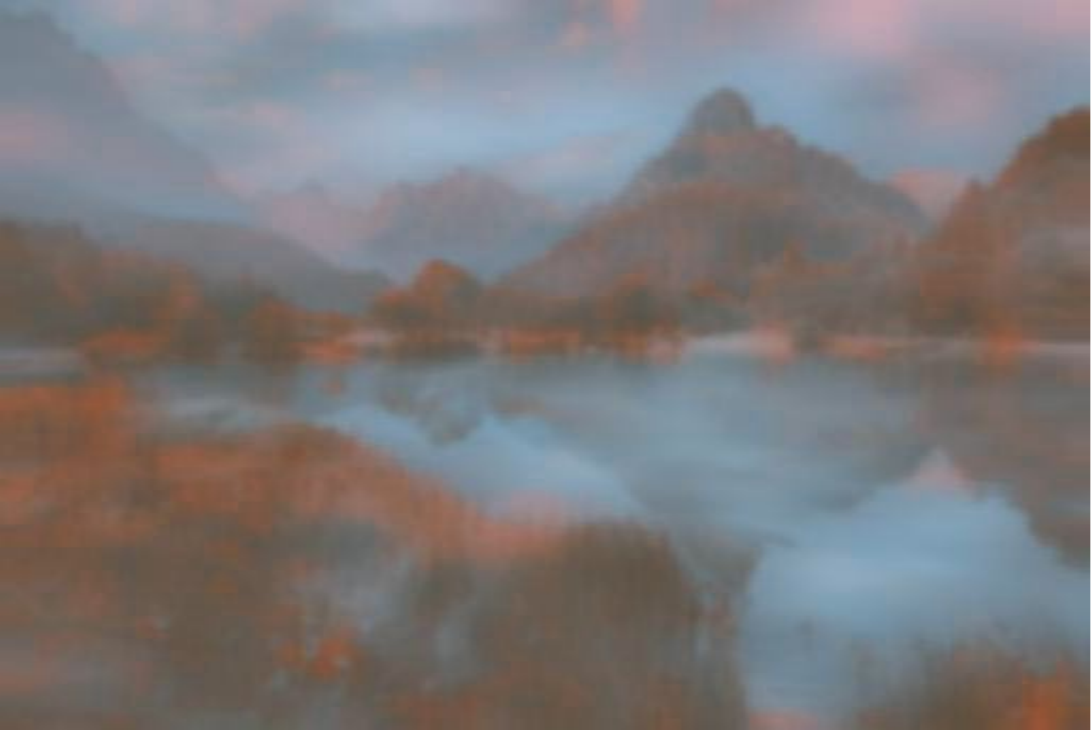}}\hspace{0.5mm}
    \subfloat[Base+FA+MST-5]{\includegraphics[width=0.18\linewidth]{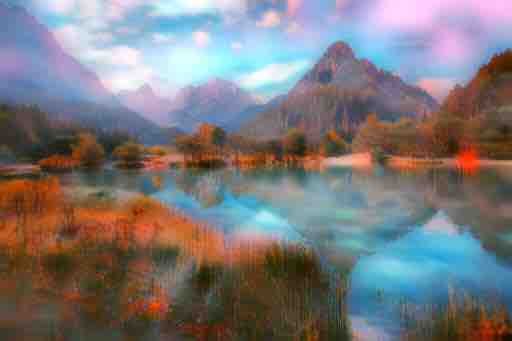}}\hspace{0.5mm}
    \subfloat[Base+FA+NS+MST-5]{\includegraphics[width=0.18\linewidth]{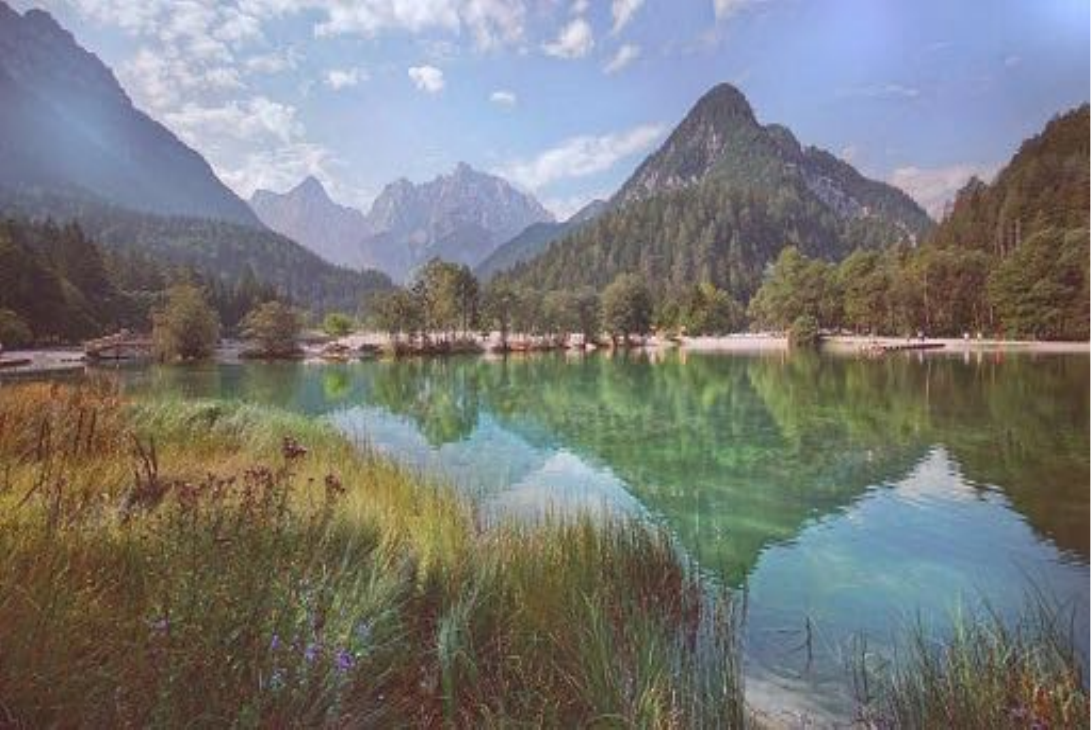}}\hspace{0.5mm}
    \subfloat[Base+FA+NS+MST-5+MST-$\infty$]{\includegraphics[width=0.18\linewidth]{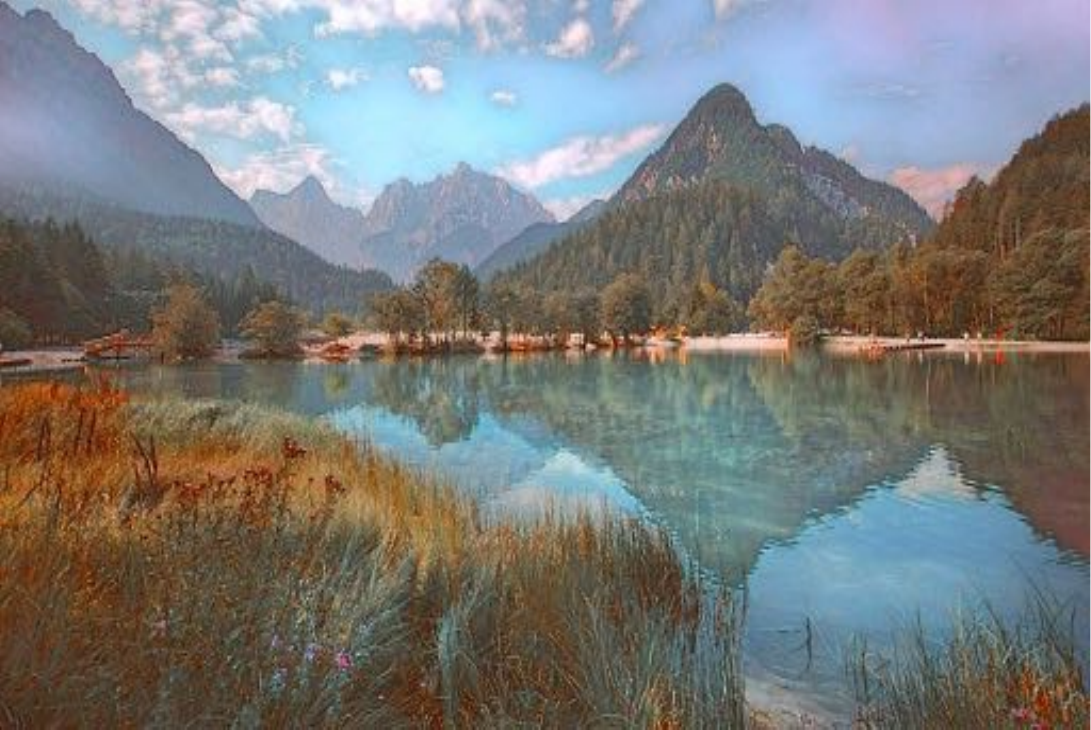}}
    \vspace{-3mm}
    \caption{\textbf{Ablation study of ArtNet and PhotoNet.}}
    \vspace{-5mm}
    \label{fig:ablation}
\end{figure*}
\section{Experiments and Empirical Validations}
In this section, we present the stylization results of our proposed algorithms in comparison with those of the state-of-the-art approaches and further provide a comprehensive empirical analysis to substantiate our observations.
%
%

\subsection{Results on artistic style transfer}
In order to demonstrate the effectiveness of the proposed ArtNet, we conduct contrast experiments on AdaIN~\cite{huang2017arbitrary} and WCT~\cite{li2017universal}, where we replace the AE part of these two approaches with ArtNet and keep other part fixed.
As Fig.~\ref{fig:comp_artistic} shows, the results by ~\cite{huang2017arbitrary} contains unpleasant artifacts such as unfaithful edge lines in brick edges (Row 2), the face (Row 3), and the background of flowers.
Such significant artifacts are clearly eliminated by the proposed method when  integrating ArtNet with AdaIN~\cite{huang2017arbitrary} as the transform.
The WCT~\cite{li2017universal} method generates images with significant distortions and a lack of local similarity, fragmented color blobs (Row 1) and twisted lines (Row 2).
In contrast, ArtNet with WCT~\cite{li2018closed} as transform creates images with straight lines (Row 2), clear color blobs (Row 1) and a clean face (Row 3).
%
%
%
To demonstrate the exceptional performance of ArtNet coupled with the ZCA in WCT~\cite{li2017universal} as the transfer module, we compare the artistic stylization results against the state-of-the-art methods.
As shown in Fig.~\ref{fig:artistic_stylization}, the method by Gatys~\etal~\cite{Gatys2016} generates images with areas of artifacts (badly stylized areas in the image of the top row and overexposed background of the human in the image in the bottom row.).
The results of AdaIN~\cite{huang2017arbitrary} contain evident artifacts that render inaccurate shapes and color blobs in flowers (Row 1) and unnatural hair of the girl (Row 2).
The WCT~\cite{li2017universal} method distorts and twists the shapes, lines, and color blobs of the transferred images, while Avatar-Net~\cite{sheng2018avatar} creates artifacts by rendering artistically matching patches to the transferred images but disregarding their semantics. This is demonstrated by  Fig.~\ref{fig:artistic_stylization} (f) where Avatar-Net~\cite{sheng2018avatar} renders ``red eyes'' (Row 1) as well as ``blue eyes and red lips'' (Row 2) arbitrarily in the generated images.

\subsection{Results on photorealistic style transfer}
We verify the effectiveness of the proposed PhotoNet (using WCT~\cite{li2017universal} as transform) by the comparison with the photorealistic stylization results of PhotoWCT~\cite{li2018closed}.
%
%
%
%
As shown in Fig.~\ref{fig:comp_photorealistic} (c), the results of PhotoWCT~\cite{li2018closed} without the post-processing for smoothing contain apparent distortions such that the sky in images in the second and third rows are distorted and twisted.
The transferred images of PhotoWCT~\cite{li2018closed} with the smoothing operation turned on are overly smooth and have low sharpness in details such that grasses (Row 1), steel frame of the Louvre (Row 2), and trees (Row 3) have been smoothed out and lost their details. 
We additionally compare the results of PhotoNet with the algorithm by Luan~\etal~\cite{luan2017deep} and PhotoWCT~\cite{li2018closed} approach.
Fig.~\ref{fig:photorealistic_stylization} shows that PhotoNet renders the leaves in trees (Row 2), the texture of woods and details of the camera lens (Row 1) considerably sharper and transfer styles much more faithfully,  demonstrating that PhotoNet outperforms the compared methods in generating visually pleasing and sharp images.
%
%

\subsection{Computational time comparison}
We conduct a computing time comparison against the state-of-the-art methods to demonstrate the efficiency of the proposed network architectures.
All approaches are tested on the same computing platform which includes an NVIDIA Tesla P100 with 16GB RAM.
We compare the computing time on content and style images with different resolutions.

\textbf{ArtNet.}~As Table~\ref{tab:efficiency} shows, Gatys~\etal~\cite{Gatys2016} is slow due to the optimization process, WCT~\cite{li2017universal} method is considerably faster but not efficient enough due to the usage of multi-level stylization (especially for high-resolution images), while ArtNet improves the  inference speed of WCT~\cite{li2017universal} by three times on large images, thanks to the avoidance of multi-level stylization.
Moreover, ArtNet generates stylized images with fewer artifacts compared with AdaIN~\cite{huang2017arbitrary} with a minor additional time cost.

\textbf{PhotoNet.}~In order to verify the superior efficiency of the proposed PhotoNet, we conduct experiments against baseline photorealistic stylization methods in terms of the computing time.
As Table~\ref{tab:efficiency} demonstrated, PhotoNet is hundreds of times faster than the method of Luan~\etal~\cite{luan2017deep} and tens of times faster than PhotoWCT~\cite{li2018closed}. It is even more time-efficient on high-resolution images.
It is worth mentioning that the method by Luan~\etal~\cite{luan2017deep} requires additional segmentation masks to assist stylization, which costs more computing time.

\vspace{-1mm}
\subsection{Empirical validation}\vspace{-2mm}
In this section, we try to provide more insights into the proposed methods through three empirical studies: (1) We first use quantitative analysis to demonstrate the quality of stylized images using a user study together with  Fr{\'e}chet Inception Distance (FID)~\cite{heusel2017gans} and total variation~\cite{rudin1992nonlinear} (TV) scores, where we are specifically interested in the sharpness of the generated images; (2) We then conduct an ablation study to validate the necessary/contribution of each component (\eg, feature aggregation, multi-stage stylization, and normalized skip connections) included in PhotoNet and ArtNet; (3) We add a case study on the image reconstruction performance of AEs used for style transfer to make sense of our simple intuition that lower AE image reconstruction error leads to better stylization performance. 
\begin{figure*}[!h]
    \centering
    \subfloat[AdaIN~\cite{huang2017arbitrary}.]{\includegraphics[width=0.15\linewidth]{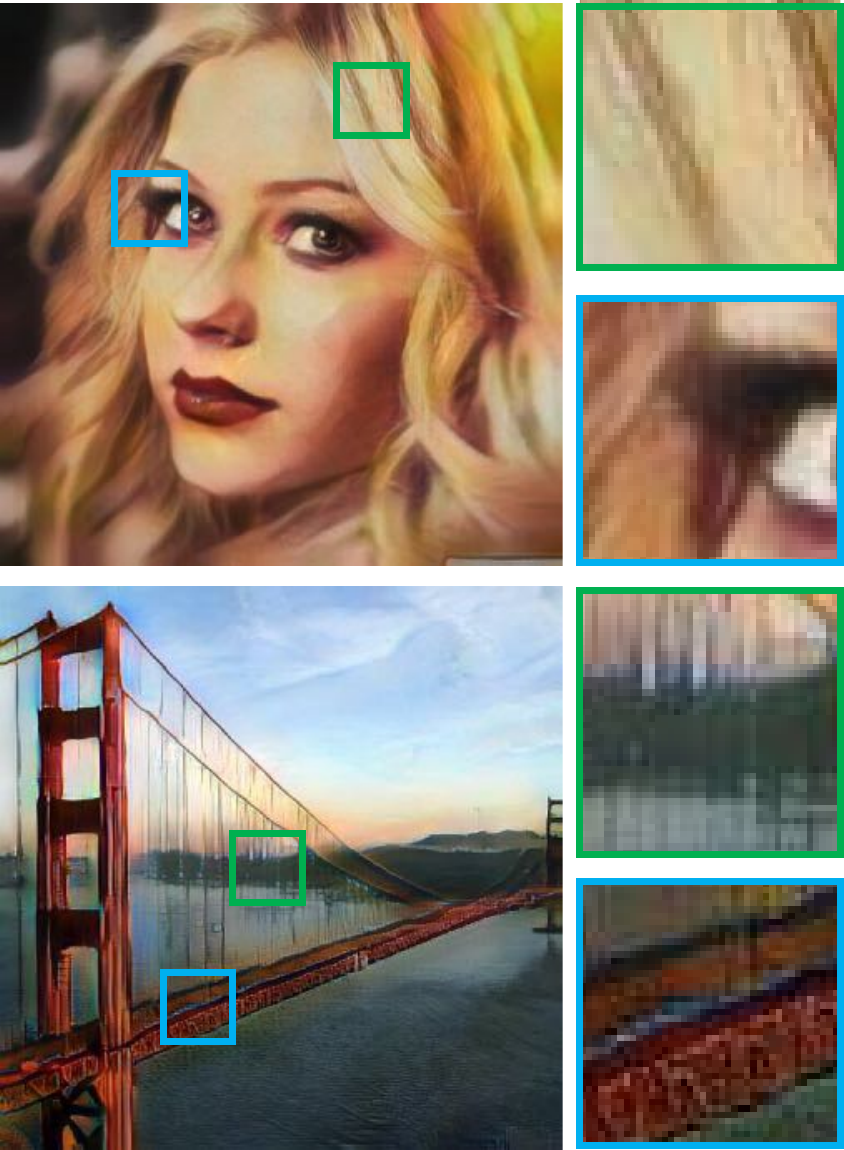}}\hspace{1mm}
    \subfloat[WCT~\cite{li2017universal}.]{\includegraphics[width=0.15\linewidth]{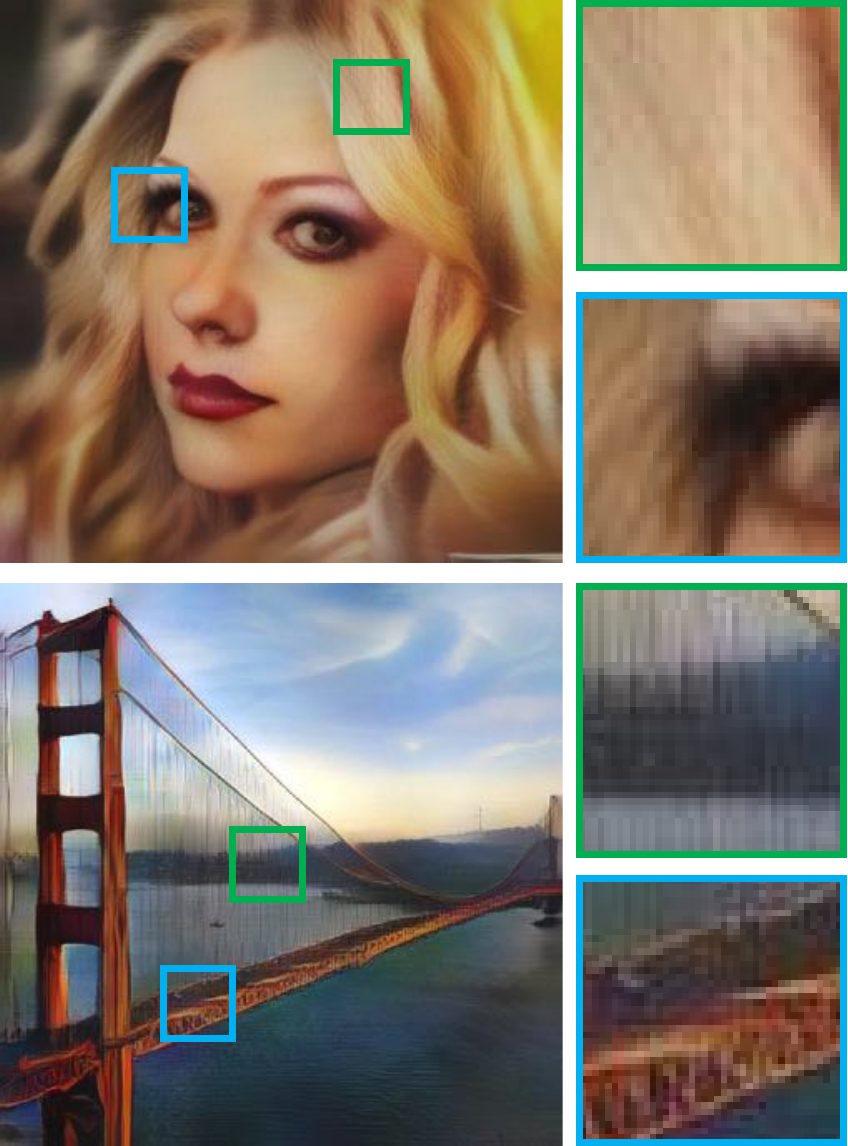}}\hspace{1mm}
    \subfloat[ArtNet]{\includegraphics[width=0.15\linewidth]{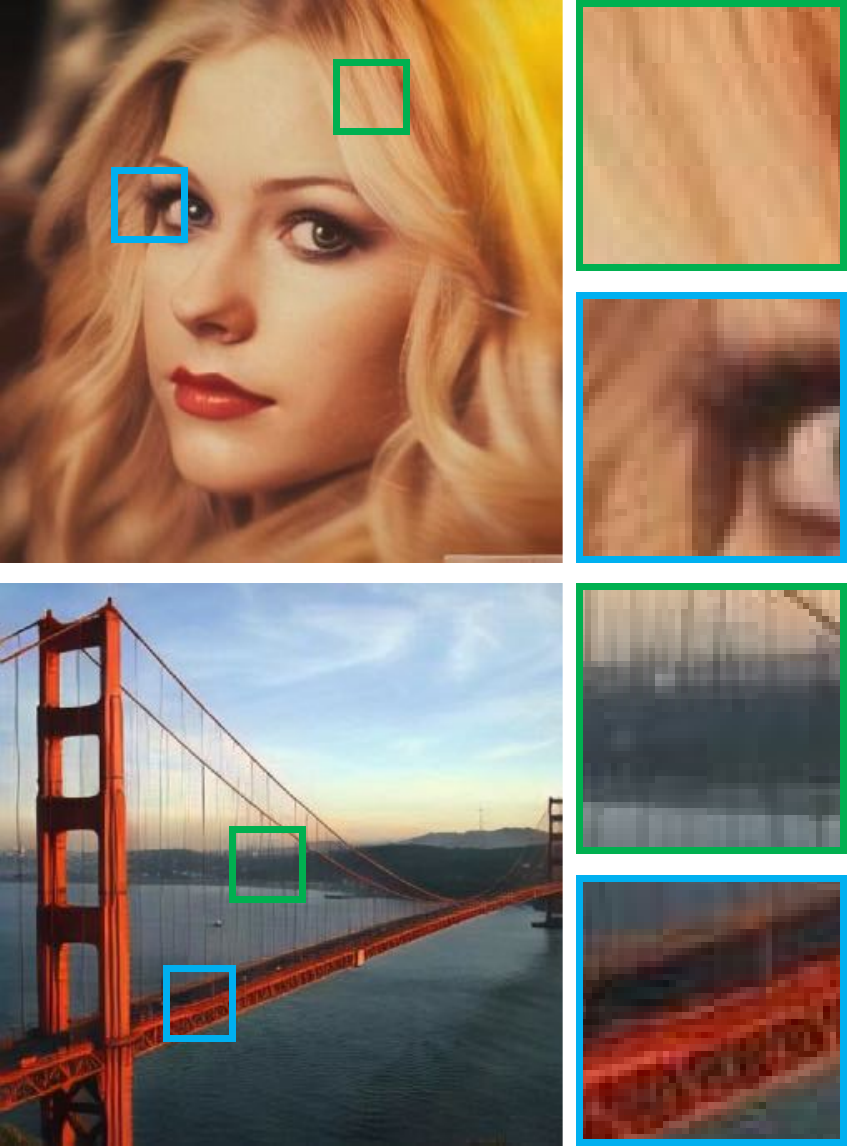}}\hspace{1mm}
    \hspace{0.5mm}
    \vline
    \hspace{2mm}
    \subfloat[PhotoWCT~\cite{li2018closed}.]{\includegraphics[width=0.15\linewidth]{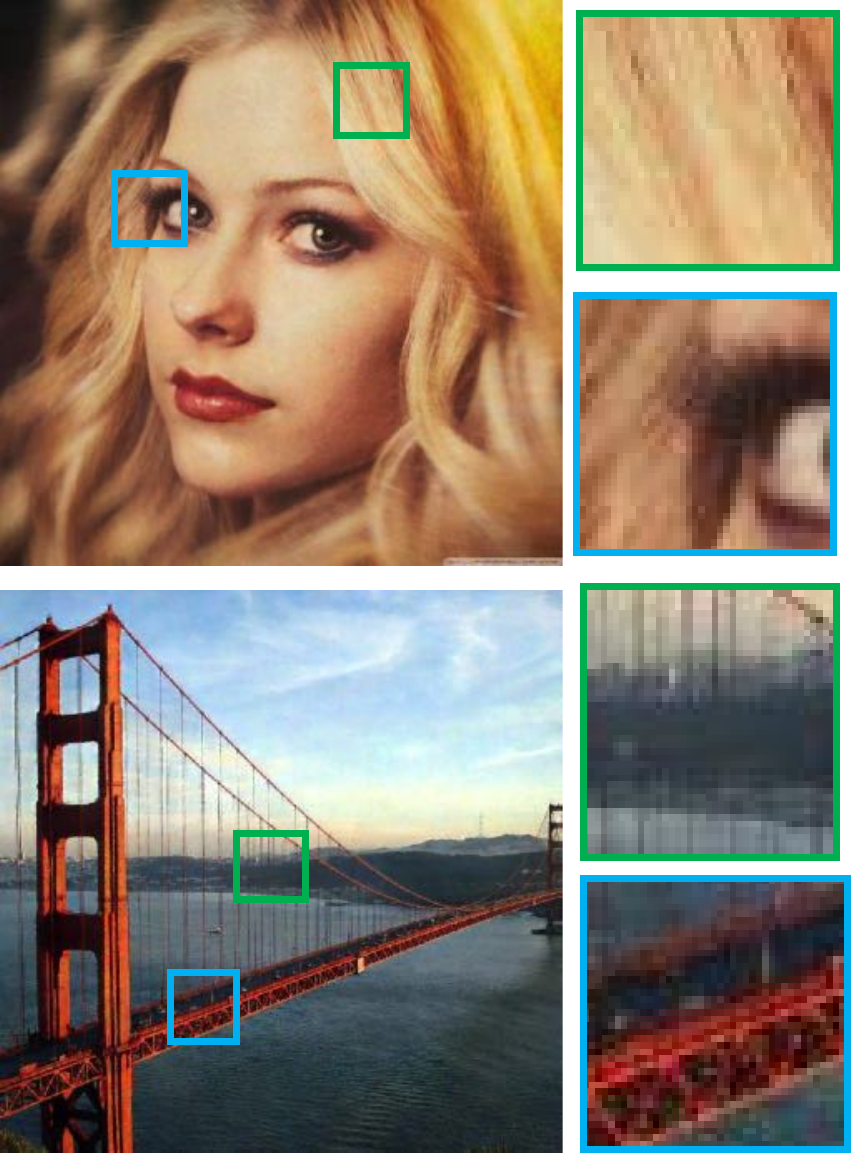}}\hspace{1mm}
    \subfloat[PhotoNet]{\includegraphics[width=0.15\linewidth]{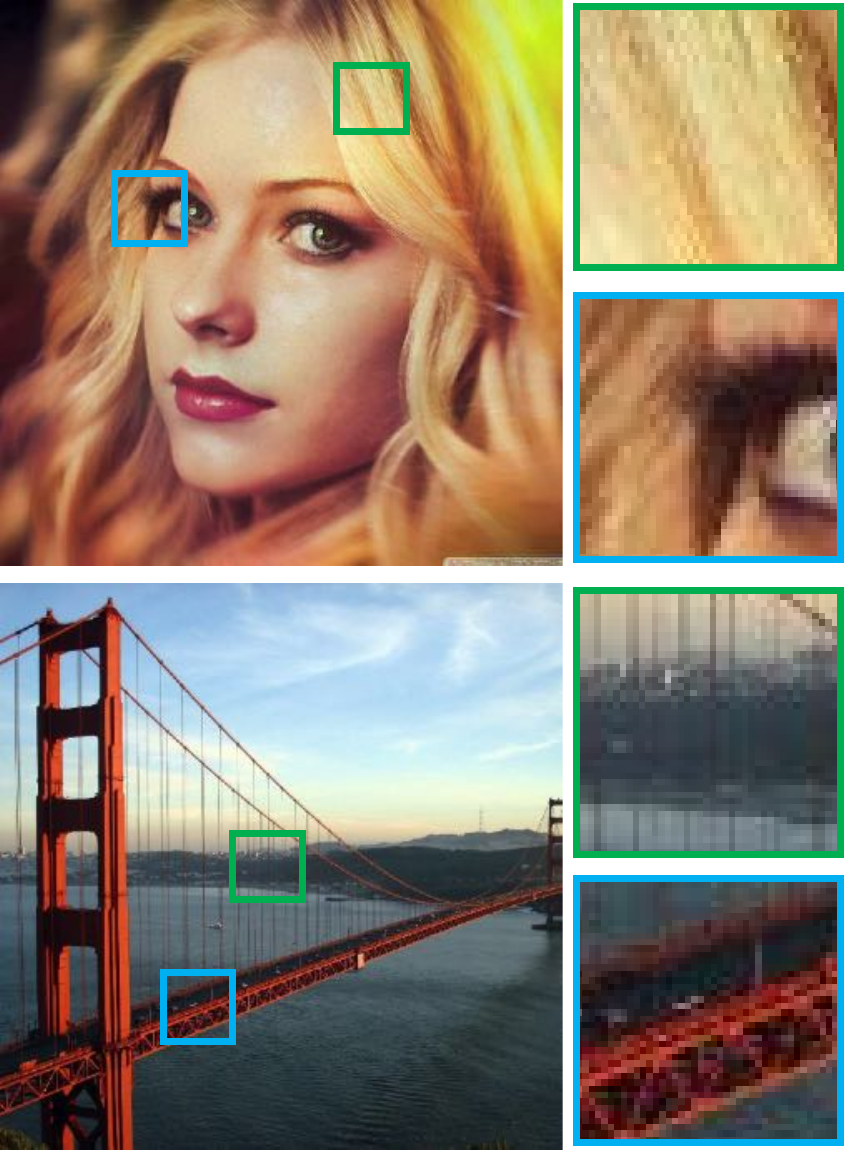}}\hspace{1mm}
    \subfloat[Ground Truth]{\includegraphics[width=0.15\linewidth]{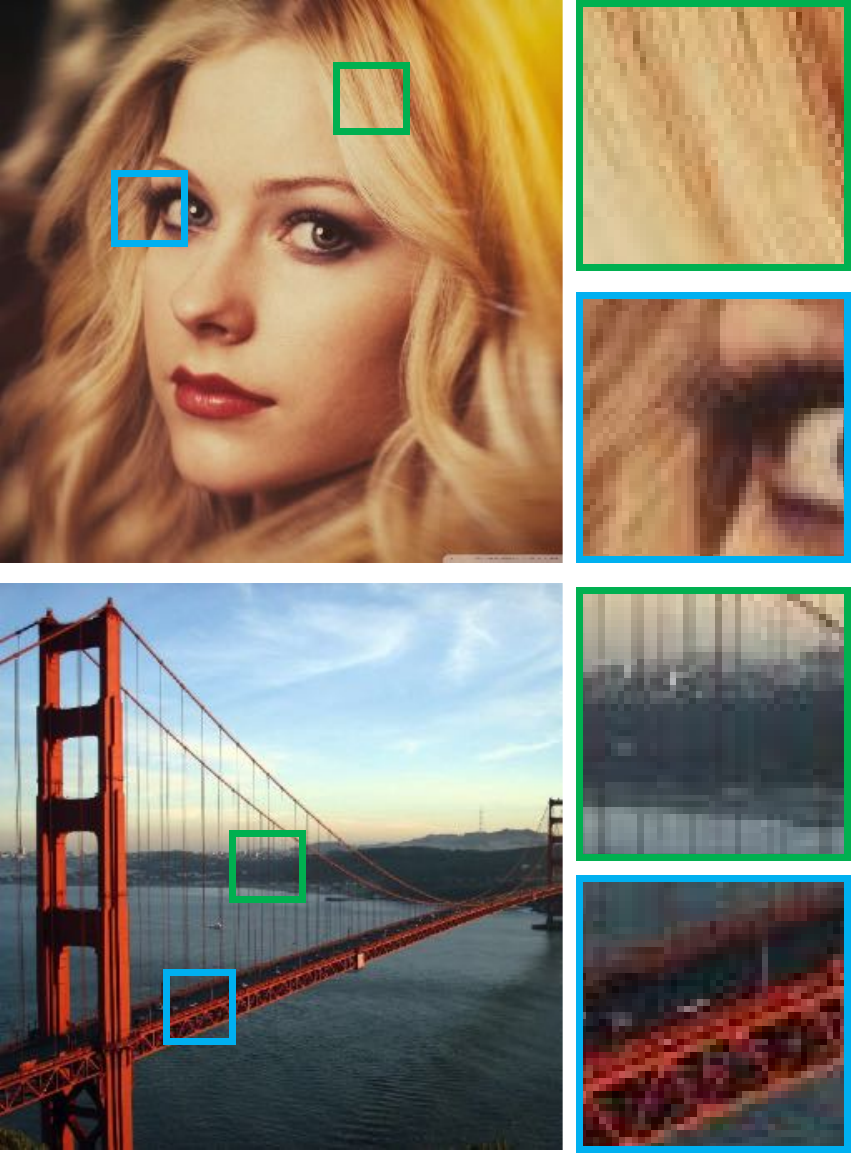}}
        \vspace{-4mm}
        \caption{\textbf{Image reconstruction results.} The proposed PhotoNet and ArtNet outperform WCT~\cite{li2017universal}, AdaIN~\cite{huang2017arbitrary} and PhotoWCT~\cite{li2018closed} in preserving details for image reconstruction (\ie, hairs, eye-slashes, and the local structure/textures in both artistic and photorealistic settings). 
        }
        \vspace{-3mm}
\label{fig:reconstruction}
\end{figure*}

\vspace{-2mm}\subsubsection{Quantitative evaluation}\vspace{-2mm}
%
In this study, artistic style transfer methods are evaluated on a dataset consisting of 12 content images and 16 style images, where each content image is  transferred into every style.
We compute the evaluation metrics and conduct the user study based on totally 192 generated images.
As for photo-realistic cases, the evaluation is based on 38 content images and their corresponding styles.

\textbf{User study.}~We conducted a user study to subjectively demonstrate the effectiveness of the proposed ArtNet and PhotoNet. We randomly select 12 content and style image pairs to evaluate the artistic style transfer methods and use 6 pairs to measure photo-realistic approaches.
For each content and style pairs, we display the results of AdaIN~\cite{huang2017arbitrary}/ArtNet(AdaIN), WCT~\cite{li2017universal}/ArtNet(WCT), and PhotoWCT~\cite{li2018closed}/PhotoNet(WCT) side-by-side and let the subject  choose the better one in terms of less artifact, less distortion, and more details,  respectively.
We collect 16 responses and a total of 288 votes.
The preference percentage of the choices are summarized in Table~\ref{tab:evaluation}, which demonstrates that using ArtNet improves over the  stylization results of WCT~\cite{li2017universal} in terms of less distortion and AdaIN~\cite{huang2017arbitrary} in terms of fewer artifacts, while PhotoNet improves over the results of PhotoWCT~\cite{li2018closed} in terms of more sharp  details.

\textbf{FID}~\cite{heusel2017gans}.~
We compute the FID score between the reference style images and transferred images by PhotoNet and state-of-the-art methods, \ie, PhotoWCT~\cite{li2018closed} for comparison.
As Table~\ref{tab:evaluation} shows, PhotoNet using WCT~\cite{li2017universal} as the transfer module outperforms PhotoWCT~\cite{li2018closed} with a higher FID score (\ie, better stylization). Note that FID was originally used to validate the image quality for domain adaption and image translation, which is close to photorealistic stylization.

\textbf{Total variation}~\cite{rudin1992nonlinear}.~We compare the total variation scores of the results by PhotoNet and PhotoWCT~\cite{li2018closed} methods.
As demonstrated in Table~\ref{tab:evaluation}, images generated by PhotoNet are of  higher total variation scores (\ie, more sharpness and details)~
than PhotoWCT~\cite{li2018closed}.
%

\vspace{-2mm}\subsubsection{Ablation study}\vspace{-2mm}
%
Note that we denote the use of feature aggregation module as \emph{FA},  the use of normalized skip connections as \emph{NS}, and \emph{MST-X} for the multi-stage style transfer module, where $\mathrm{X=3,5,\infty}$ refers to the incorporation of transfer modules (like WCT~\cite{li2017universal} or AdaIN~\cite{huang2017arbitrary}) in the first three stage of the decoder, all stages of the decoder, and all normalized skip connections, respectively.

\textbf{ArtNet.}~As was shown in Fig.~\ref{fig:ablation} Row 1, FA and MST-3 improve the stylization effects in succession.
However, as shown in Fig.~\ref{fig:ablation} (e), placing the transfer module in the  low-level stages of the decoder hurt the results of the stylization.%

\textbf{PhotoNet.}~We present the ablation study results of PhotoNet in  Fig.~\ref{fig:ablation} Row 2, which demonstrates the effectiveness of NS.
Moreover, Fig.~\ref{fig:ablation} (i) shows that conducting style transfer in  normalized skip connections can further improve the visual effects of the transferred images (red flowers and blue sky in the image are highlighted by MST-$\infty$.).

\begin{table}
 \caption{\textbf{Image reconstruction evaluation.} All the listed algorithms are evaluated on the $512\times512$ images.}
 \vspace{-3mm}
    \centering
    \scriptsize
    \begin{tabular}{lccccc}
        \toprule
        Method & AdaIN & WCT & PhotoWCT & \textbf{ArtNet} & \textbf{PhotoNet} \\
        \midrule
        Error & 86611.57 & 83196.87 & 76755.60 & \textbf{75021.59} & \textbf{74643.66} \\
        \bottomrule
    \end{tabular}
\label{tab:reconstruction}
       \vspace{-3mm}
\end{table}
\vspace{-2mm}\subsubsection{Image reconstruction}\vspace{-2mm}
One major finding of our work is that lower image reconstruction error of AEs leads to better stylization performance. We present the images generated by vanilla AEs used in AdaIN~\cite{huang2017arbitrary}, WCT~\cite{li2017universal}, PhotoWCT~\cite{li2018closed}, as well as ArtNet and PhotoNet in Fig.~\ref{fig:reconstruction}.
Images generated by AdaIN~\cite{huang2017arbitrary} contains significant artifacts in both two cases, e.g., changes in the lip color of the girl in the top row and twisted cables/steel frame of the bridge in the bottom row.
The WCT~\cite{li2017universal} method distorts images where the hairlines of the girl and cables of the bridge are blurred.
ArtNet renders the clearest reconstruction result among the compared artistic stylization methods.
As for photorealistic cases, PhotoNet outperforms PhotoWCT~\cite{li2018closed} in terms of the sharpness in details, e.g., the steel frame of the bridge, hairlines, and eyelash of the girl contain clearer details.
We quantitatively evaluate the performance of the proposed algorithms against the  state-of-the-art baseline methods by computing the mean squared error as defined by Eq.~\ref{eq:reconstruction} between the original and the reconstructed images on a randomly selected dataset.
\begin{equation}
    error = \sum\limits_{i=1}^N\left( \| I_{in} - I_{out} \|_F \right) / N,
    \label{eq:reconstruction}
\end{equation}
where $N$ denotes the number of the selected images and $N=13$ here.
As shown in Table~\ref{tab:reconstruction}, the AE used in ArtNet and PhotoNet achieve better image reconstruction performance compared to all other methods.

\section{Conclusions}
In this paper, we present two network architectures to address artistic and photorealistic style transfer, respectively. ArtNet outperforms the artistic stylization results of the existing methods by introducing a feature aggregation operation and a multi-stage stylization module, which also avoids the use of  multi-round computation for stylization to speed up the transfer process. In addition, PhotoNet utilizes normalized skip connections to preserve details of the transferred images, thus generating rich-detailed and well-stylized images. Our extensive experiments include visual comparisons, quantitative comparisons, and a thorough ablation study to show that the proposed approaches have the ability to remarkably improve the stylization effects for both artistic and photo-realistic stylization while reducing the time consumption dramatically especially for photo-realistic transfer algorithms. In the future, we will try to combine the proposed networks with  newly proposed style transfer modules such as Avatar-Net~\cite{sheng2018avatar} and the method by Gu~\etal~\cite{gu2018arbitrary} to further improve the style transfer results.
\balance

\clearpage
\appendix

\begin{center}
\huge{\textbf{Supplementary Material}}
\end{center}

\section{Network Training Setting}
We train the ArtNet and PhotoNet with the reconstruction and perceptual loss functions,
\begin{equation}
    \mathcal{L} = \alpha \cdot \mathcal{L}_{recon} + \left( 1 - \alpha \right) \cdot\mathcal{L}_{precep},
\end{equation}
where $\alpha$ is used to balance tow loss terms. We set $\alpha = 0.5$ during training. In addition, we use the Adam method~\cite{kingma2014adam} and set the learning rate to be $1e^{-4}$. We train both the ArtNet and PhotoNet for $5$ epoches with the fixed learning rate. The training process spends about eight hours on a NVIDIA Tesla P100 GPU with 16GB GPU RAM.

\section{User Control}
\begin{figure*}[tbp]
    \centering
    \subfloat[Content]{
    \begin{minipage}[t]{0.128\textwidth}
    \centering
    \includegraphics[width=\textwidth]{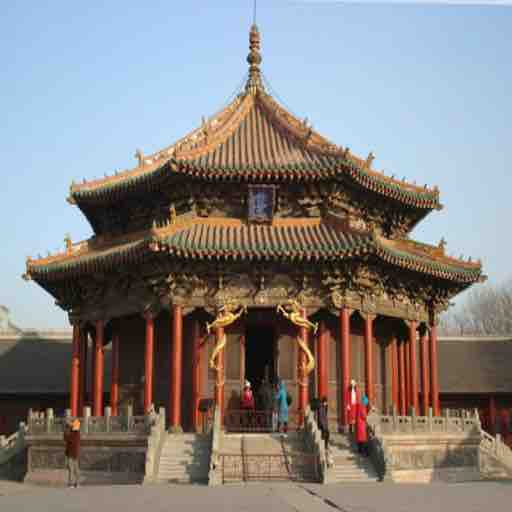}\\[0.5mm]
    \includegraphics[width=\textwidth]{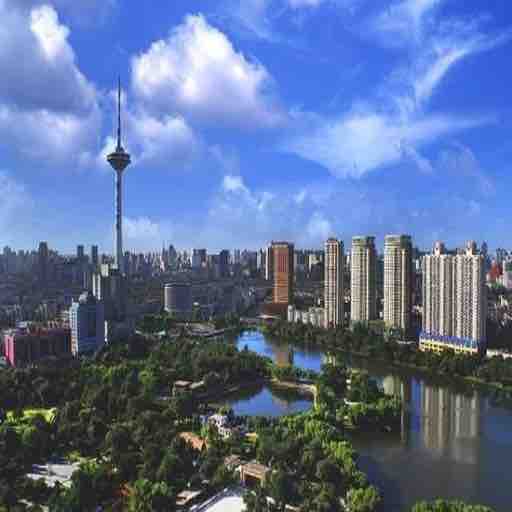}\\[0.5mm]
    \includegraphics[width=\textwidth]{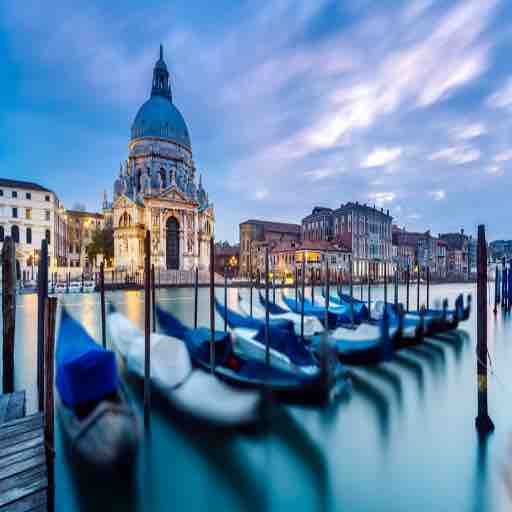}\\[0.5mm]
    \includegraphics[width=\textwidth]{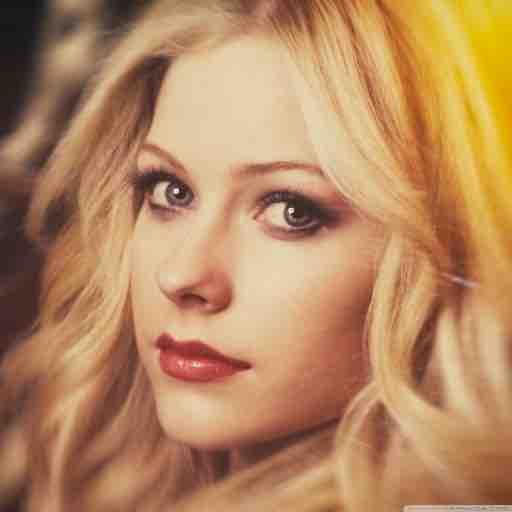}
    \end{minipage}
    }
    \hspace{-1.5mm}
    \subfloat[$\beta = 0.2$]{
    \begin{minipage}[t]{0.128\textwidth}
    \centering
    \includegraphics[width=\textwidth]{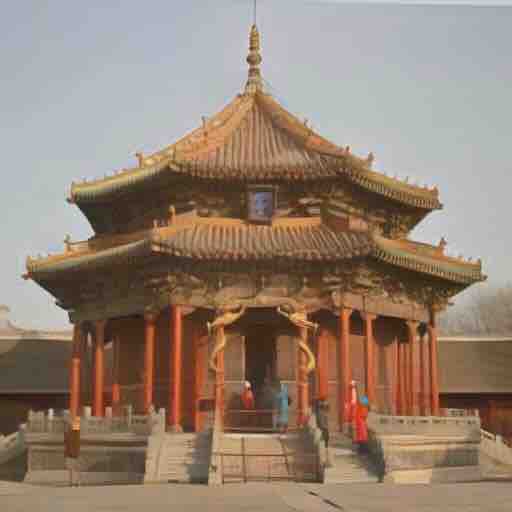}\\[0.5mm]
    \includegraphics[width=\textwidth]{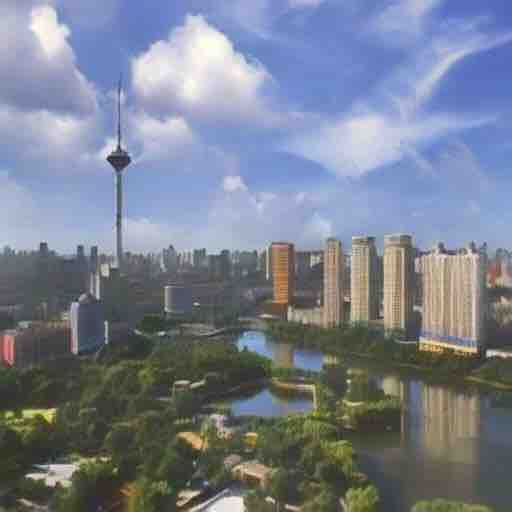}\\[0.5mm]
    \includegraphics[width=\textwidth]{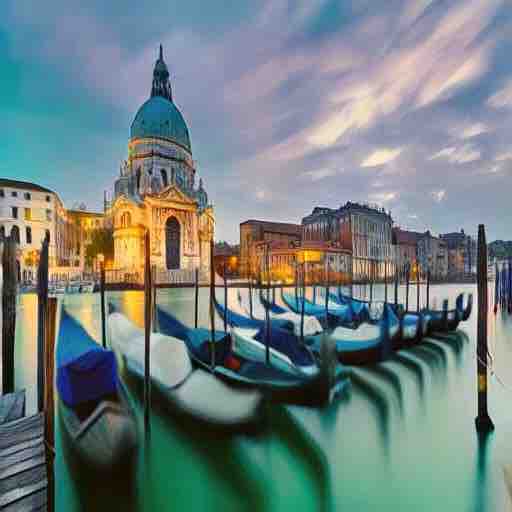}\\[0.5mm]
    \includegraphics[width=\textwidth]{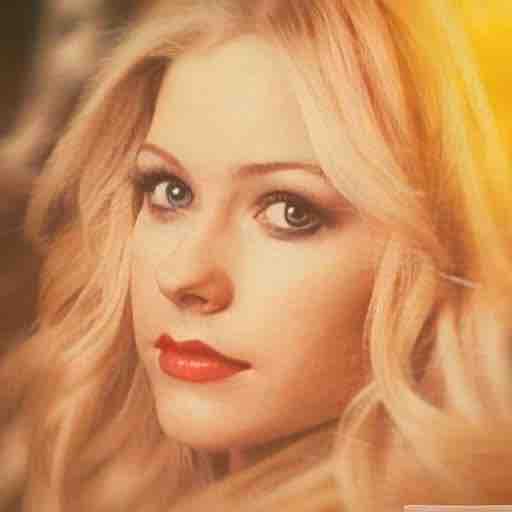}
    \end{minipage}
    }
    \hspace{-1.5mm}
    \subfloat[$\beta = 0.4$]{
    \begin{minipage}[t]{0.128\textwidth}
    \centering
    \includegraphics[width=\textwidth]{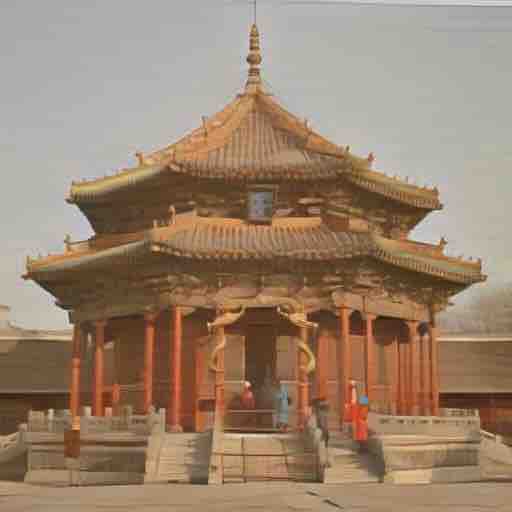}\\[0.5mm]
    \includegraphics[width=\textwidth]{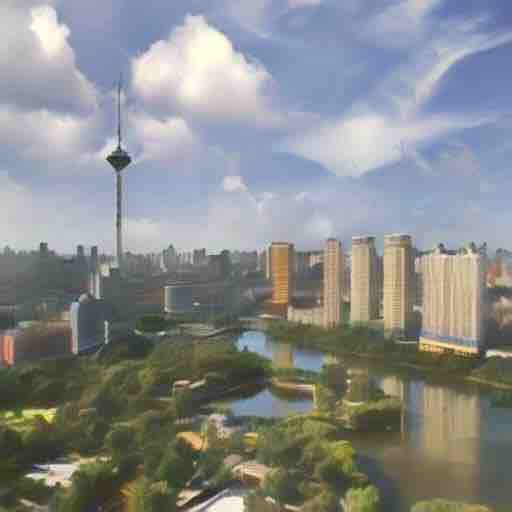}\\[0.5mm]
    \includegraphics[width=\textwidth]{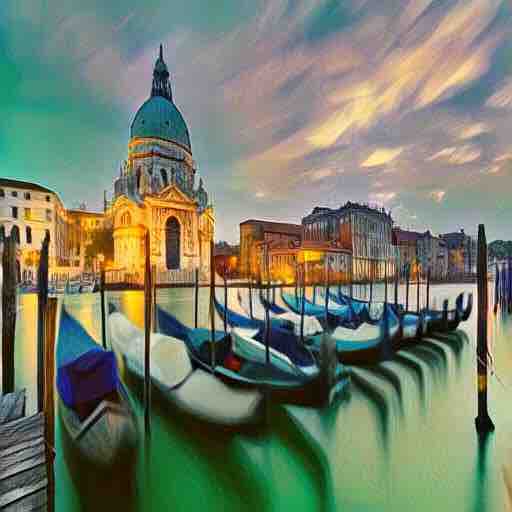}\\[0.5mm]
    \includegraphics[width=\textwidth]{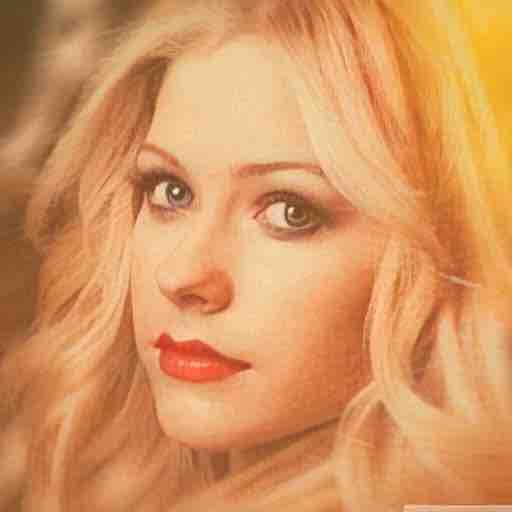}
    \end{minipage}
    }
    \hspace{-1.5mm}
    \subfloat[$\beta = 0.6$]{
    \begin{minipage}[t]{0.128\textwidth}
    \centering
    \includegraphics[width=\textwidth]{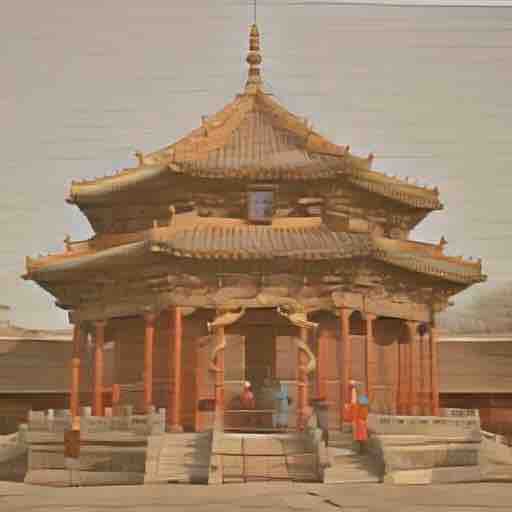}\\[0.5mm]
    \includegraphics[width=\textwidth]{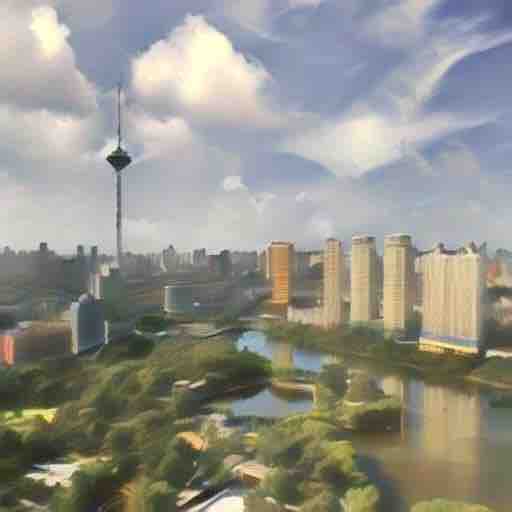}\\[0.5mm]
    \includegraphics[width=\textwidth]{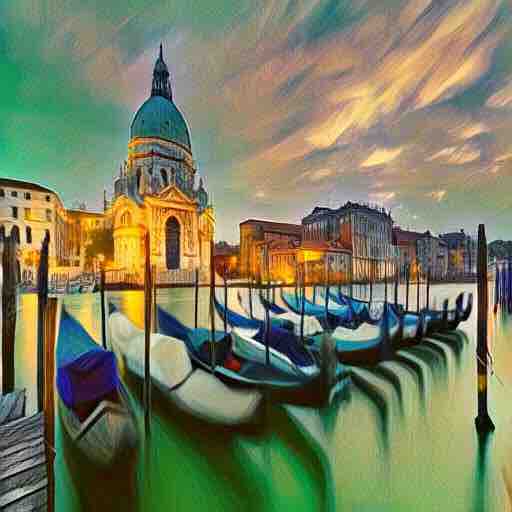}\\[0.5mm]
    \includegraphics[width=\textwidth]{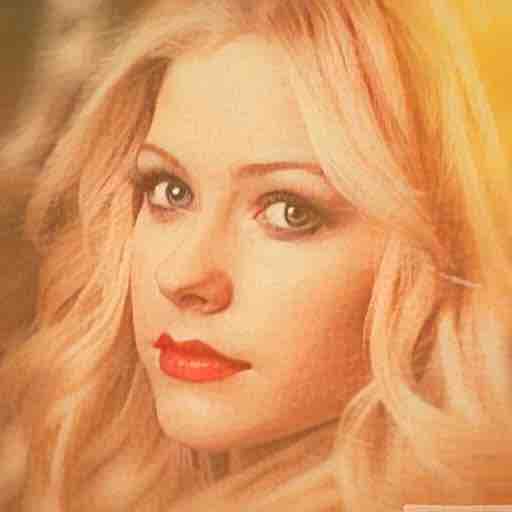}
    \end{minipage}
    }
    \hspace{-1.5mm}
    \subfloat[$\beta = 0.8$]{
    \begin{minipage}[t]{0.128\textwidth}
    \centering
    \includegraphics[width=\textwidth]{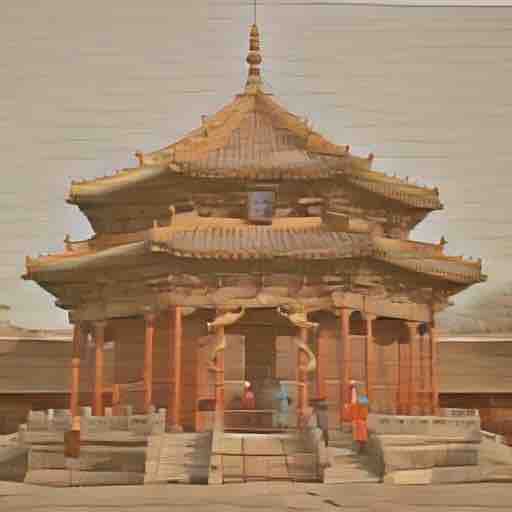}\\[0.5mm]
    \includegraphics[width=\textwidth]{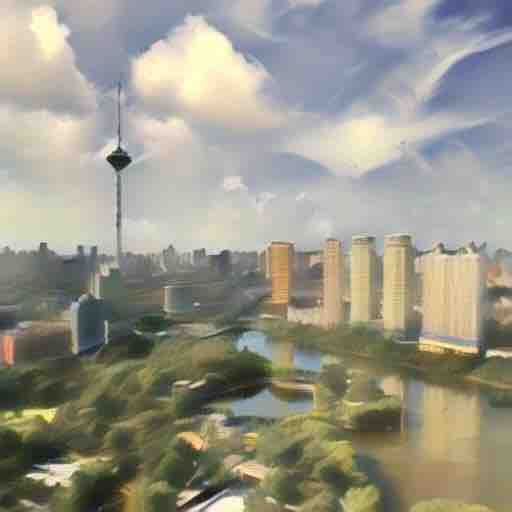}\\[0.5mm]
    \includegraphics[width=\textwidth]{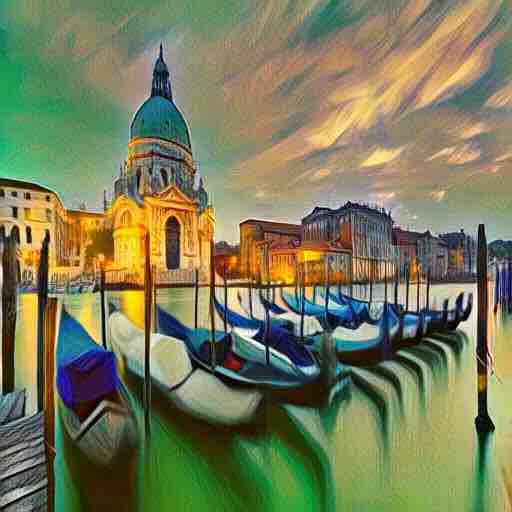}\\[0.5mm]
    \includegraphics[width=\textwidth]{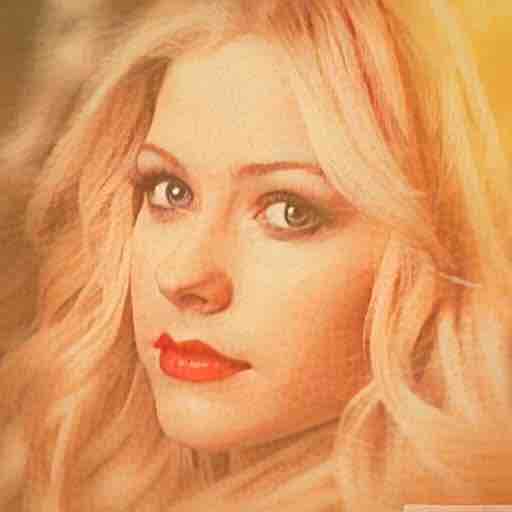}
    \end{minipage}
    }
    \hspace{-1.5mm}
    \subfloat[$\beta = 1.0$]{
    \begin{minipage}[t]{0.128\textwidth}
    \centering
    \includegraphics[width=\textwidth]{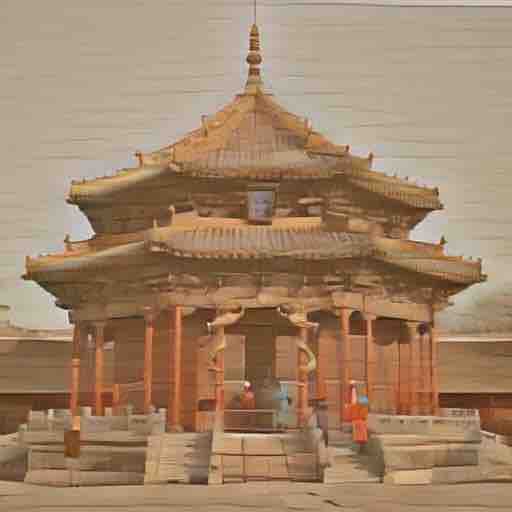}\\[0.5mm]
    \includegraphics[width=\textwidth]{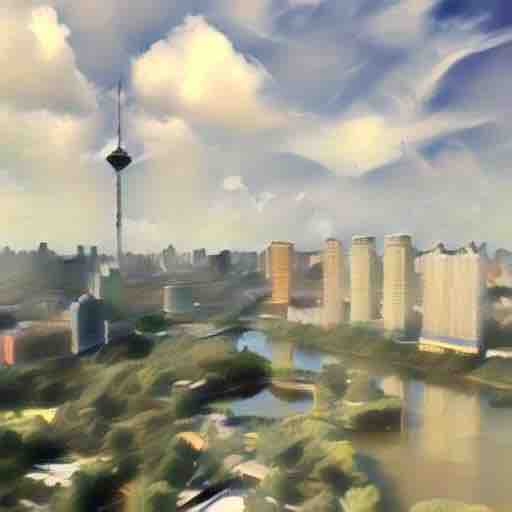}\\[0.5mm]
    \includegraphics[width=\textwidth]{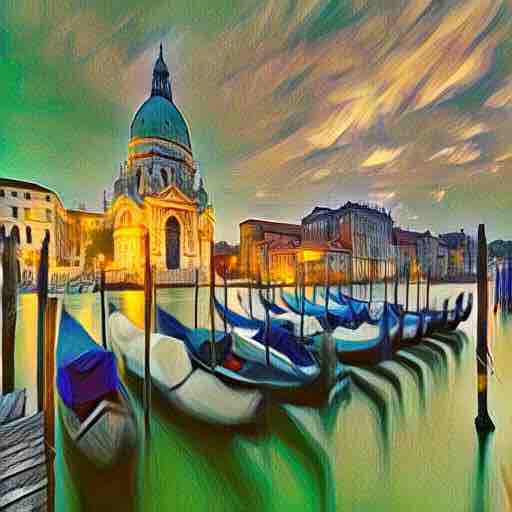}\\[0.5mm]
    \includegraphics[width=\textwidth]{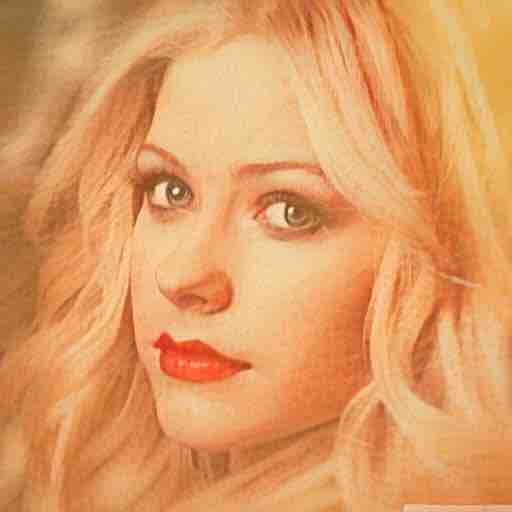}
    \end{minipage}
    }
    \hspace{-1.5mm}
    \subfloat[Style]{
    \begin{minipage}[t]{0.128\textwidth}
    \centering
    \includegraphics[width=\textwidth]{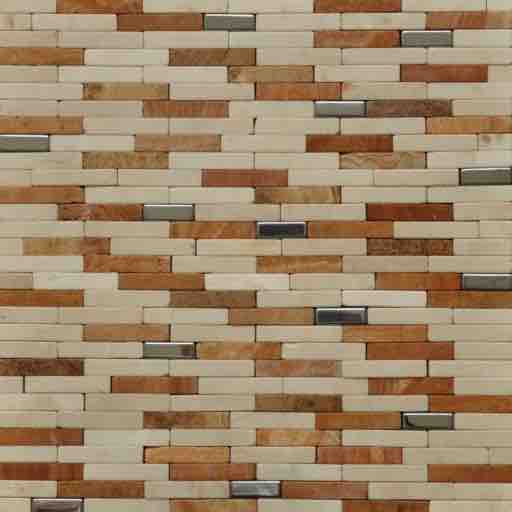}\\[0.5mm]
    \includegraphics[width=\textwidth]{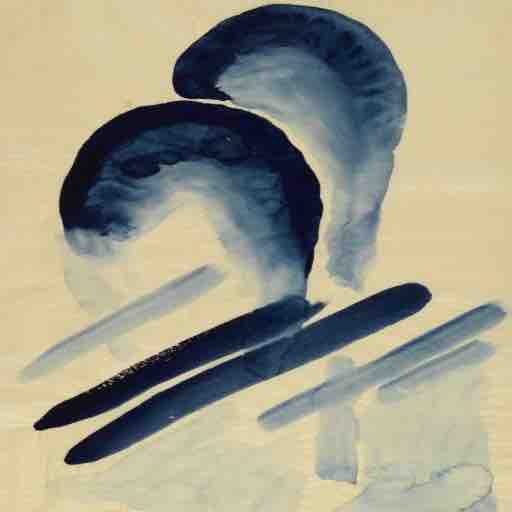}\\[0.5mm]
    \includegraphics[width=\textwidth]{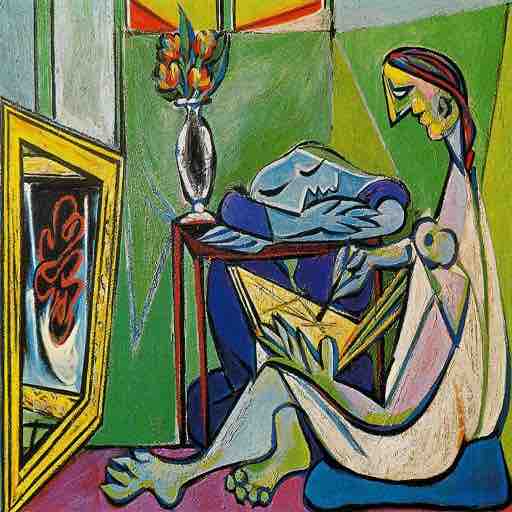}\\[0.5mm]
    \includegraphics[width=\textwidth]{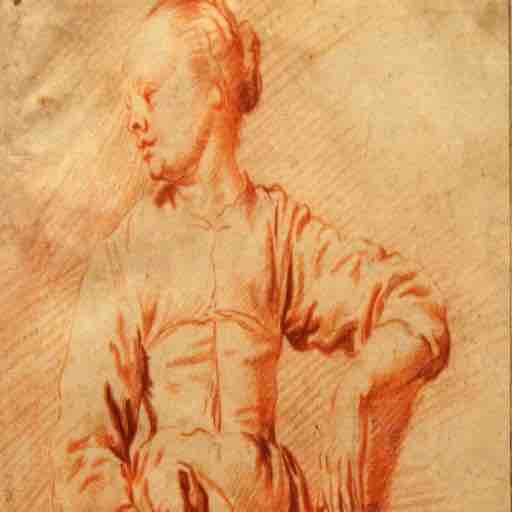}
    \end{minipage}
    }
    \caption{\textbf{Artistic style transfer results by the ArtNet(AdaIN) with control factor $\beta$ ranging from $0.2$ to $1.0$.}}
    \label{fig:art_adain_user_control}
\end{figure*}

\begin{figure*}[tbp]
    \centering
    \subfloat[Content]{
    \begin{minipage}[t]{0.128\textwidth}
    \centering
    \includegraphics[width=\textwidth]{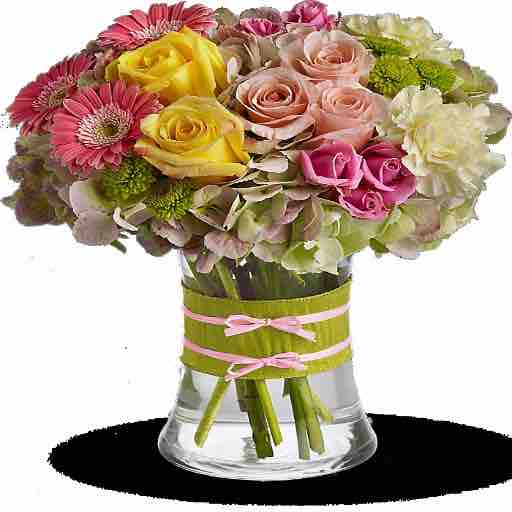}\\[0.5mm]
    \includegraphics[width=\textwidth]{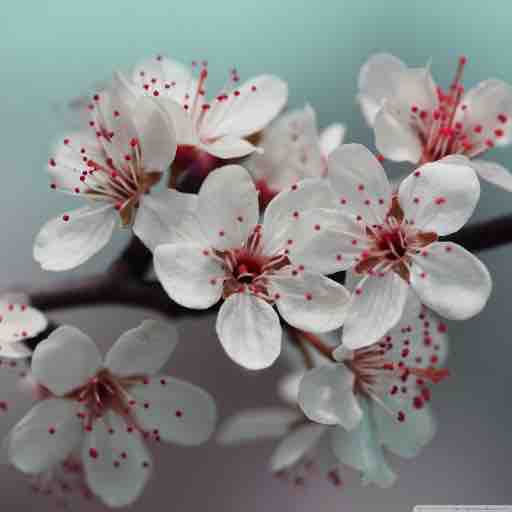}\\[0.5mm]
    \includegraphics[width=\textwidth]{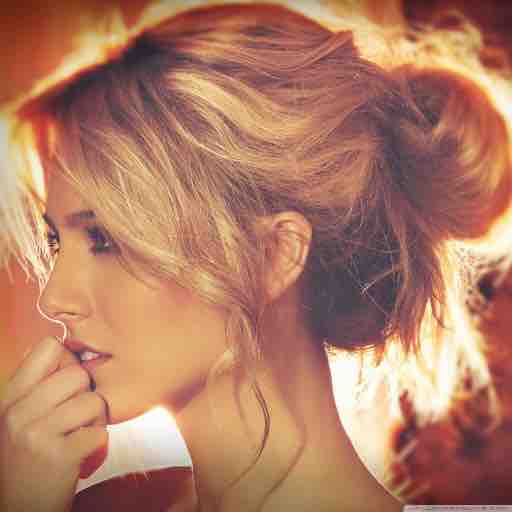}\\[0.5mm]
    \includegraphics[width=\textwidth]{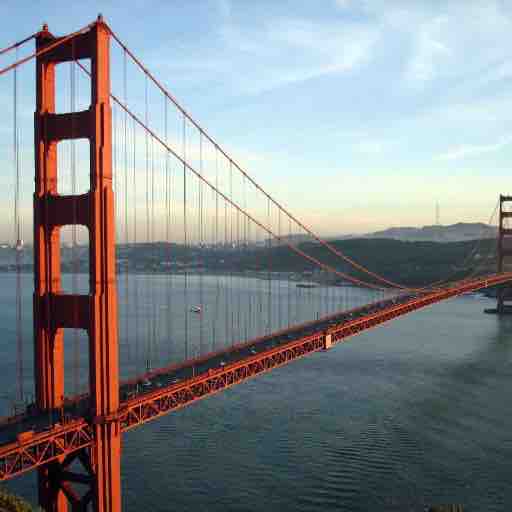}
    \end{minipage}
    }
    \hspace{-1.5mm}
    \subfloat[$\beta = 0.2$]{
    \begin{minipage}[t]{0.128\textwidth}
    \centering
    \includegraphics[width=\textwidth]{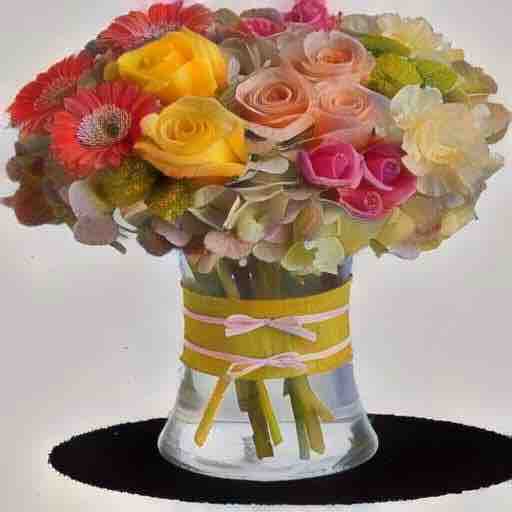}\\[0.5mm]
    \includegraphics[width=\textwidth]{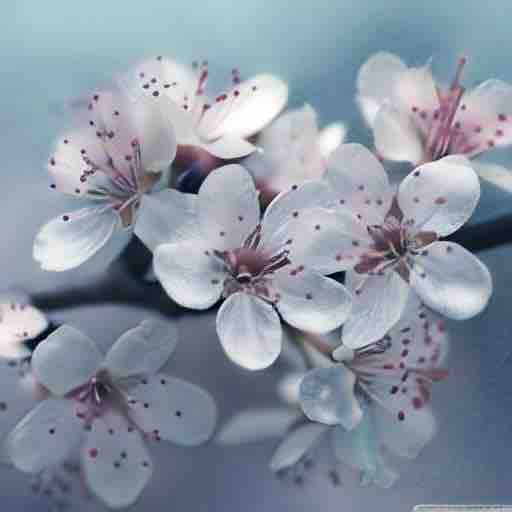}\\[0.5mm]
    \includegraphics[width=\textwidth]{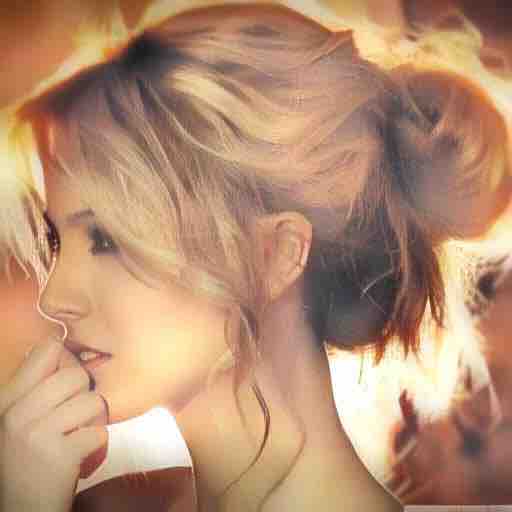}\\[0.5mm]
    \includegraphics[width=\textwidth]{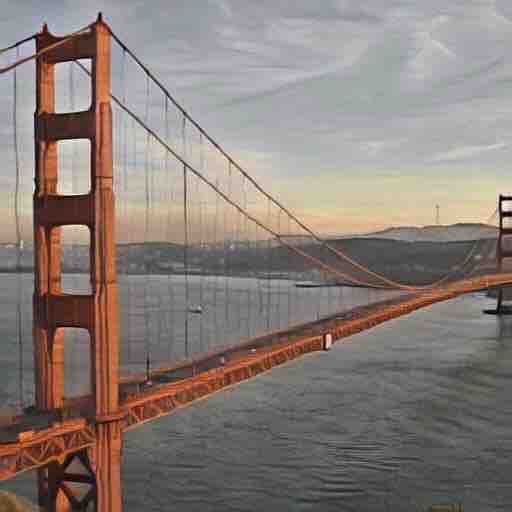}
    \end{minipage}
    }
    \hspace{-1.5mm}
    \subfloat[$\beta = 0.4$]{
    \begin{minipage}[t]{0.128\textwidth}
    \centering
    \includegraphics[width=\textwidth]{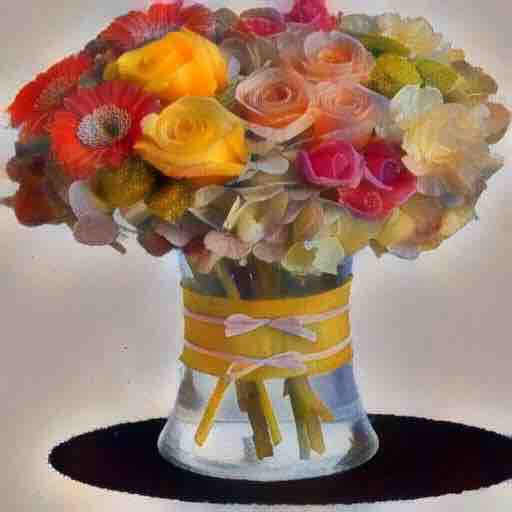}\\[0.5mm]
    \includegraphics[width=\textwidth]{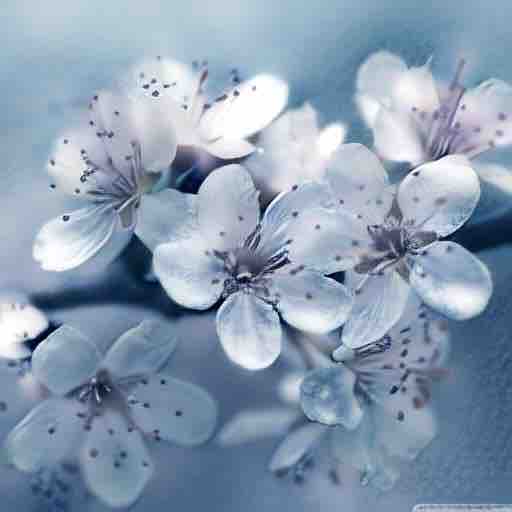}\\[0.5mm]
    \includegraphics[width=\textwidth]{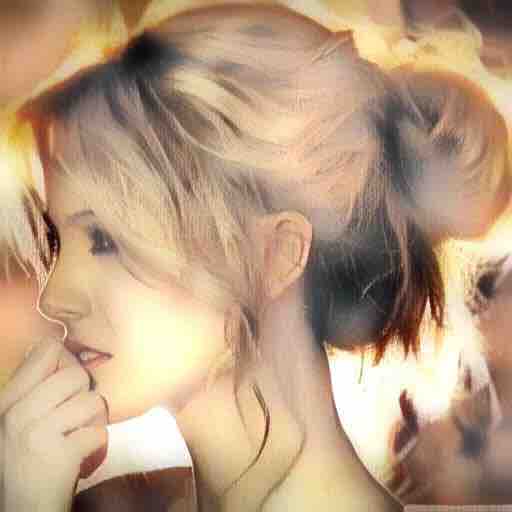}\\[0.5mm]
    \includegraphics[width=\textwidth]{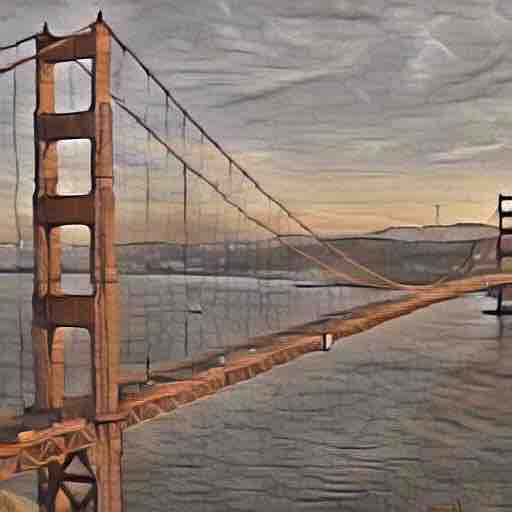}
    \end{minipage}
    }
    \hspace{-1.5mm}
    \subfloat[$\beta = 0.6$]{
    \begin{minipage}[t]{0.128\textwidth}
    \centering
    \includegraphics[width=\textwidth]{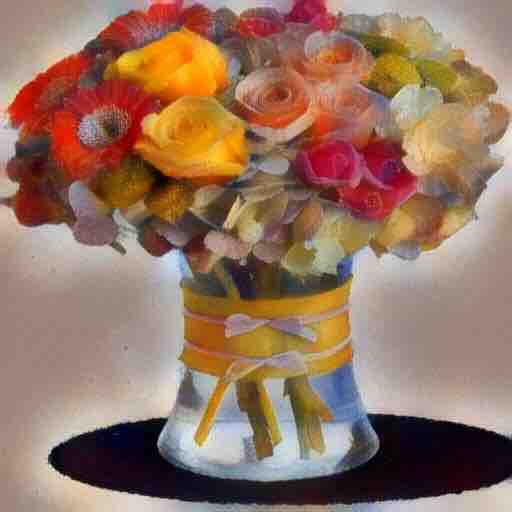}\\[0.5mm]
    \includegraphics[width=\textwidth]{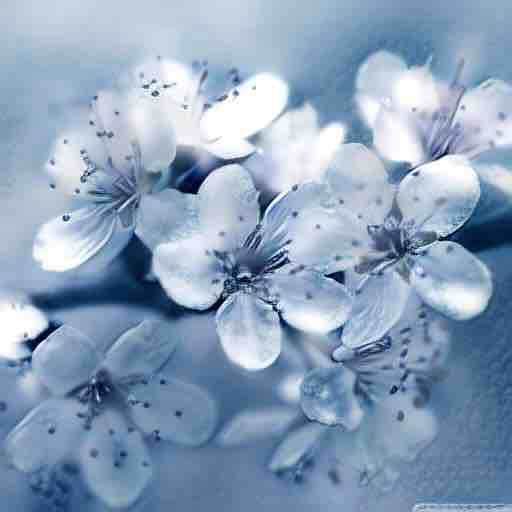}\\[0.5mm]
    \includegraphics[width=\textwidth]{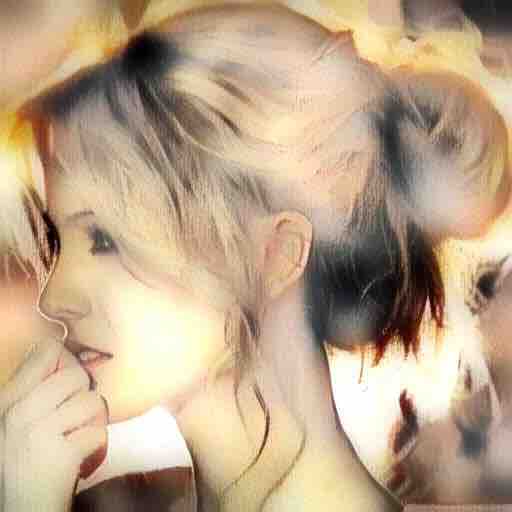}\\[0.5mm]
    \includegraphics[width=\textwidth]{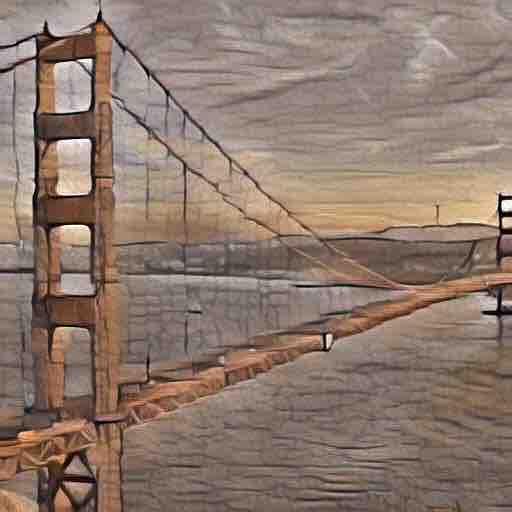}
    \end{minipage}
    }
    \hspace{-1.5mm}
    \subfloat[$\beta = 0.8$]{
    \begin{minipage}[t]{0.128\textwidth}
    \centering
    \includegraphics[width=\textwidth]{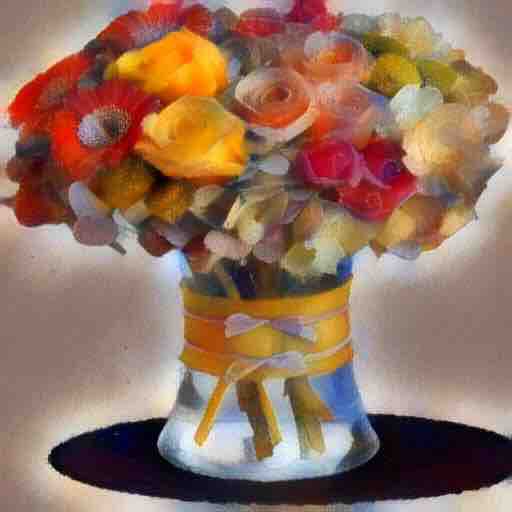}\\[0.5mm]
    \includegraphics[width=\textwidth]{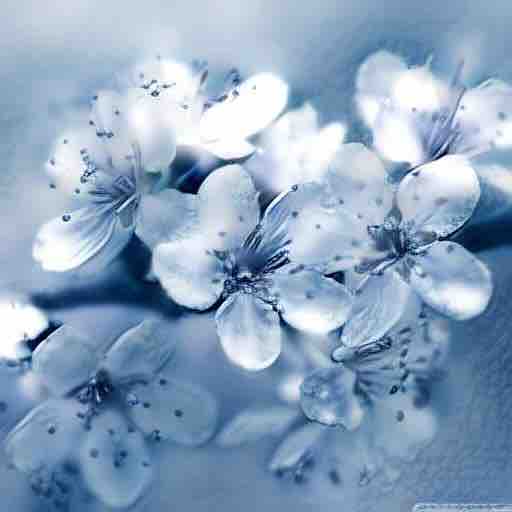}\\[0.5mm]
    \includegraphics[width=\textwidth]{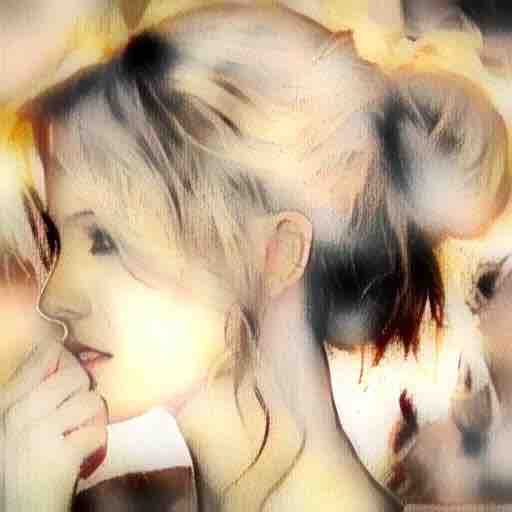}\\[0.5mm]
    \includegraphics[width=\textwidth]{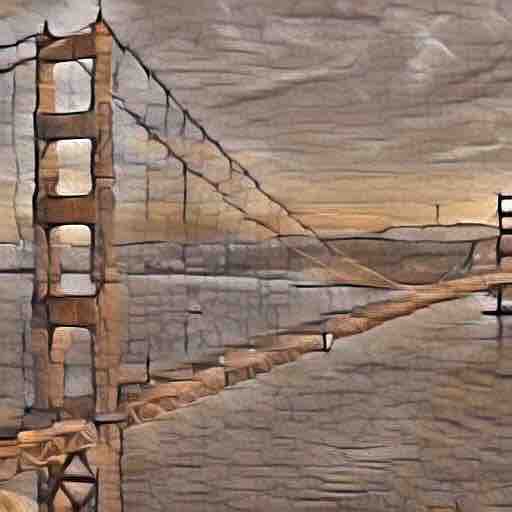}
    \end{minipage}
    }
    \hspace{-1.5mm}
    \subfloat[$\beta = 1.0$]{
    \begin{minipage}[t]{0.128\textwidth}
    \centering
    \includegraphics[width=\textwidth]{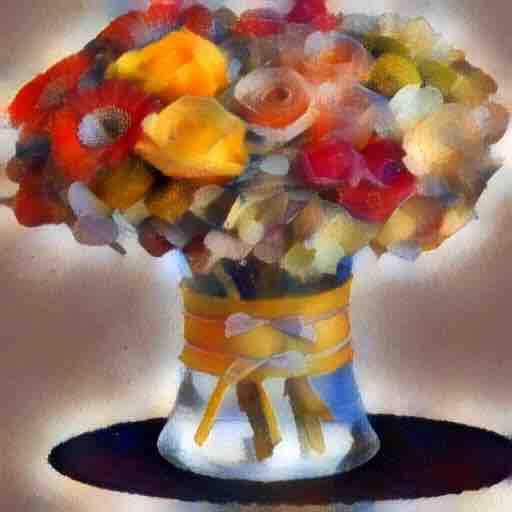}\\[0.5mm]
    \includegraphics[width=\textwidth]{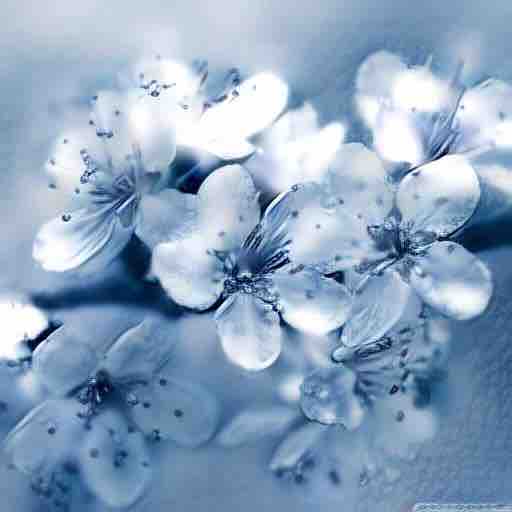}\\[0.5mm]
    \includegraphics[width=\textwidth]{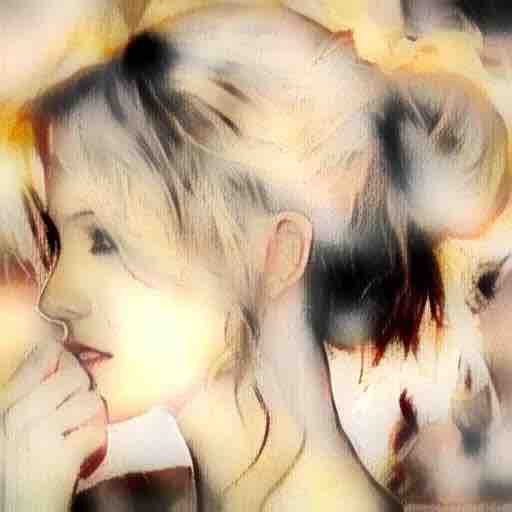}\\[0.5mm]
    \includegraphics[width=\textwidth]{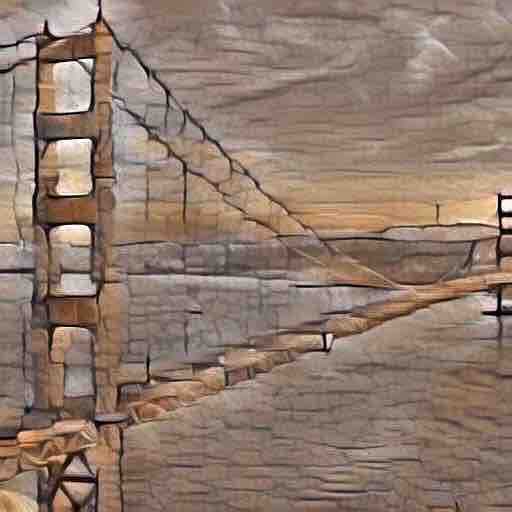}
    \end{minipage}
    }
    \hspace{-1.5mm}
    \subfloat[Style]{
    \begin{minipage}[t]{0.128\textwidth}
    \centering
    \includegraphics[width=\textwidth]{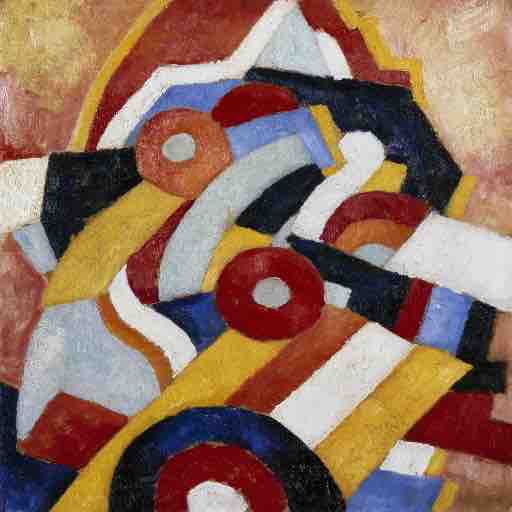}\\[0.5mm]
    \includegraphics[width=\textwidth]{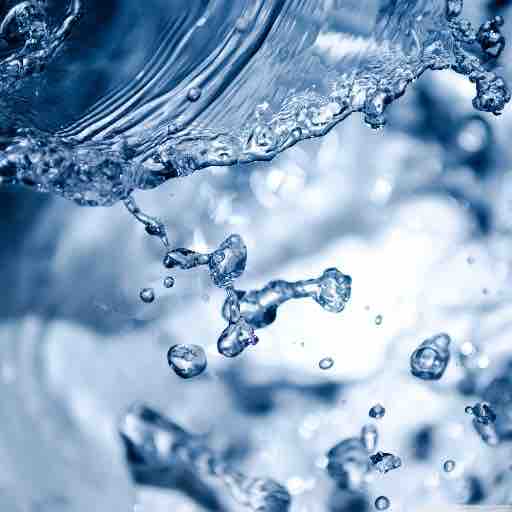}\\[0.5mm]
    \includegraphics[width=\textwidth]{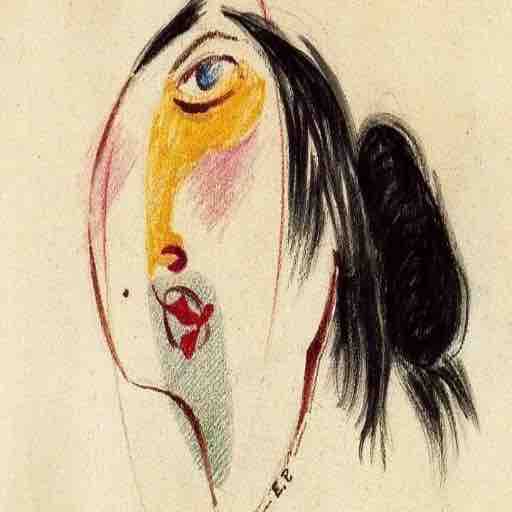}\\[0.5mm]
    \includegraphics[width=\textwidth]{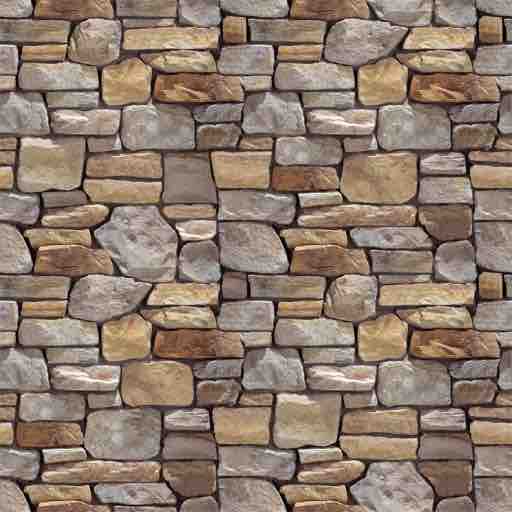}
    \end{minipage}
    }
    \caption{\textbf{Artistic style transfer results by the ArtNet(WCT) with control factor $\beta$ ranging from $0.2$ to $1.0$.}}
    \label{fig:art_wct_user_control}
\end{figure*}

\begin{figure*}[tbp]
    \centering
    \subfloat[Content]{
    \begin{minipage}[t]{0.128\textwidth}
    \centering
    \includegraphics[width=\textwidth]{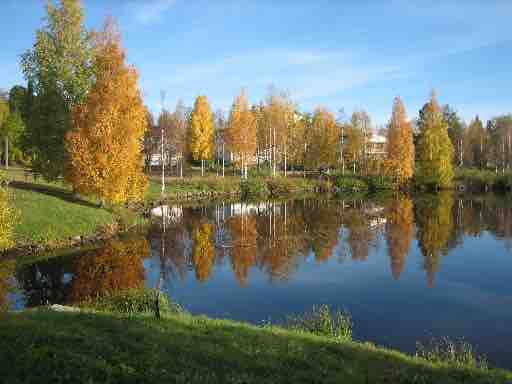}\\[0.5mm]
    \includegraphics[width=\textwidth]{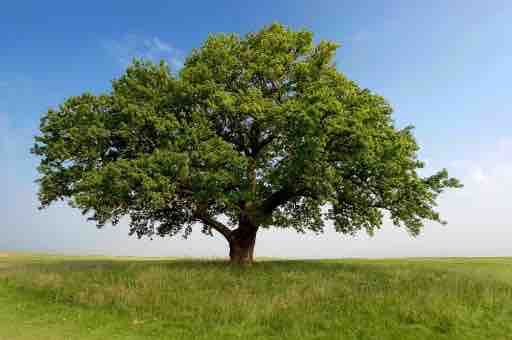}\\[0.5mm]
    \includegraphics[width=\textwidth]{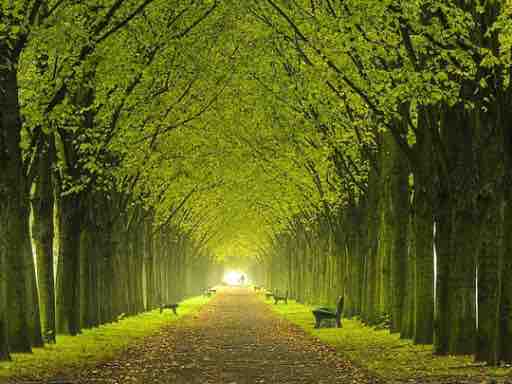}\\[0.5mm]
    \includegraphics[width=\textwidth]{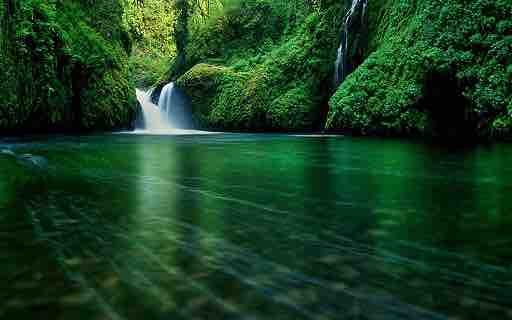}
    \end{minipage}
    }
    \hspace{-1.5mm}
    \subfloat[$\beta = 0.2$]{
    \begin{minipage}[t]{0.128\textwidth}
    \centering
    \includegraphics[width=\textwidth]{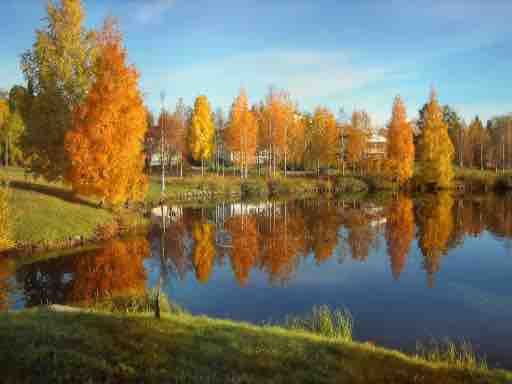}\\[0.5mm]
    \includegraphics[width=\textwidth]{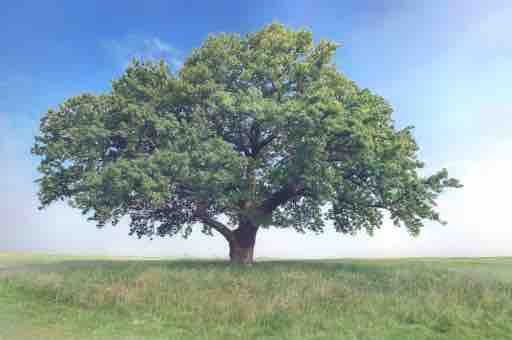}\\[0.5mm]
    \includegraphics[width=\textwidth]{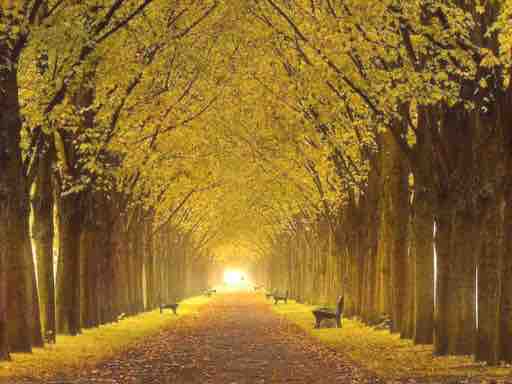}\\[0.5mm]
    \includegraphics[width=\textwidth]{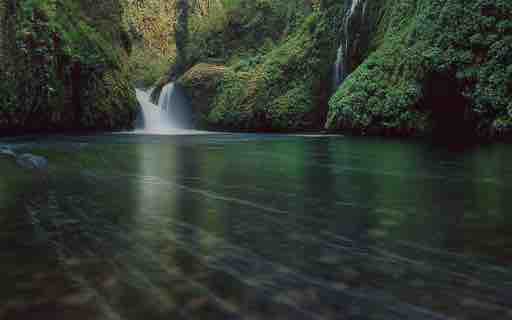}
    \end{minipage}
    }
    \hspace{-1.5mm}
    \subfloat[$\beta = 0.4$]{
    \begin{minipage}[t]{0.128\textwidth}
    \centering
    \includegraphics[width=\textwidth]{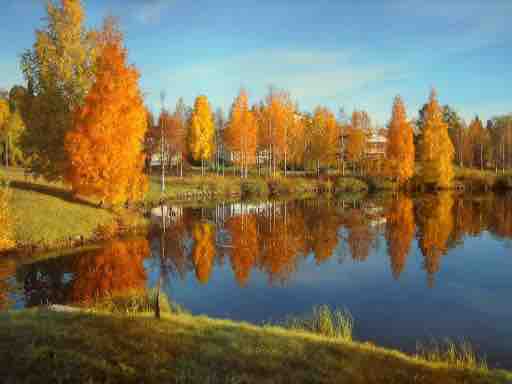}\\[0.5mm]
    \includegraphics[width=\textwidth]{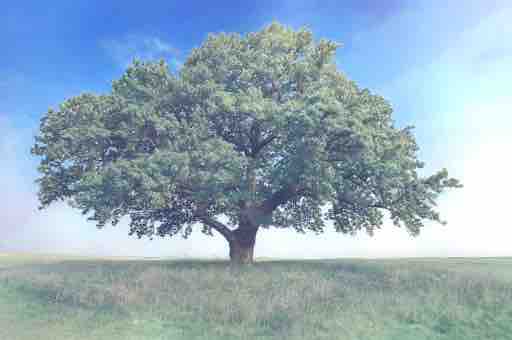}\\[0.5mm]
    \includegraphics[width=\textwidth]{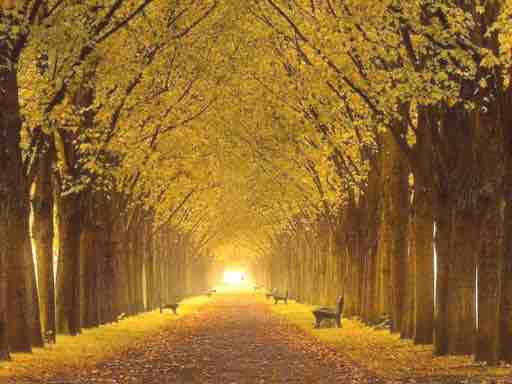}\\[0.5mm]
    \includegraphics[width=\textwidth]{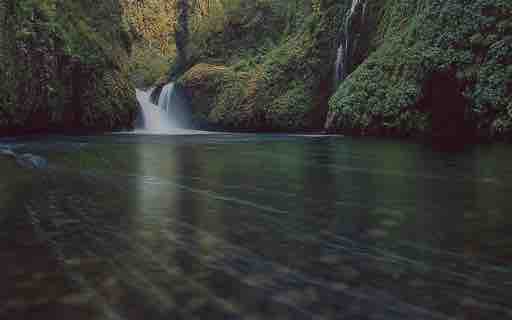}
    \end{minipage}
    }
    \hspace{-1.5mm}
    \subfloat[$\beta = 0.6$]{
    \begin{minipage}[t]{0.128\textwidth}
    \centering
    \includegraphics[width=\textwidth]{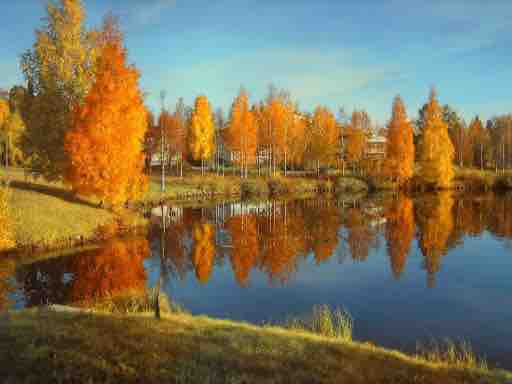}\\[0.5mm]
    \includegraphics[width=\textwidth]{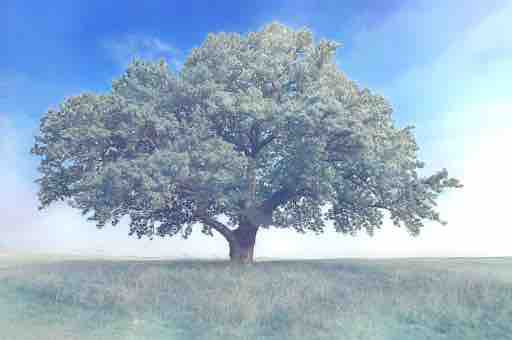}\\[0.5mm]
    \includegraphics[width=\textwidth]{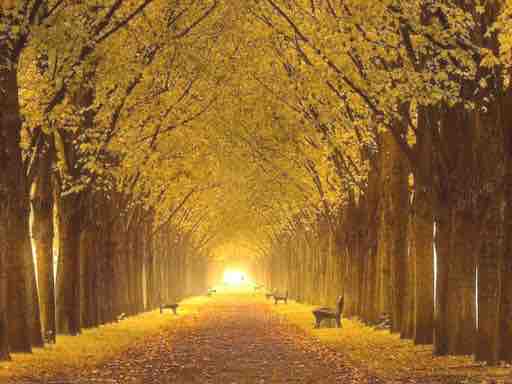}\\[0.5mm]
    \includegraphics[width=\textwidth]{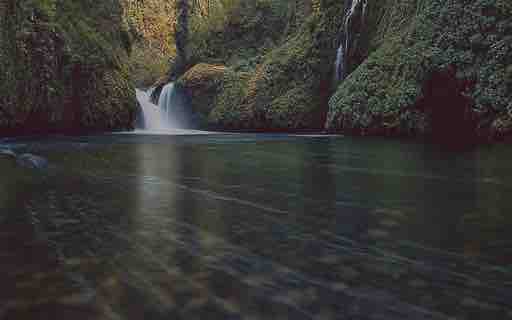}
    \end{minipage}
    }
    \hspace{-1.5mm}
    \subfloat[$\beta = 0.8$]{
    \begin{minipage}[t]{0.128\textwidth}
    \centering
    \includegraphics[width=\textwidth]{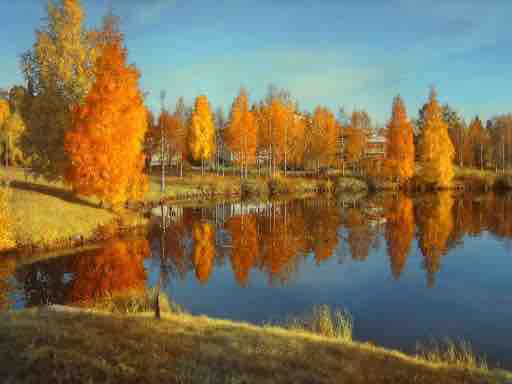}\\[0.5mm]
    \includegraphics[width=\textwidth]{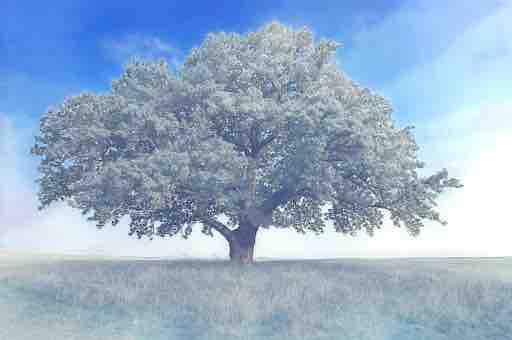}\\[0.5mm]
    \includegraphics[width=\textwidth]{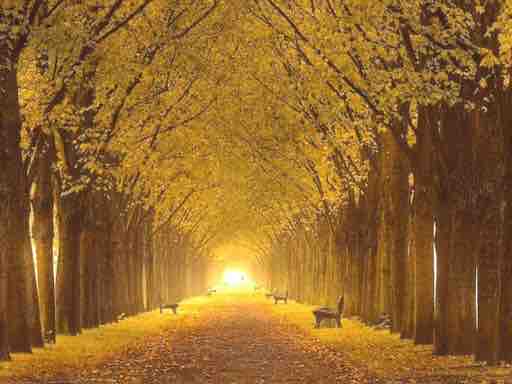}\\[0.5mm]
    \includegraphics[width=\textwidth]{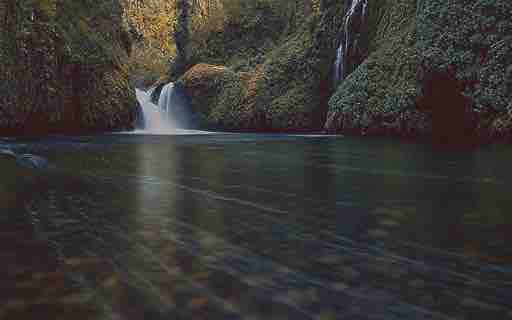}
    \end{minipage}
    }
    \hspace{-1.5mm}
    \subfloat[$\beta = 1.0$]{
    \begin{minipage}[t]{0.128\textwidth}
    \centering
    \includegraphics[width=\textwidth]{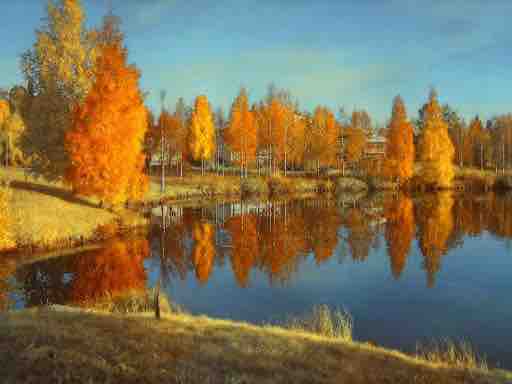}\\[0.5mm]
    \includegraphics[width=\textwidth]{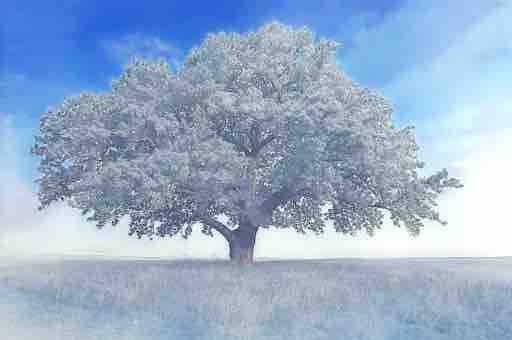}\\[0.5mm]
    \includegraphics[width=\textwidth]{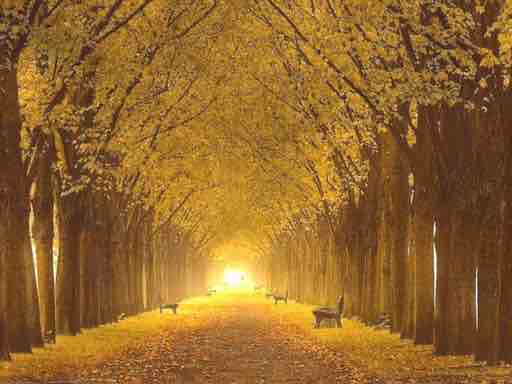}\\[0.5mm]
    \includegraphics[width=\textwidth]{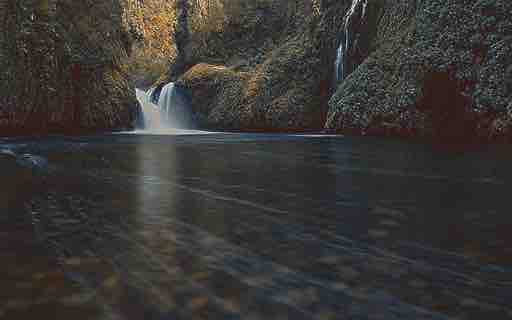}
    \end{minipage}
    }
    \hspace{-1.5mm}
    \subfloat[Style]{
    \begin{minipage}[t]{0.128\textwidth}
    \centering
    \includegraphics[width=\textwidth]{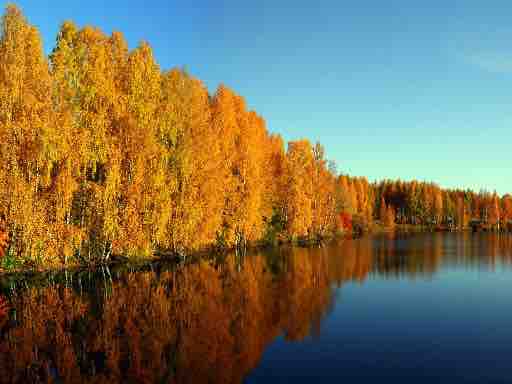}\\[0.5mm]
    \includegraphics[width=\textwidth]{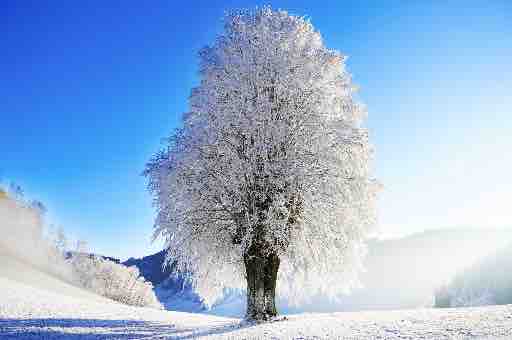}\\[0.5mm]
    \includegraphics[width=\textwidth]{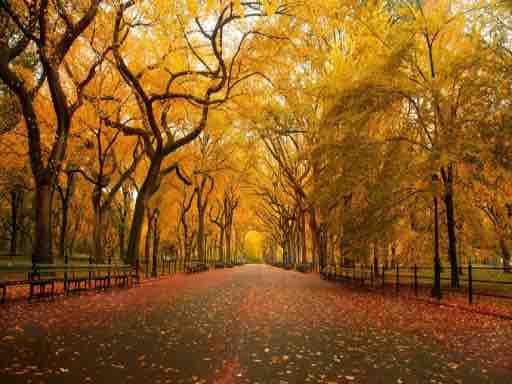}\\[0.5mm]
    \includegraphics[width=\textwidth]{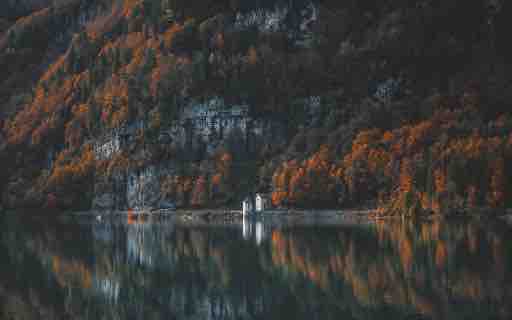}
    \end{minipage}
    }
    \caption{\textbf{Photorealistic style transfer results by the PhotoNet(WCT) with control factor $\beta$ ranging from $0.2$ to $1.0$.}}
    \label{fig:photo_user_control}
\end{figure*}

We conduct extensive experiments to demonstrate the ability of the proposed ArtNet and PhotoNet that enable flexible user control of the stylization effects as~\cite{li2018closed, huang2017arbitrary, li2017universal} do.
We introduce an user control module at the end of each style transfer module, which mixes the transferred and the content features as,
\begin{equation}
    \mathcal{F}_{out} = \beta \cdot \mathcal{F}_{transferred} + \left( 1 - \beta \right) \cdot \mathcal{F}_{content},
\end{equation}
where $\mathcal{F}$ represents deep feature maps, $\beta$ is a factor to let the user to control the degree of stylization effects. We present the artistic style transfer results with $\beta$ ranging from $0.2$ to $1.0$ in Figs.~\ref{fig:art_adain_user_control}\ref{fig:art_wct_user_control} and the photorealistic stylization results with the same user control setting in Fig.~\ref{fig:photo_user_control}. The generated images demonstrate that the proposed ArtNet and PhotoNet achieve incremental style transfer effects with changing control factors, which facilitates the flexible user control to the degree of the proposed style transfer algorithms.

\section{Style Transfer Results}
We present more style transfer results of the proposed ArtNet and PhotoNet in comparison to the AdaIN~\cite{huang2017arbitrary}, WCT~\cite{li2017universal}, and PhotoWCT~\cite{li2018closed} respectively.
\begin{figure*}[h]
    \centering
    \subfloat[Content]{
    \begin{minipage}[t]{0.208\textwidth}
    \centering
    \includegraphics[width=\textwidth]{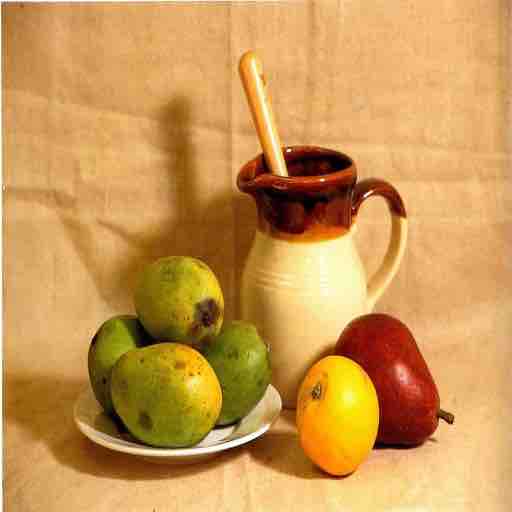}\\[0.5mm]
    \includegraphics[width=\textwidth]{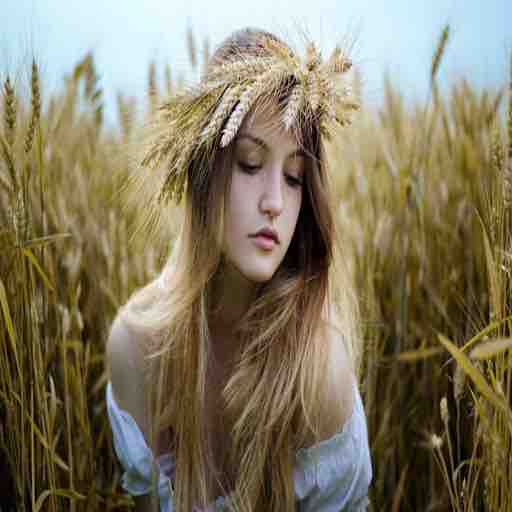}\\[0.5mm]
    \includegraphics[width=\textwidth]{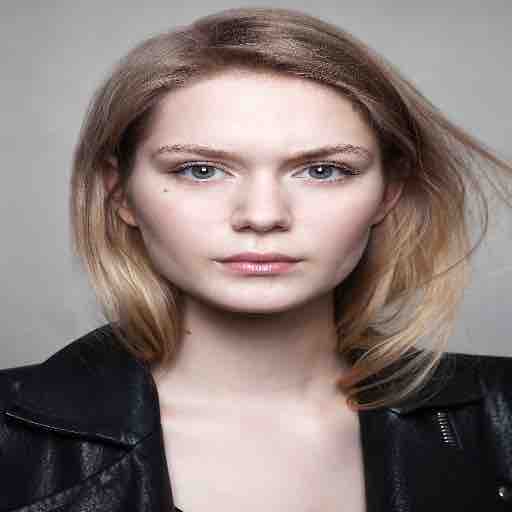}\\[0.5mm]
    \includegraphics[width=\textwidth]{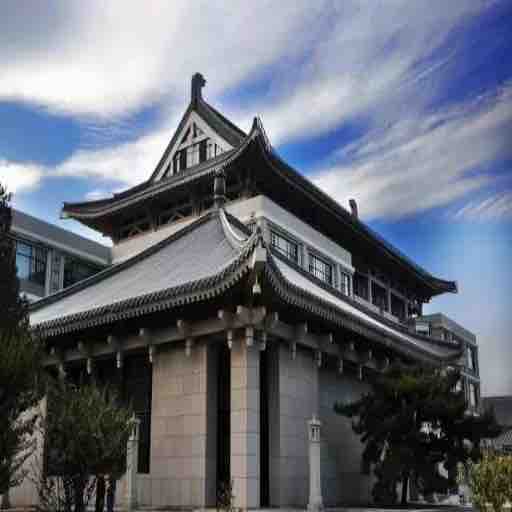}\\[0.5mm]
    \includegraphics[width=\textwidth]{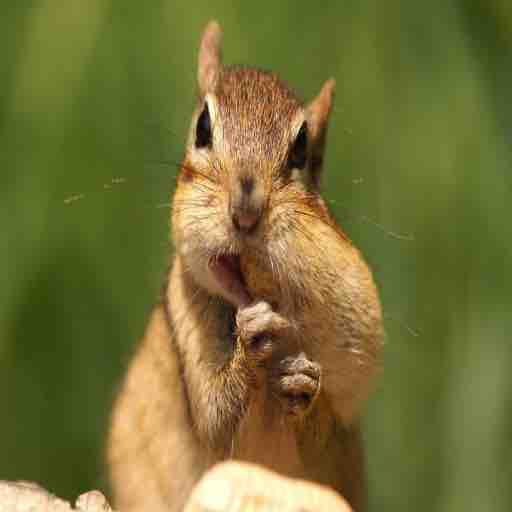}
    \end{minipage}
    }
    \hspace{-1.5mm}
    \subfloat[Style]{
    \begin{minipage}[t]{0.208\textwidth}
    \centering
    \includegraphics[width=\textwidth]{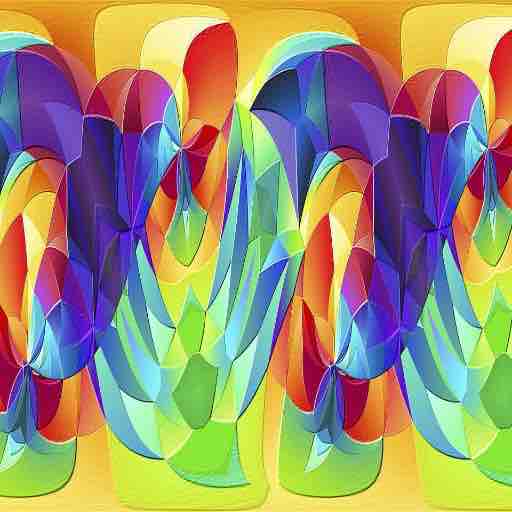}\\[0.5mm]
    \includegraphics[width=\textwidth]{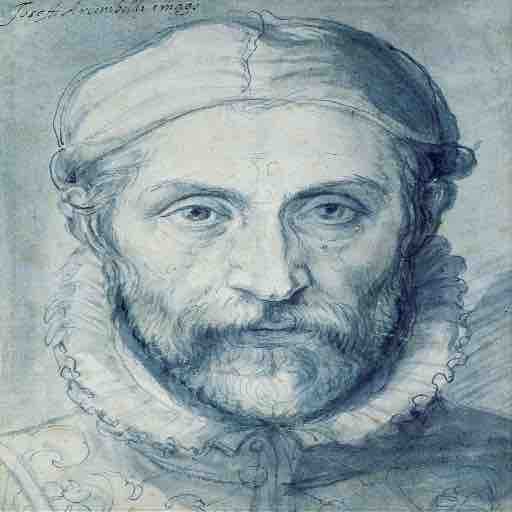}\\[0.5mm]
    \includegraphics[width=\textwidth]{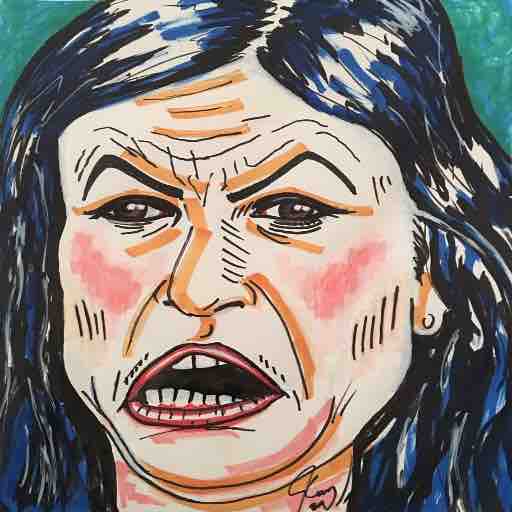}\\[0.5mm]
    \includegraphics[width=\textwidth]{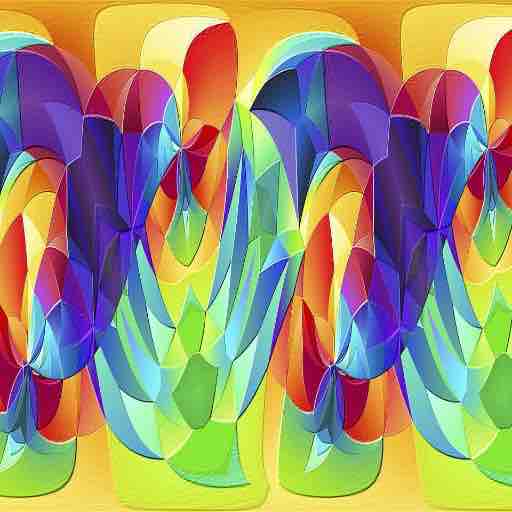}\\[0.5mm]
    \includegraphics[width=\textwidth]{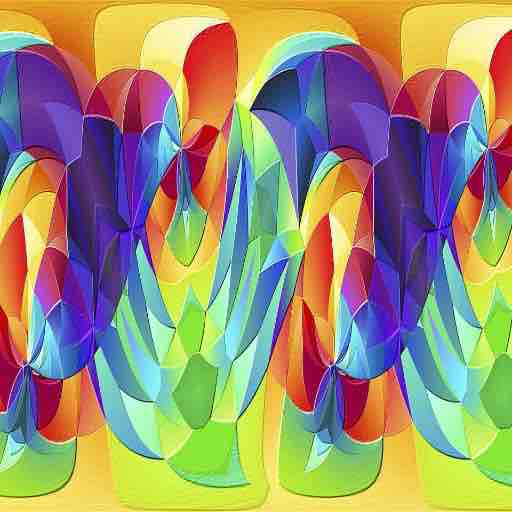}
    \end{minipage}
    }
    \hspace{-1.5mm}
    \subfloat[AdaIN~\cite{huang2017arbitrary}]{
    \begin{minipage}[t]{0.208\textwidth}
    \centering
    \includegraphics[width=\textwidth]{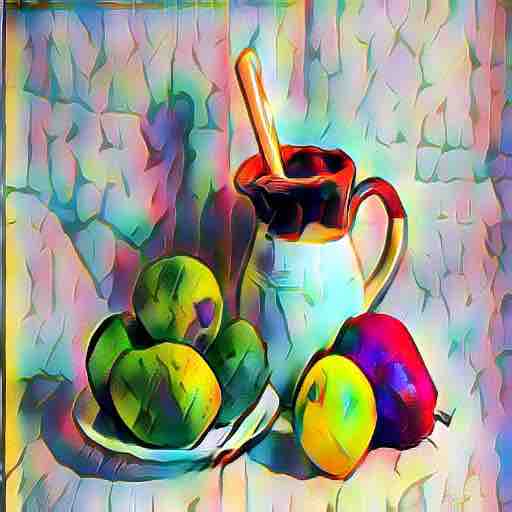}\\[0.5mm]
    \includegraphics[width=\textwidth]{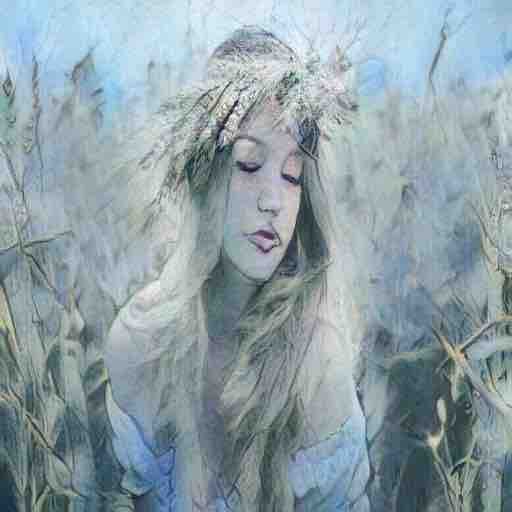}\\[0.5mm]
    \includegraphics[width=\textwidth]{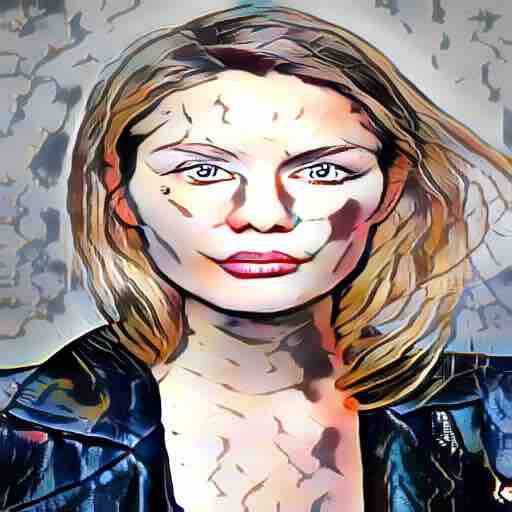}\\[0.5mm]
    \includegraphics[width=\textwidth]{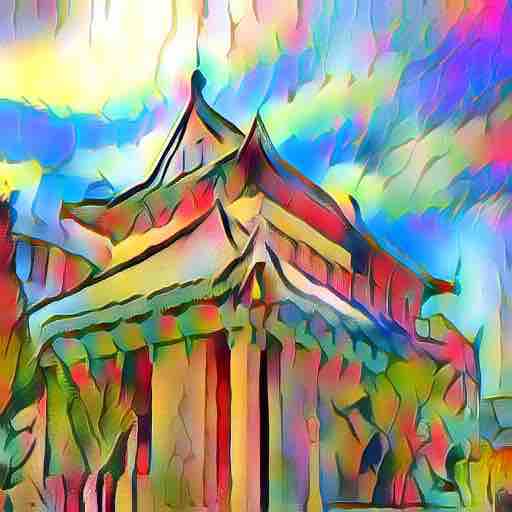}\\[0.5mm]
    \includegraphics[width=\textwidth]{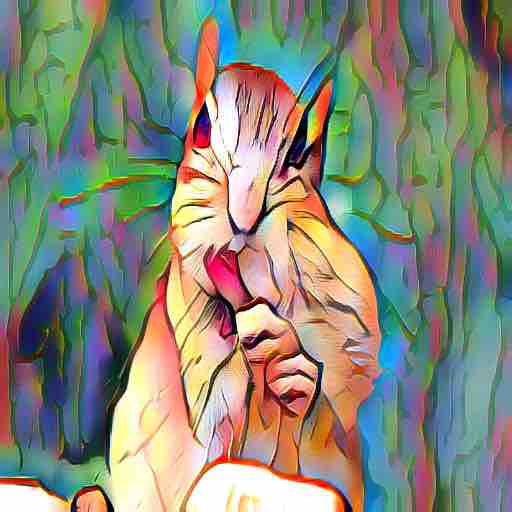}
    \end{minipage}
    }
    \hspace{-1.5mm}
    \subfloat[ArtNet(AdaIN)]{
    \begin{minipage}[t]{0.208\textwidth}
    \centering
    \includegraphics[width=\textwidth]{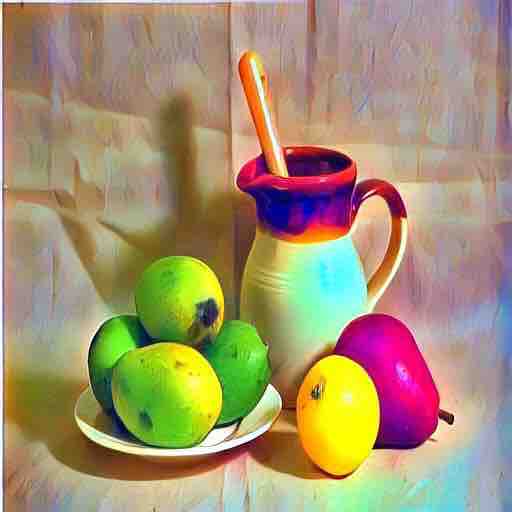}\\[0.5mm]
    \includegraphics[width=\textwidth]{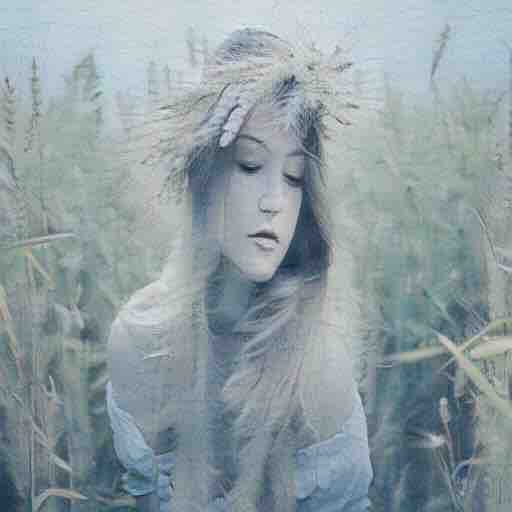}\\[0.5mm]
    \includegraphics[width=\textwidth]{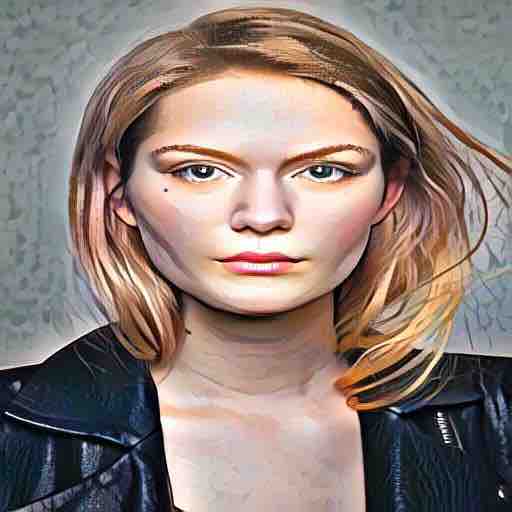}\\[0.5mm]
    \includegraphics[width=\textwidth]{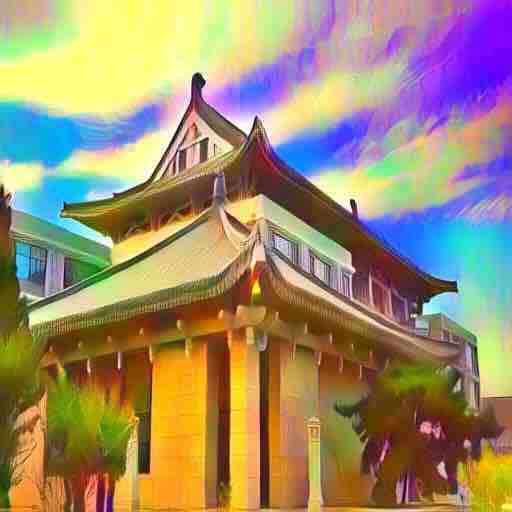}\\[0.5mm]
    \includegraphics[width=\textwidth]{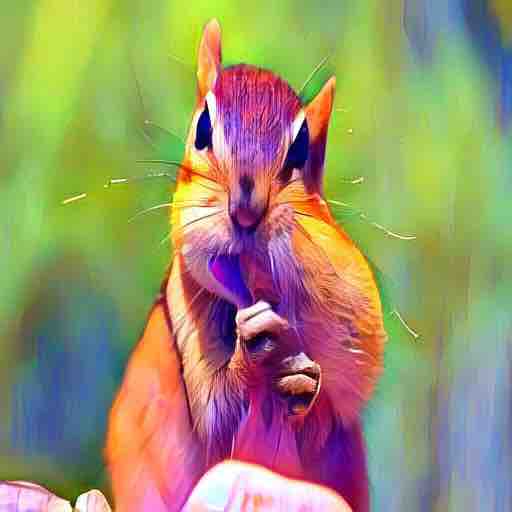}
    \end{minipage}
    }
    \caption{\textbf{Artistic style transfer comparison between the ArtNet(AdaIN) and AdaIN~\cite{huang2017arbitrary}.}}
    \label{fig:art_adain_1}
\end{figure*}
\begin{figure*}[h]
    \centering
    \subfloat[Content]{
    \begin{minipage}[t]{0.208\textwidth}
    \centering
    \includegraphics[width=\textwidth]{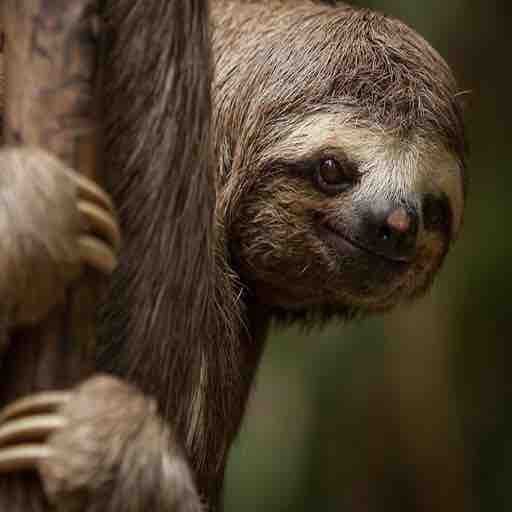}\\[0.5mm]
    \includegraphics[width=\textwidth]{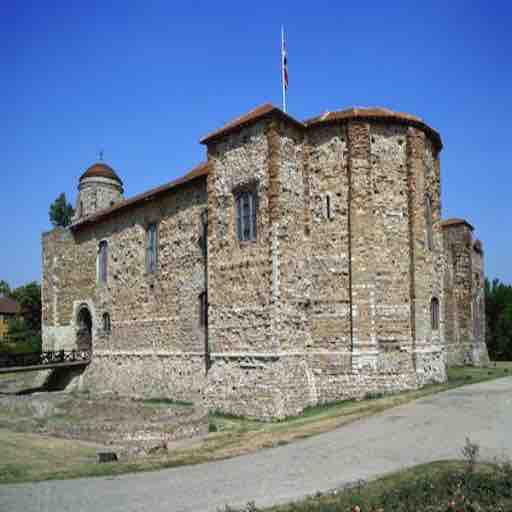}\\[0.5mm]
    \includegraphics[width=\textwidth]{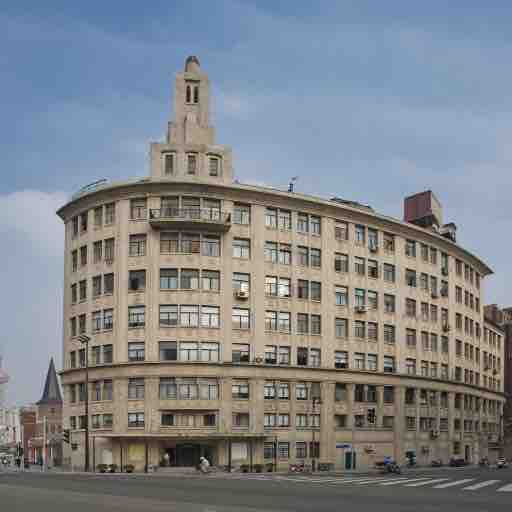}\\[0.5mm]
    \includegraphics[width=\textwidth]{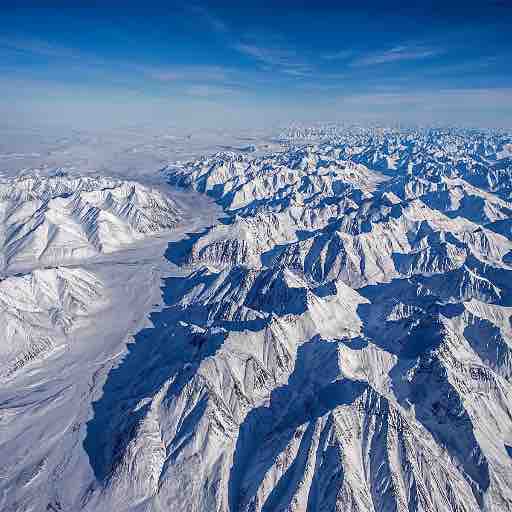}\\[0.5mm]
    \includegraphics[width=\textwidth]{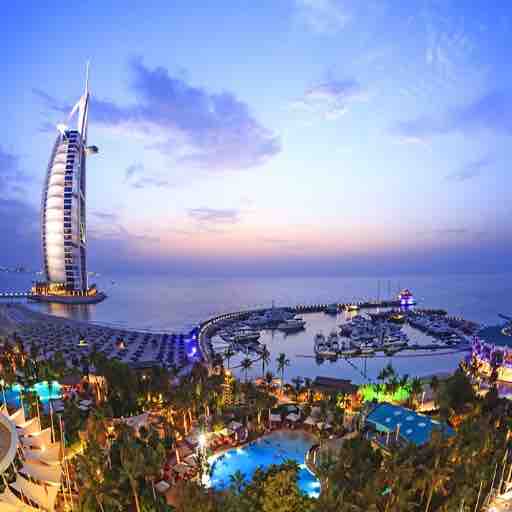}
    \end{minipage}
    }
    \hspace{-1.5mm}
    \subfloat[Style]{
    \begin{minipage}[t]{0.208\textwidth}
    \centering
    \includegraphics[width=\textwidth]{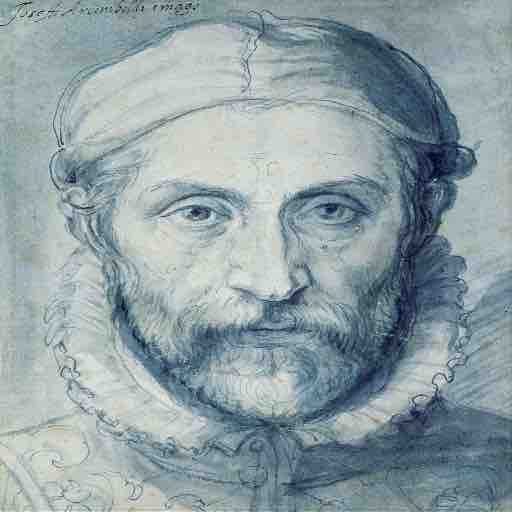}\\[0.5mm]
    \includegraphics[width=\textwidth]{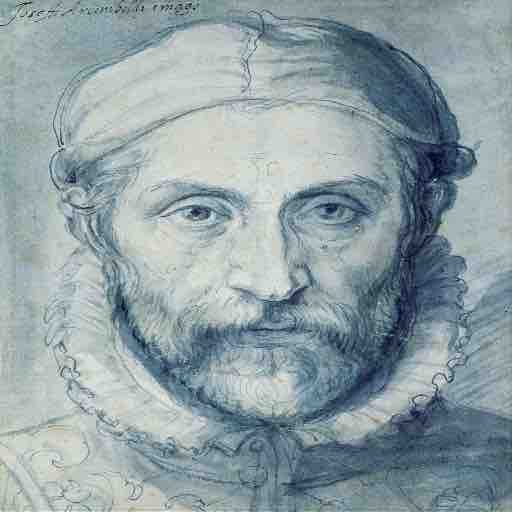}\\[0.5mm]
    \includegraphics[width=\textwidth]{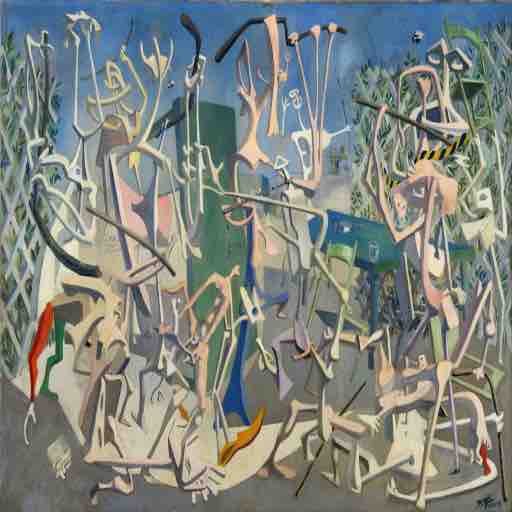}\\[0.5mm]
    \includegraphics[width=\textwidth]{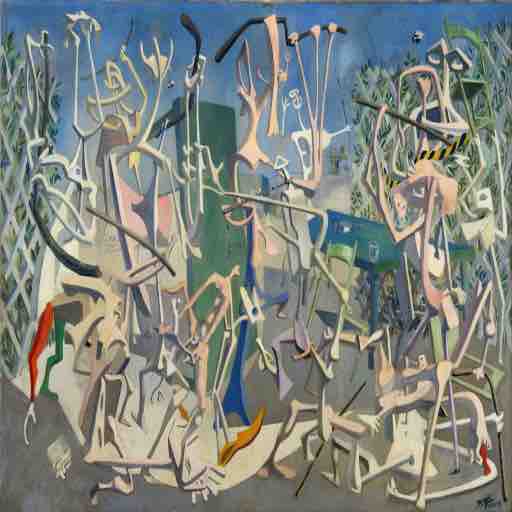}\\[0.5mm]
    \includegraphics[width=\textwidth]{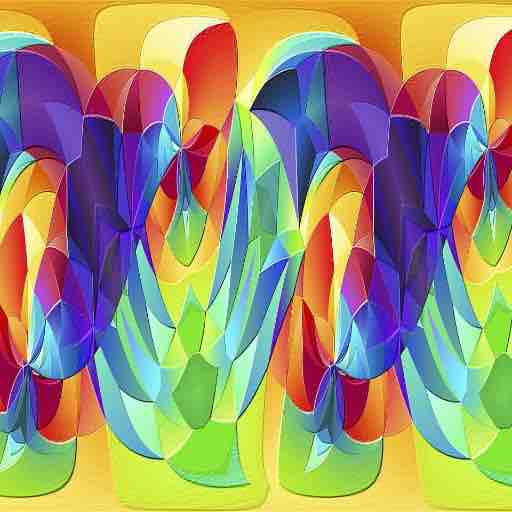}
    \end{minipage}
    }
    \hspace{-1.5mm}
    \subfloat[AdaIN~\cite{huang2017arbitrary}]{
    \begin{minipage}[t]{0.208\textwidth}
    \centering
    \includegraphics[width=\textwidth]{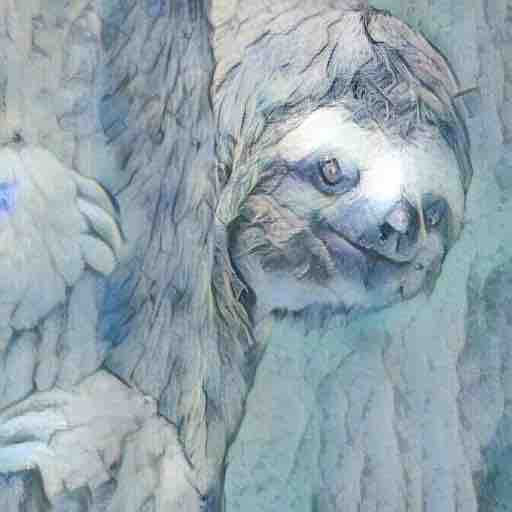}\\[0.5mm]
    \includegraphics[width=\textwidth]{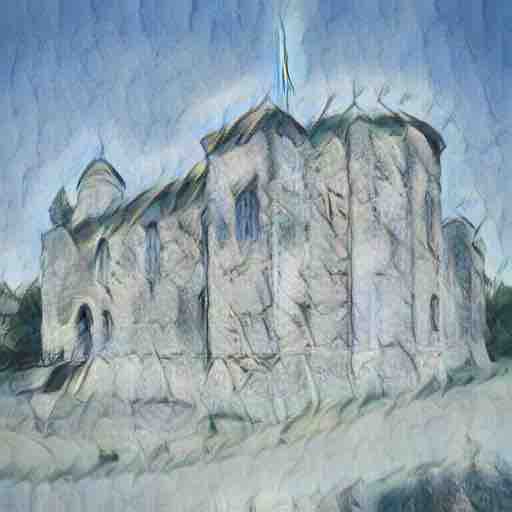}\\[0.5mm]
    \includegraphics[width=\textwidth]{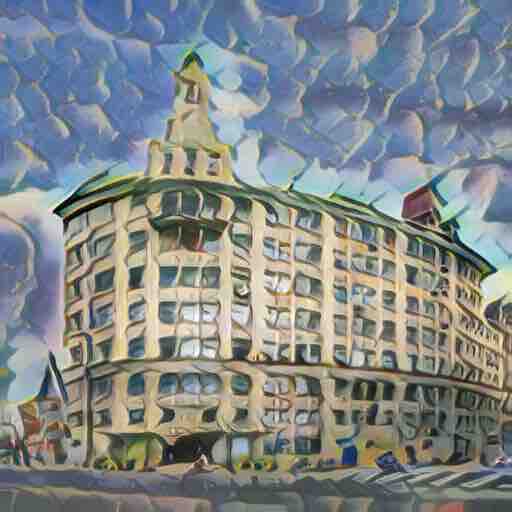}\\[0.5mm]
    \includegraphics[width=\textwidth]{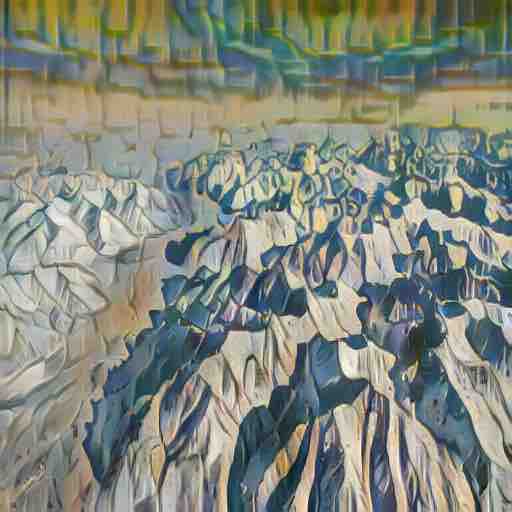}\\[0.5mm]
    \includegraphics[width=\textwidth]{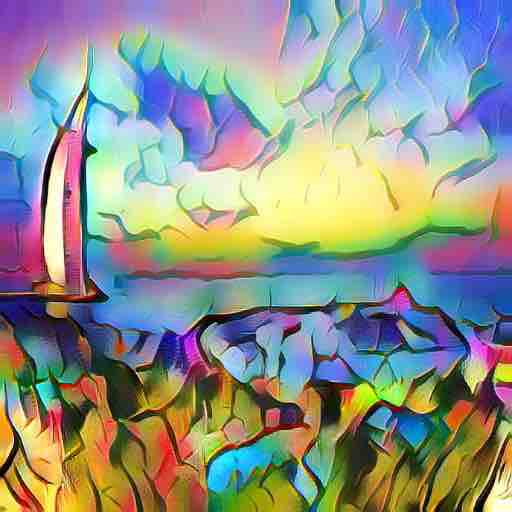}
    \end{minipage}
    }
    \hspace{-1.5mm}
    \subfloat[ArtNet(AdaIN)]{
    \begin{minipage}[t]{0.208\textwidth}
    \centering
    \includegraphics[width=\textwidth]{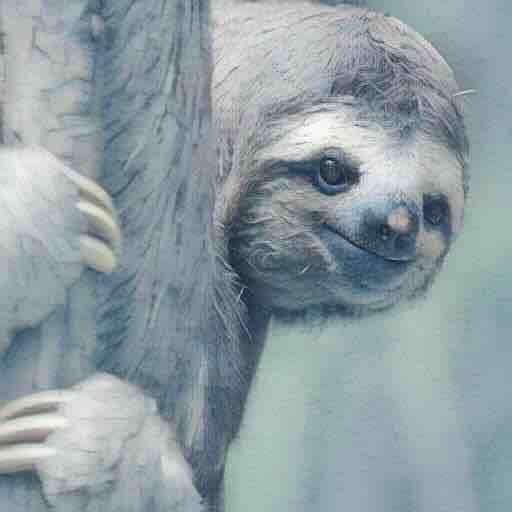}\\[0.5mm]
    \includegraphics[width=\textwidth]{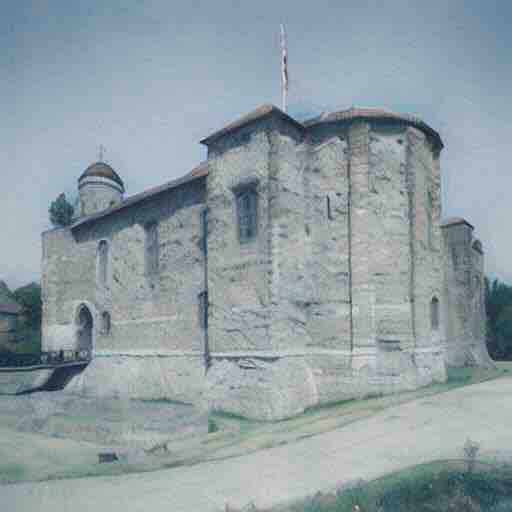}\\[0.5mm]
    \includegraphics[width=\textwidth]{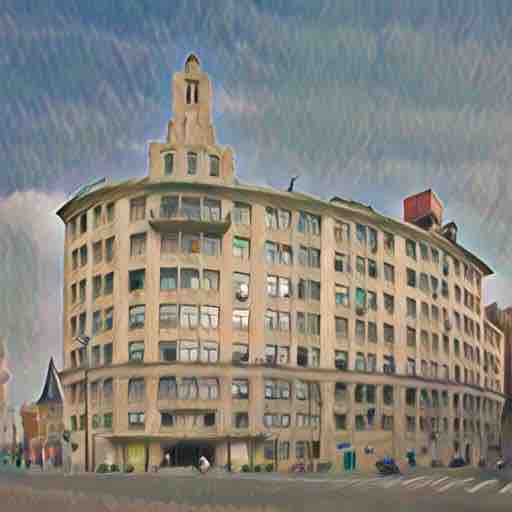}\\[0.5mm]
    \includegraphics[width=\textwidth]{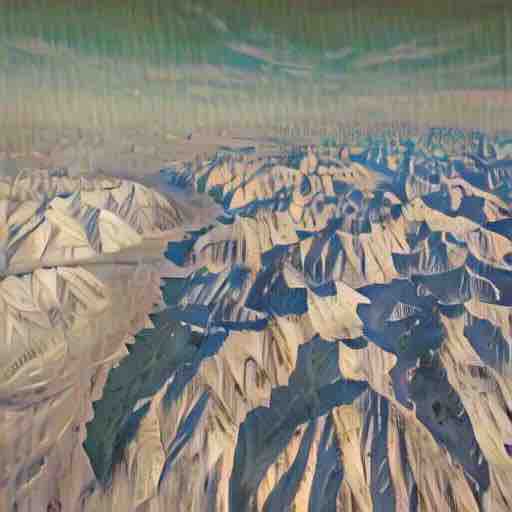}\\[0.5mm]
    \includegraphics[width=\textwidth]{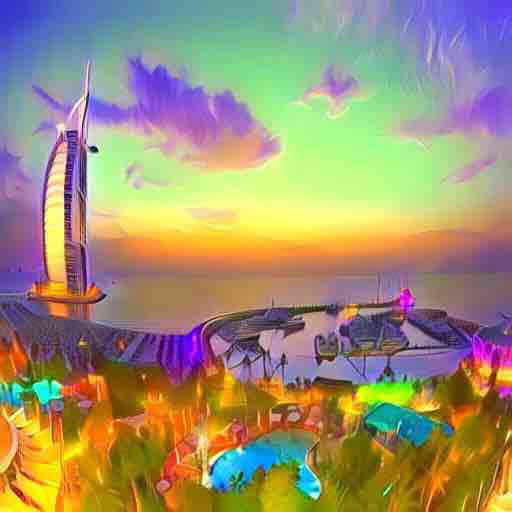}
    \end{minipage}
    }
    \caption{\textbf{Artistic style transfer comparison between the ArtNet(AdaIN) and AdaIN~\cite{huang2017arbitrary}.}}
    \label{fig:art_adain_2}
\end{figure*}
\begin{figure*}[h]
    \centering
    \subfloat[Content]{
    \begin{minipage}[t]{0.208\textwidth}
    \centering
    \includegraphics[width=\textwidth]{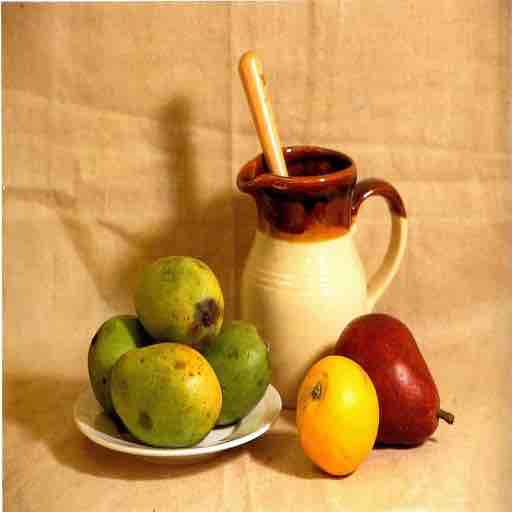}\\[0.5mm]
    \includegraphics[width=\textwidth]{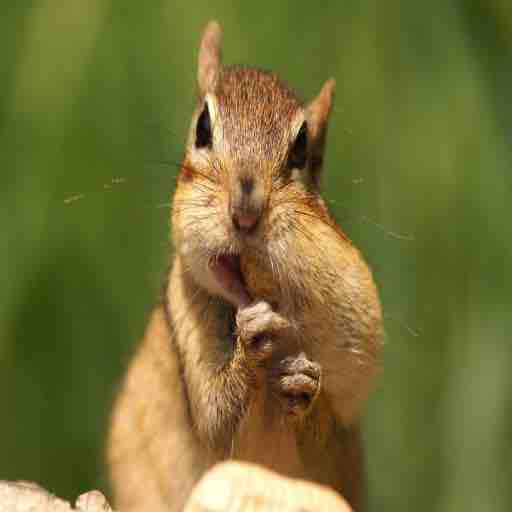}\\[0.5mm]
    \includegraphics[width=\textwidth]{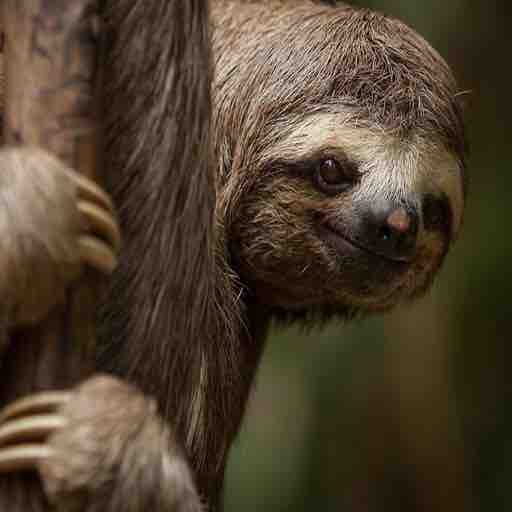}\\[0.5mm]
    \includegraphics[width=\textwidth]{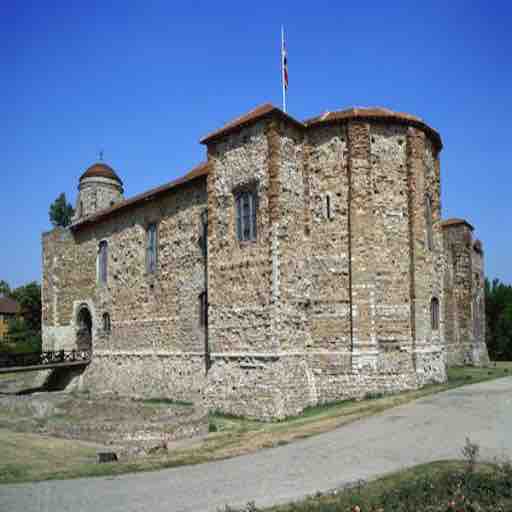}\\[0.5mm]
    \includegraphics[width=\textwidth]{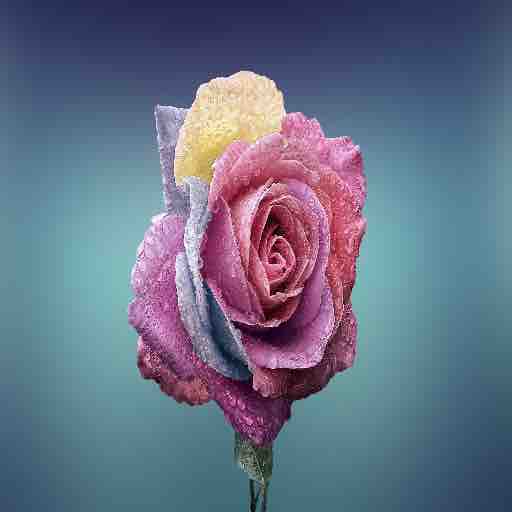}
    \end{minipage}
    }
    \hspace{-1.5mm}
    \subfloat[Style]{
    \begin{minipage}[t]{0.208\textwidth}
    \centering
    \includegraphics[width=\textwidth]{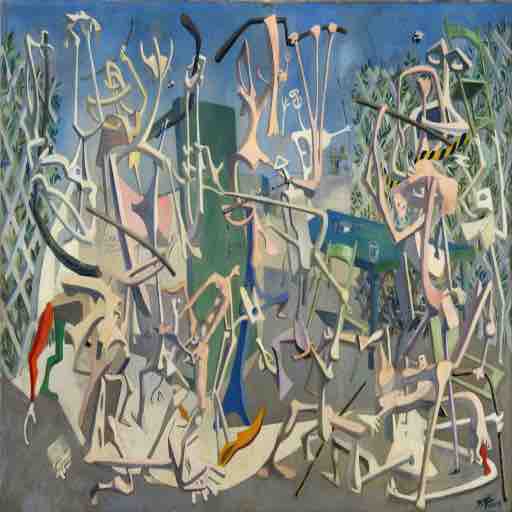}\\[0.5mm]
    \includegraphics[width=\textwidth]{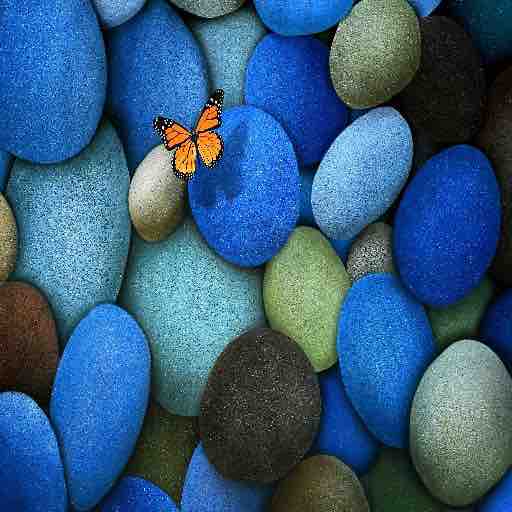}\\[0.5mm]
    \includegraphics[width=\textwidth]{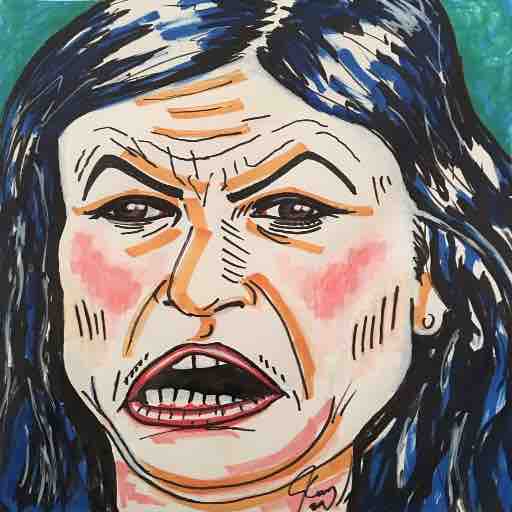}\\[0.5mm]
    \includegraphics[width=\textwidth]{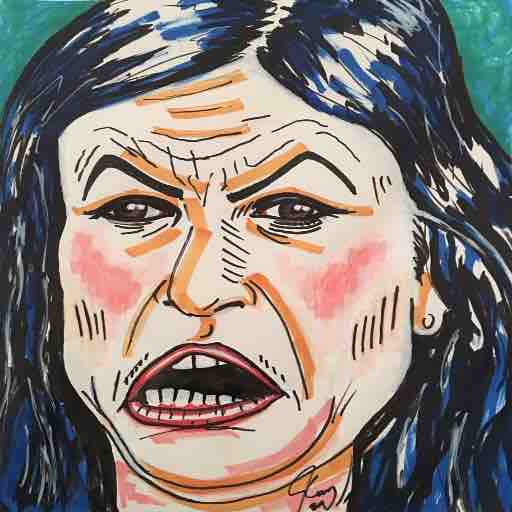}\\[0.5mm]
    \includegraphics[width=\textwidth]{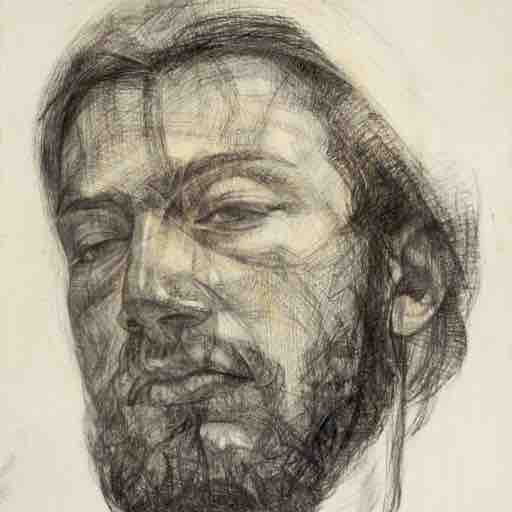}
    \end{minipage}
    }
    \hspace{-1.5mm}
    \subfloat[AdaIN~\cite{huang2017arbitrary}]{
    \begin{minipage}[t]{0.208\textwidth}
    \centering
    \includegraphics[width=\textwidth]{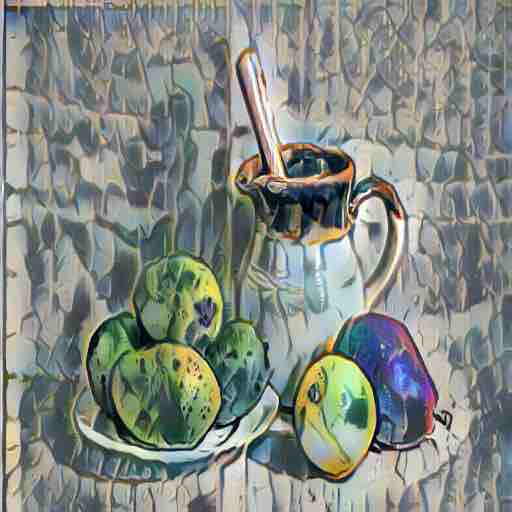}\\[0.5mm]
    \includegraphics[width=\textwidth]{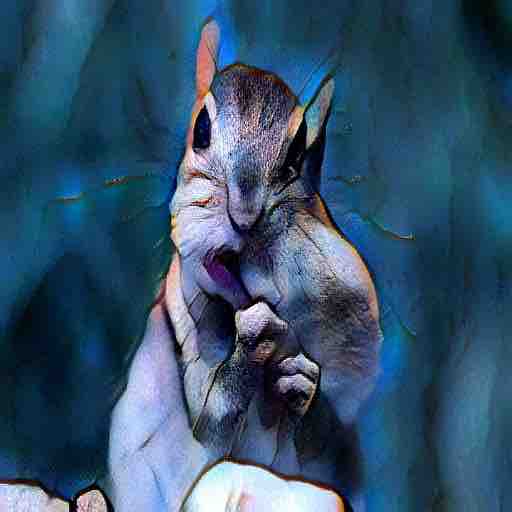}\\[0.5mm]
    \includegraphics[width=\textwidth]{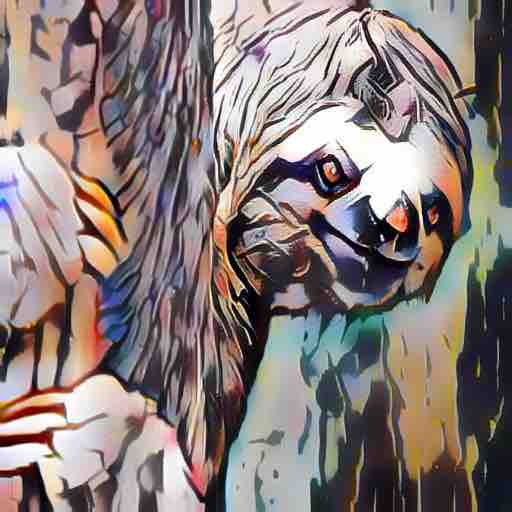}\\[0.5mm]
    \includegraphics[width=\textwidth]{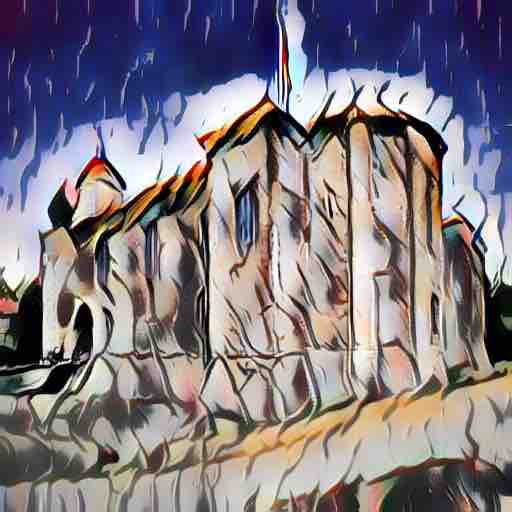}\\[0.5mm]
    \includegraphics[width=\textwidth]{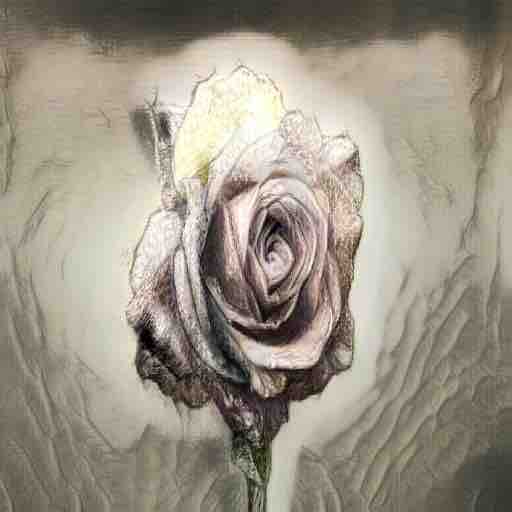}
    \end{minipage}
    }
    \hspace{-1.5mm}
    \subfloat[ArtNet(AdaIN)]{
    \begin{minipage}[t]{0.208\textwidth}
    \centering
    \includegraphics[width=\textwidth]{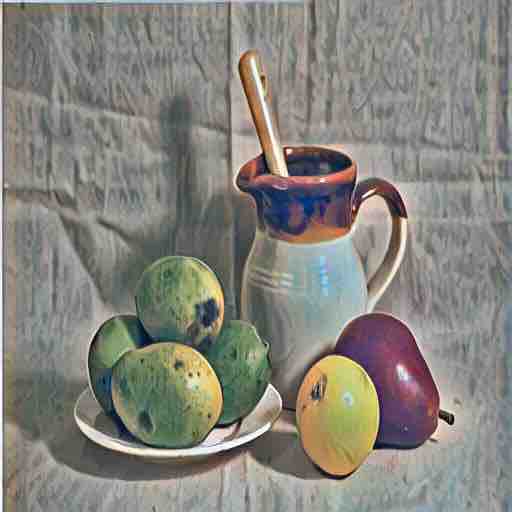}\\[0.5mm]
    \includegraphics[width=\textwidth]{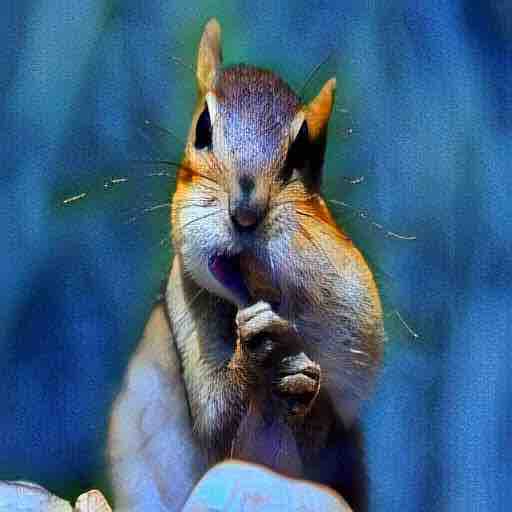}\\[0.5mm]
    \includegraphics[width=\textwidth]{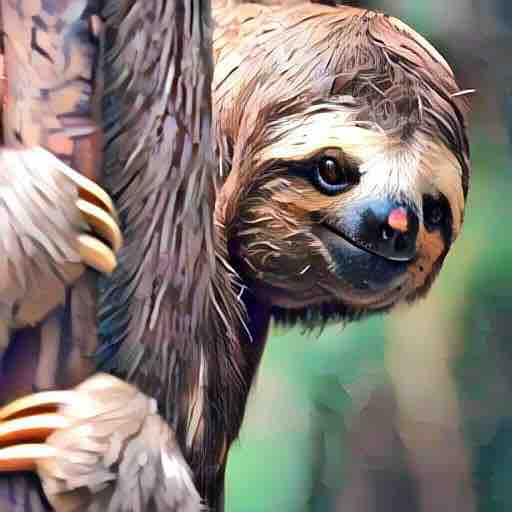}\\[0.5mm]
    \includegraphics[width=\textwidth]{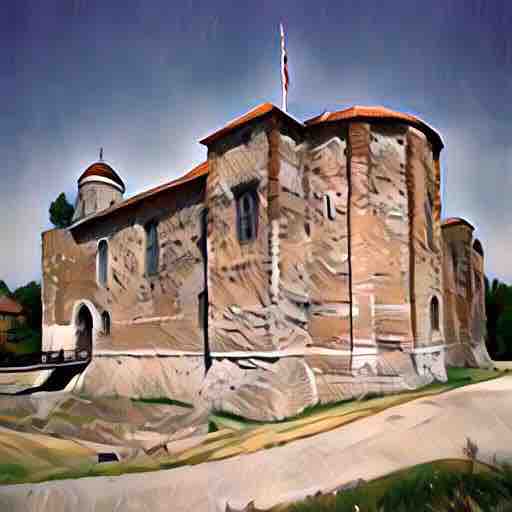}\\[0.5mm]
    \includegraphics[width=\textwidth]{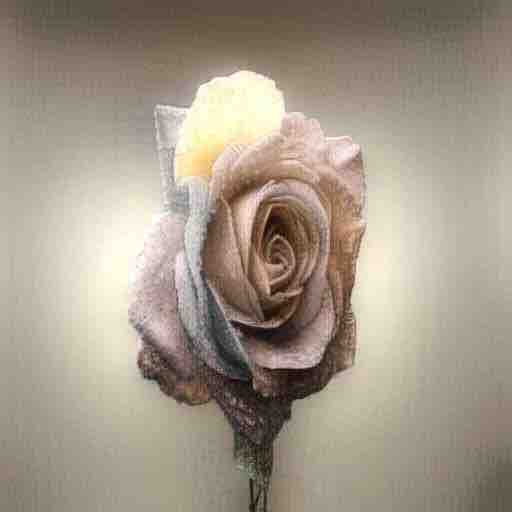}
    \end{minipage}
    }
    \caption{\textbf{Artistic style transfer comparison between the ArtNet(AdaIN) and AdaIN~\cite{huang2017arbitrary}.}}
    \label{fig:art_adain_3}
\end{figure*}
\begin{figure*}[h]
    \centering
    \subfloat[Content]{
    \begin{minipage}[t]{0.208\textwidth}
    \centering
    \includegraphics[width=\textwidth]{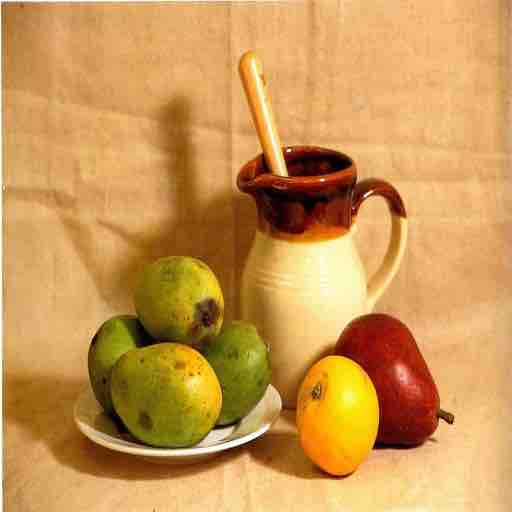}\\[0.5mm]
    \includegraphics[width=\textwidth]{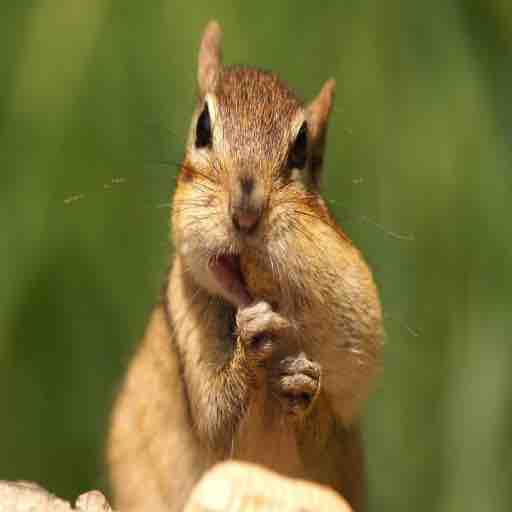}\\[0.5mm]
    \includegraphics[width=\textwidth]{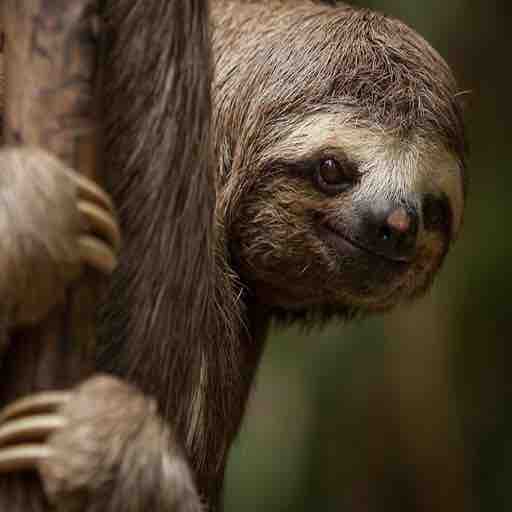}\\[0.5mm]
    \includegraphics[width=\textwidth]{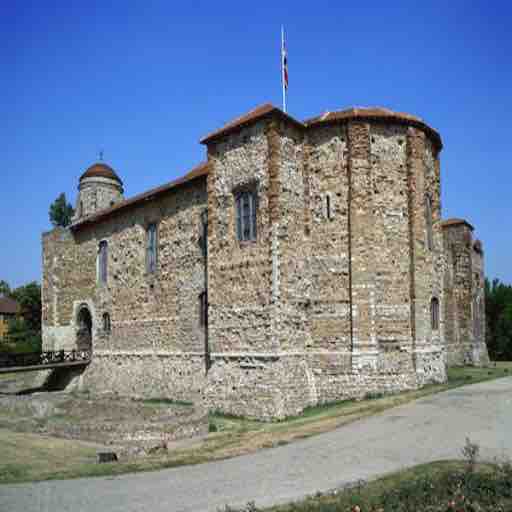}\\[0.5mm]
    \includegraphics[width=\textwidth]{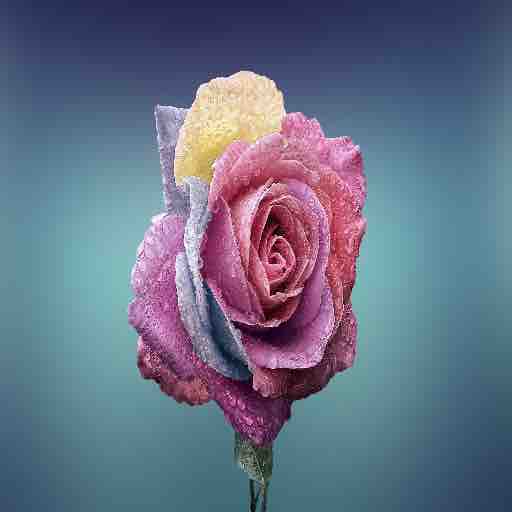}
    \end{minipage}
    }
    \hspace{-1.5mm}
    \subfloat[Style]{
    \begin{minipage}[t]{0.208\textwidth}
    \centering
    \includegraphics[width=\textwidth]{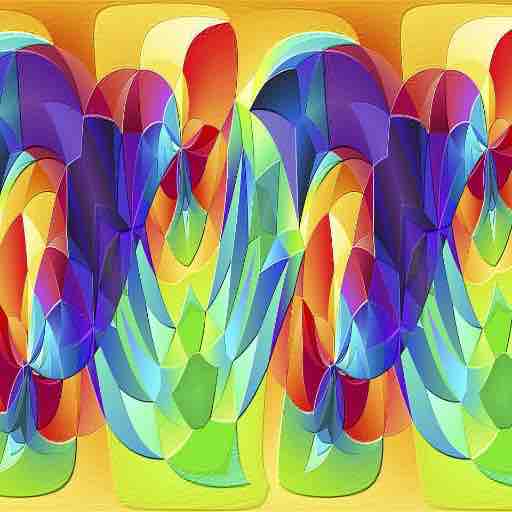}\\[0.5mm]
    \includegraphics[width=\textwidth]{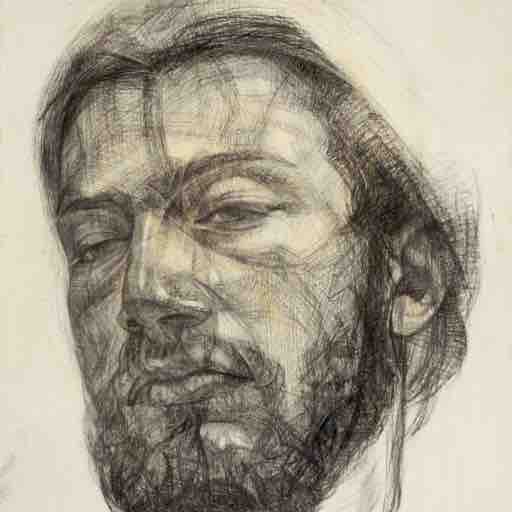}\\[0.5mm]
    \includegraphics[width=\textwidth]{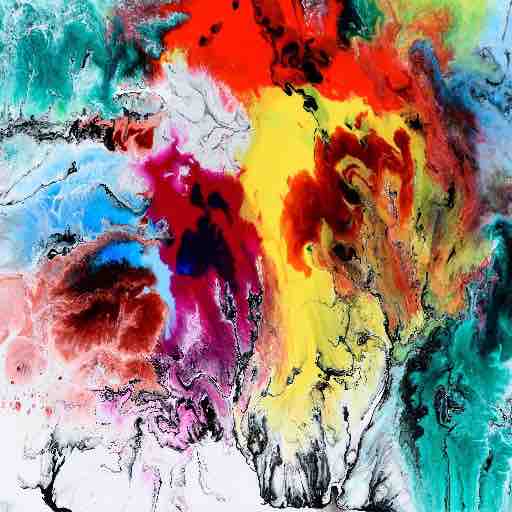}\\[0.5mm]
    \includegraphics[width=\textwidth]{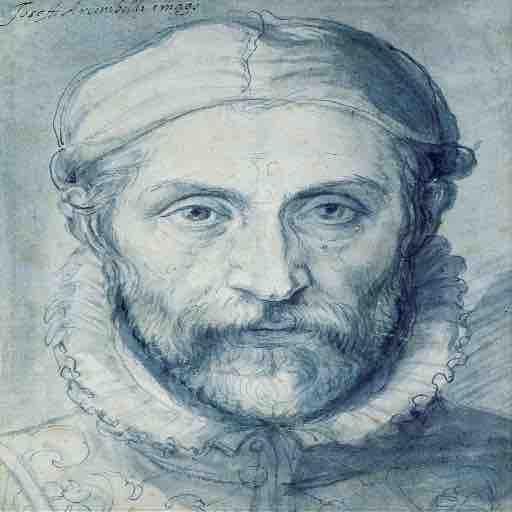}\\[0.5mm]
    \includegraphics[width=\textwidth]{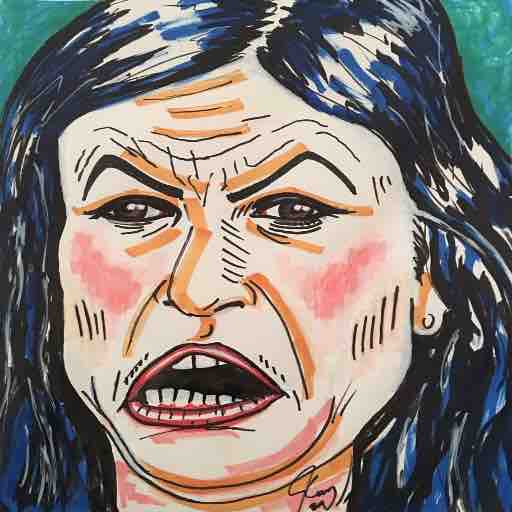}
    \end{minipage}
    }
    \hspace{-1.5mm}
    \subfloat[WCT~\cite{li2017universal}]{
    \begin{minipage}[t]{0.208\textwidth}
    \centering
    \includegraphics[width=\textwidth]{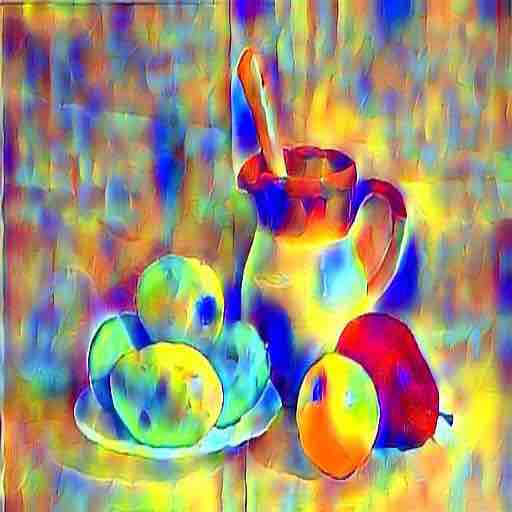}\\[0.5mm]
    \includegraphics[width=\textwidth]{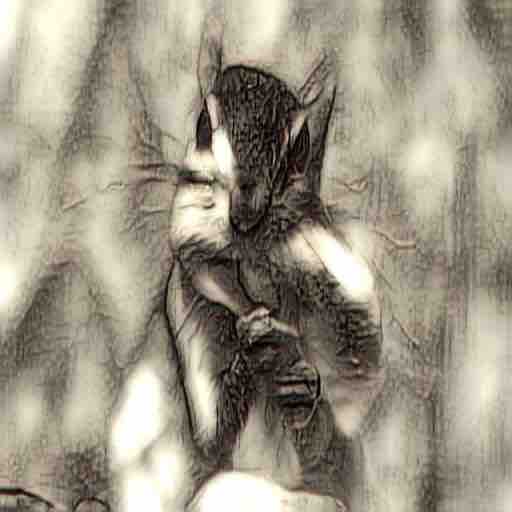}\\[0.5mm]
    \includegraphics[width=\textwidth]{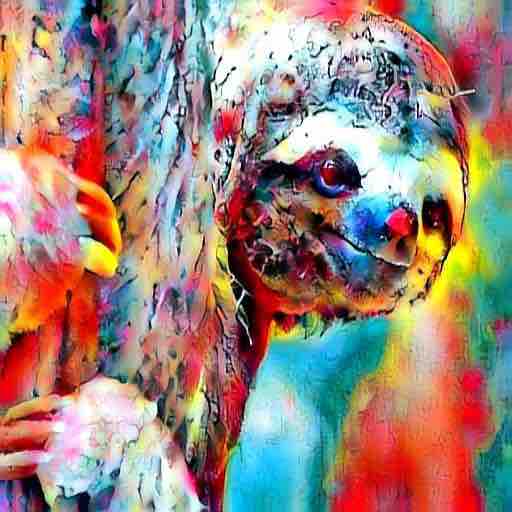}\\[0.5mm]
    \includegraphics[width=\textwidth]{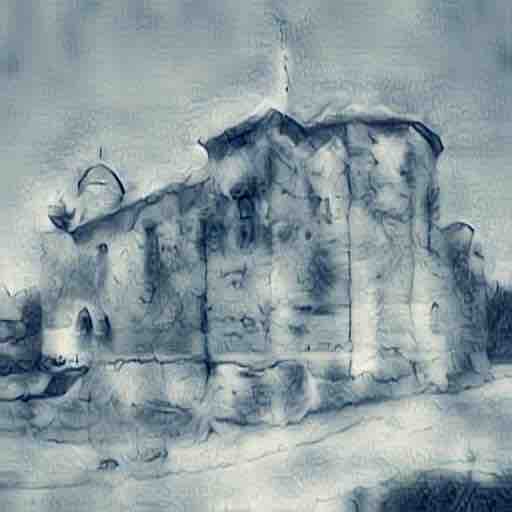}\\[0.5mm]
    \includegraphics[width=\textwidth]{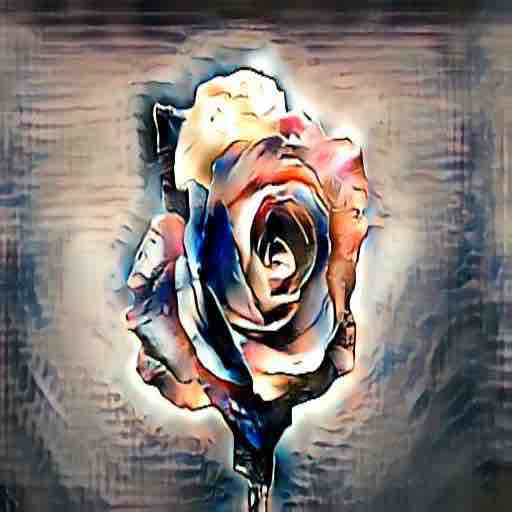}
    \end{minipage}
    }
    \hspace{-1.5mm}
    \subfloat[ArtNet(WCT)]{
    \begin{minipage}[t]{0.208\textwidth}
    \centering
    \includegraphics[width=\textwidth]{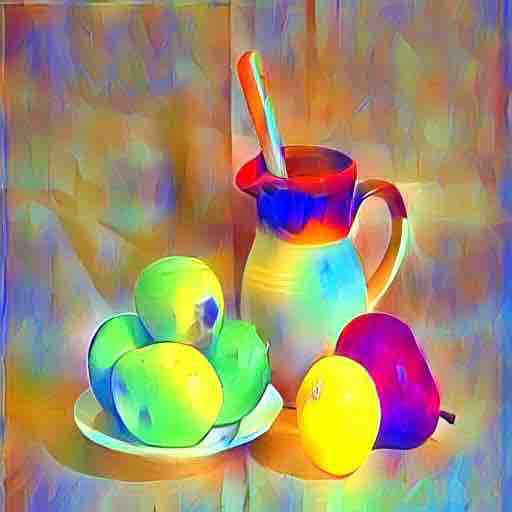}\\[0.5mm]
    \includegraphics[width=\textwidth]{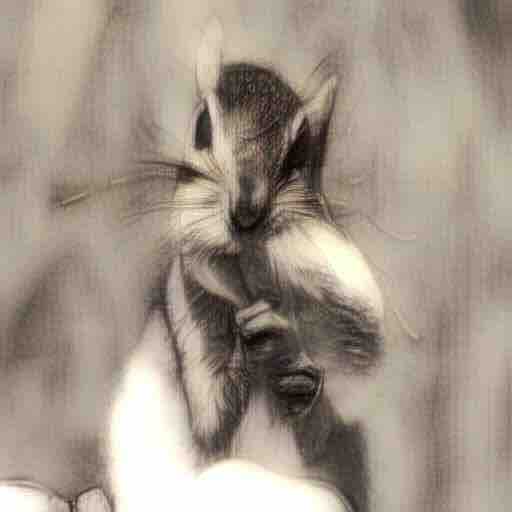}\\[0.5mm]
    \includegraphics[width=\textwidth]{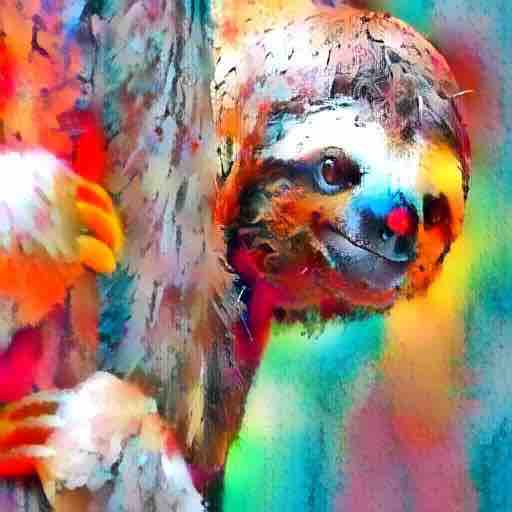}\\[0.5mm]
    \includegraphics[width=\textwidth]{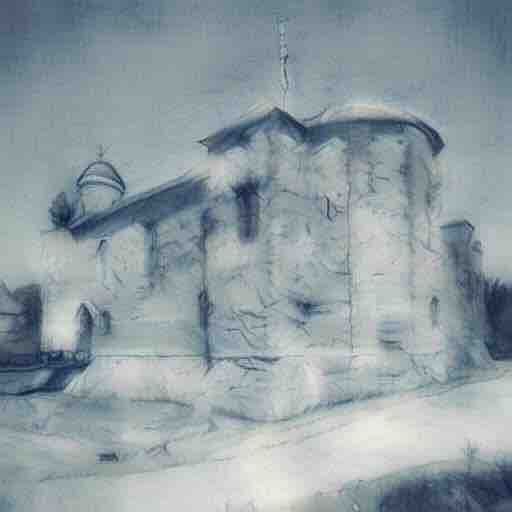}\\[0.5mm]
    \includegraphics[width=\textwidth]{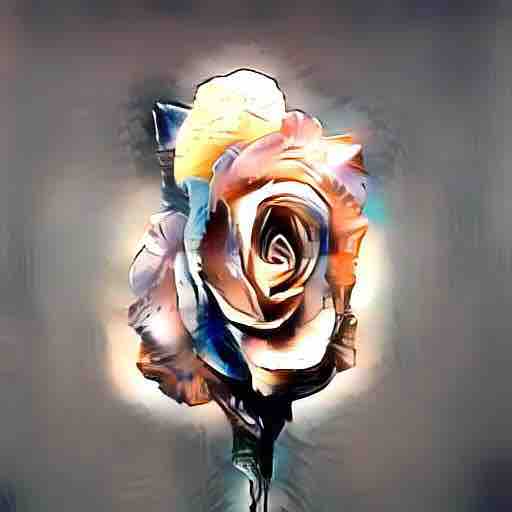}
    \end{minipage}
    }
    \caption{\textbf{Artistic style transfer comparison between the ArtNet(WCT) and WCT~\cite{li2017universal}.}}
    \label{fig:art_wct_1}
\end{figure*}
\begin{figure*}[h]
    \centering
    \subfloat[Content]{
    \begin{minipage}[t]{0.208\textwidth}
    \centering
    \includegraphics[width=\textwidth]{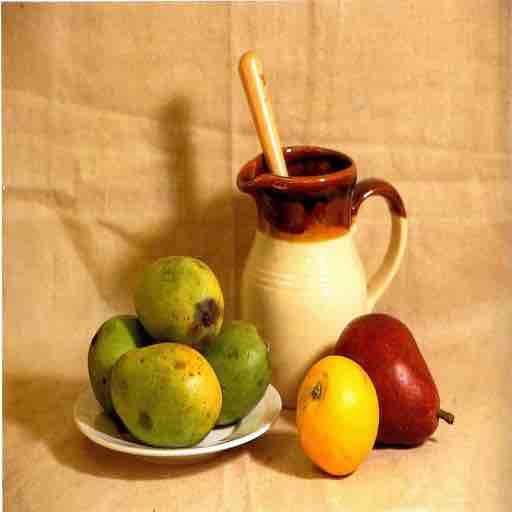}\\[0.5mm]
    \includegraphics[width=\textwidth]{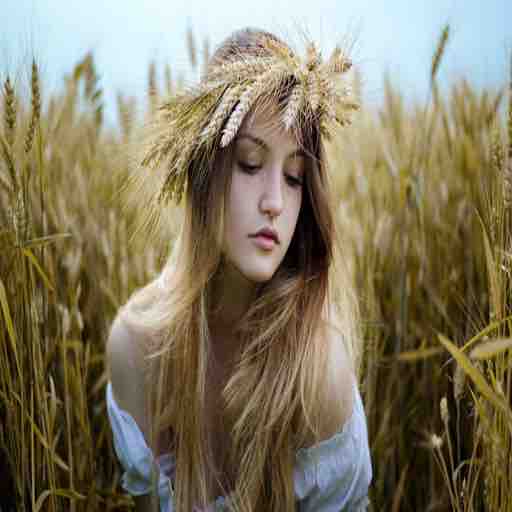}\\[0.5mm]
    \includegraphics[width=\textwidth]{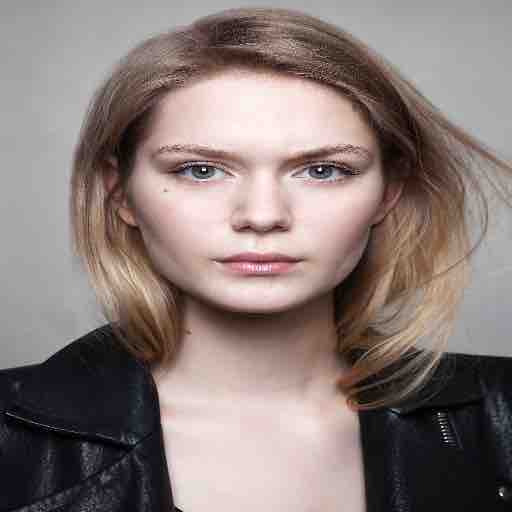}\\[0.5mm]
    \includegraphics[width=\textwidth]{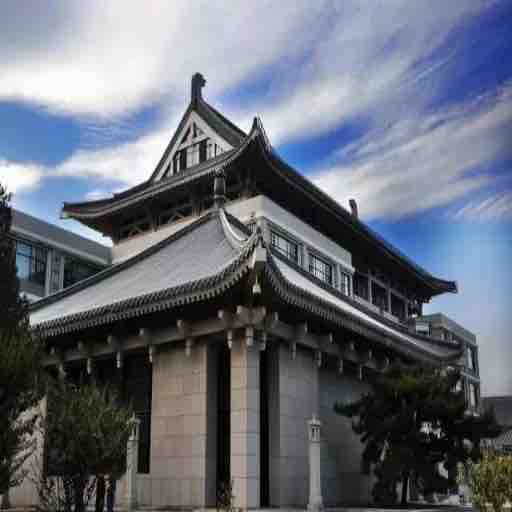}\\[0.5mm]
    \includegraphics[width=\textwidth]{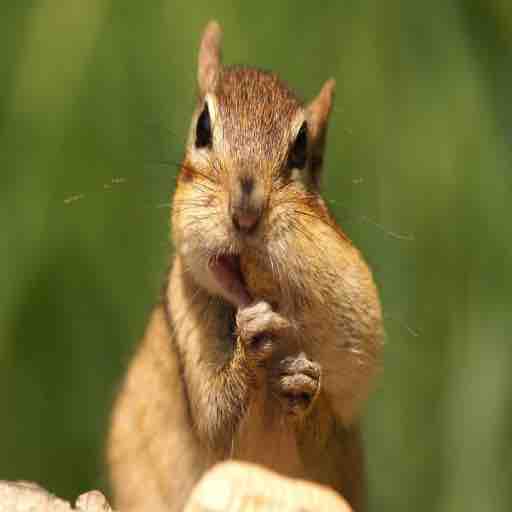}
    \end{minipage}
    }
    \hspace{-1.5mm}
    \subfloat[Style]{
    \begin{minipage}[t]{0.208\textwidth}
    \centering
    \includegraphics[width=\textwidth]{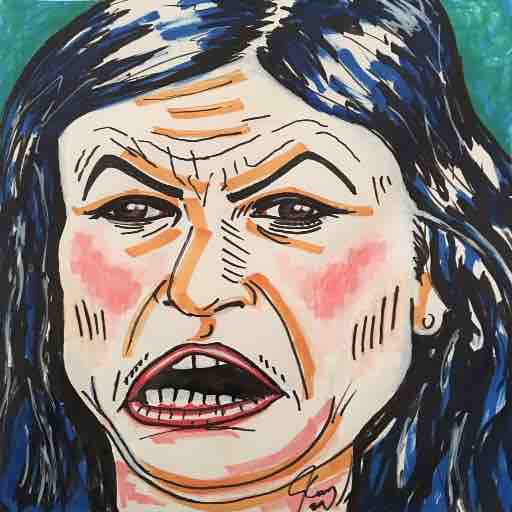}\\[0.5mm]
    \includegraphics[width=\textwidth]{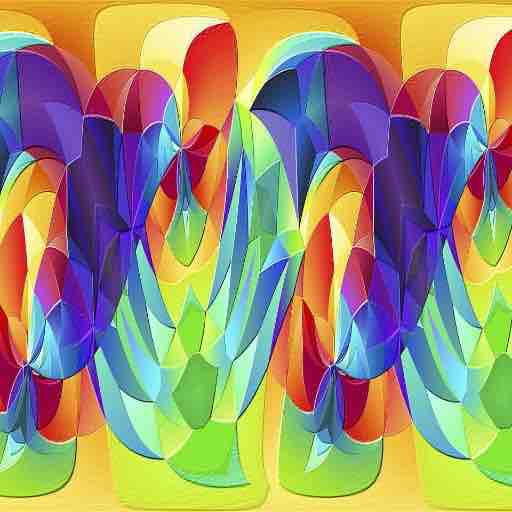}\\[0.5mm]
    \includegraphics[width=\textwidth]{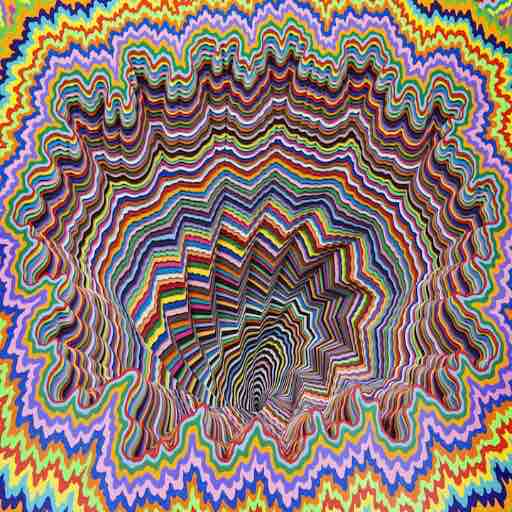}\\[0.5mm]
    \includegraphics[width=\textwidth]{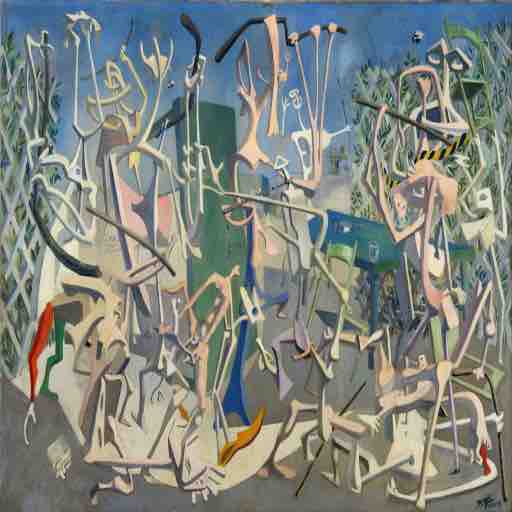}\\[0.5mm]
    \includegraphics[width=\textwidth]{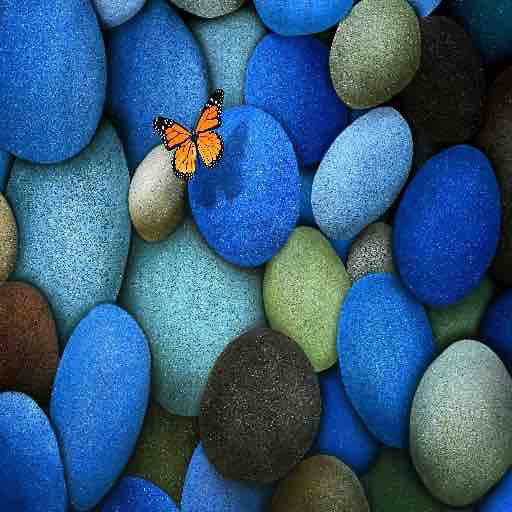}
    \end{minipage}
    }
    \hspace{-1.5mm}
    \subfloat[WCT~\cite{li2017universal}]{
    \begin{minipage}[t]{0.208\textwidth}
    \centering
    \includegraphics[width=\textwidth]{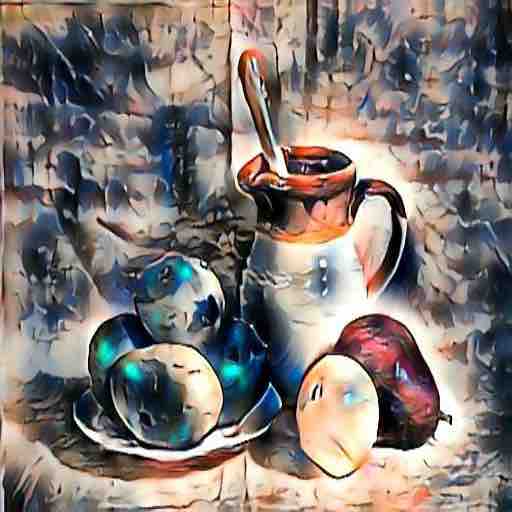}\\[0.5mm]
    \includegraphics[width=\textwidth]{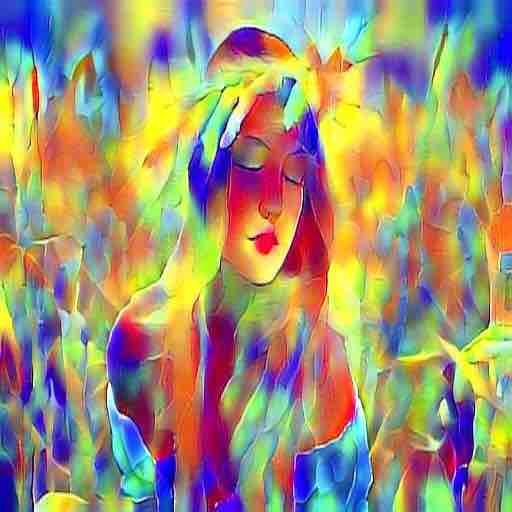}\\[0.5mm]
    \includegraphics[width=\textwidth]{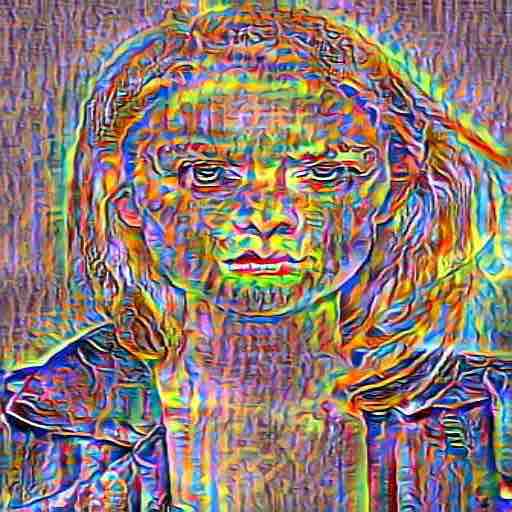}\\[0.5mm]
    \includegraphics[width=\textwidth]{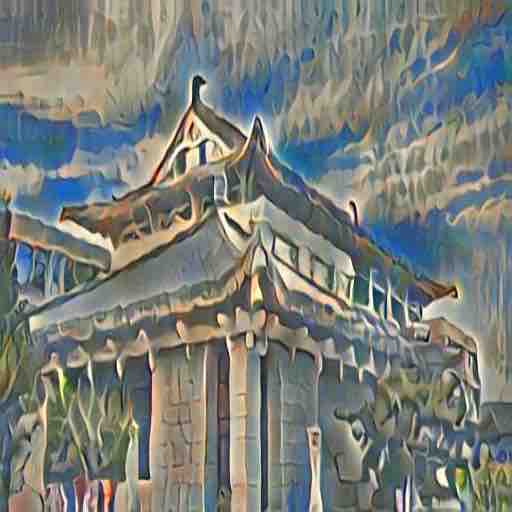}\\[0.5mm]
    \includegraphics[width=\textwidth]{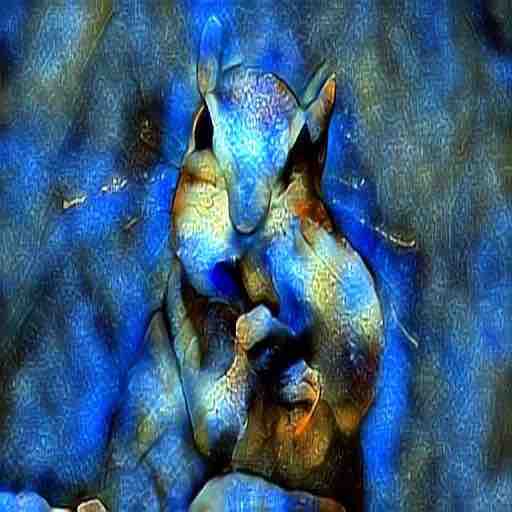}
    \end{minipage}
    }
    \hspace{-1.5mm}
    \subfloat[ArtNet(WCT)]{
    \begin{minipage}[t]{0.208\textwidth}
    \centering
    \includegraphics[width=\textwidth]{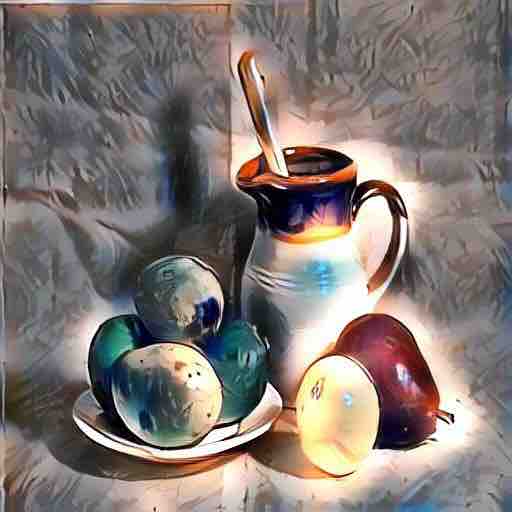}\\[0.5mm]
    \includegraphics[width=\textwidth]{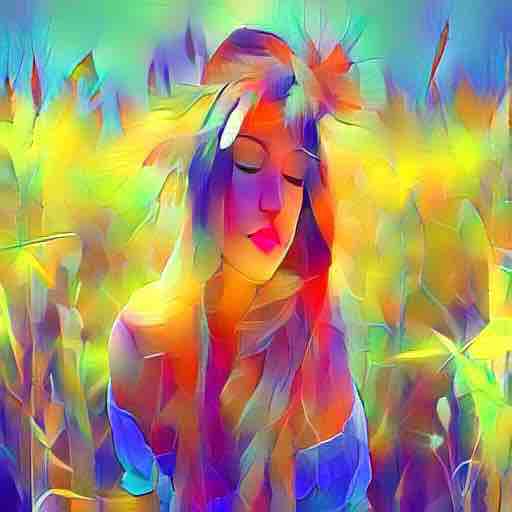}\\[0.5mm]
    \includegraphics[width=\textwidth]{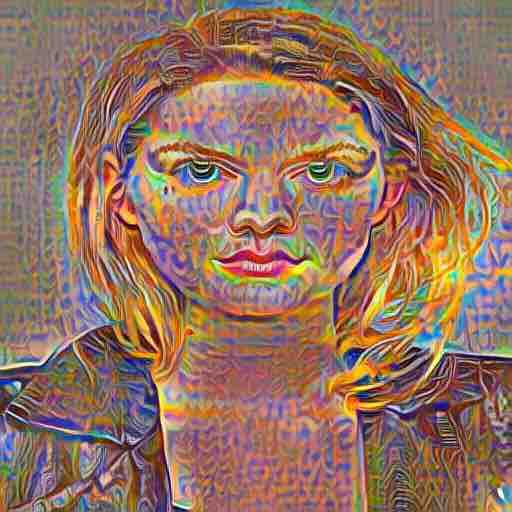}\\[0.5mm]
    \includegraphics[width=\textwidth]{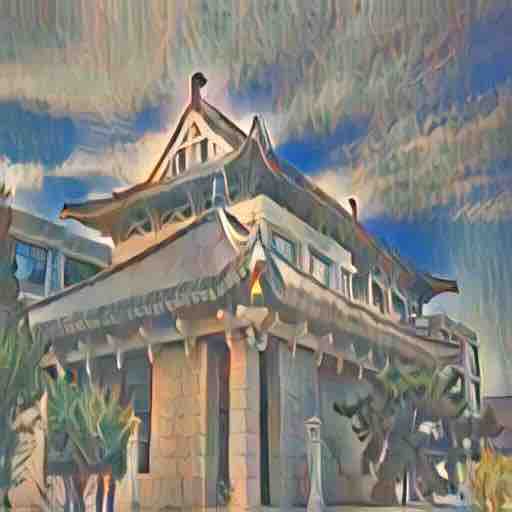}\\[0.5mm]
    \includegraphics[width=\textwidth]{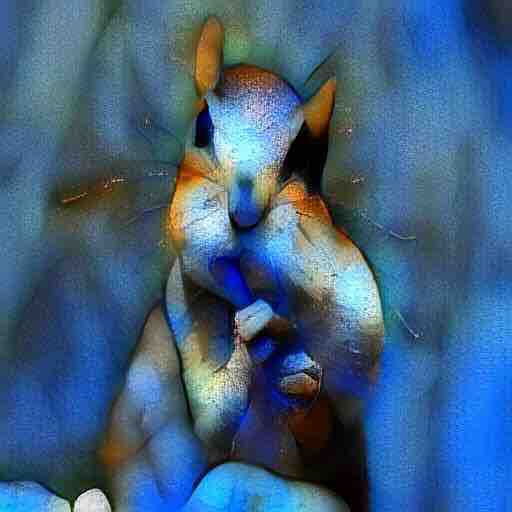}
    \end{minipage}
    }
    \caption{\textbf{Artistic style transfer comparison between the ArtNet(WCT) and WCT~\cite{li2017universal}.}}
    \label{fig:art_wct_2}
\end{figure*}
\begin{figure*}[h]
    \centering
    \subfloat[Content]{
    \begin{minipage}[t]{0.208\textwidth}
    \centering
    \includegraphics[width=\textwidth]{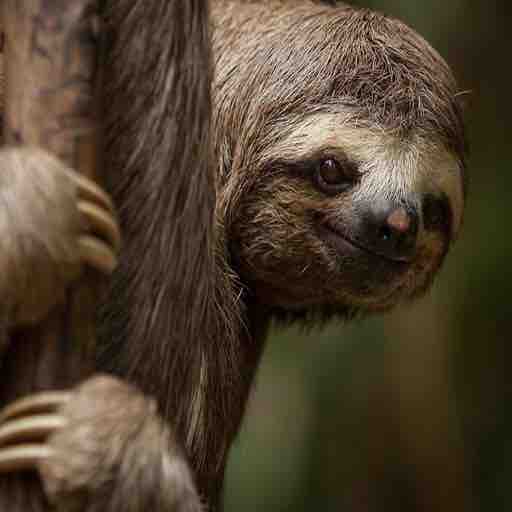}\\[0.5mm]
    \includegraphics[width=\textwidth]{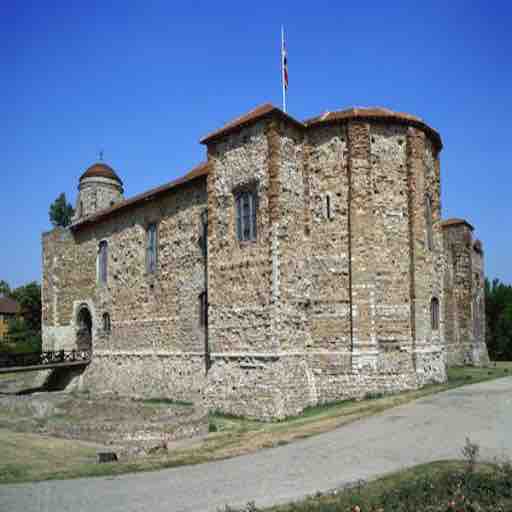}\\[0.5mm]
    \includegraphics[width=\textwidth]{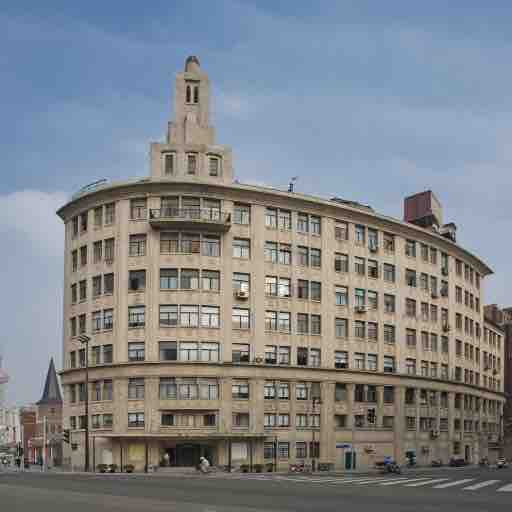}\\[0.5mm]
    \includegraphics[width=\textwidth]{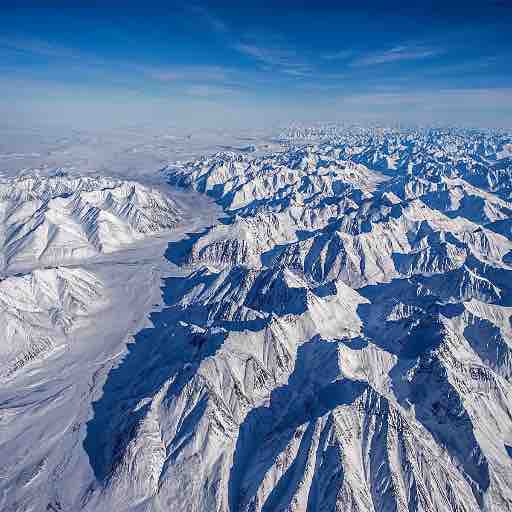}\\[0.5mm]
    \includegraphics[width=\textwidth]{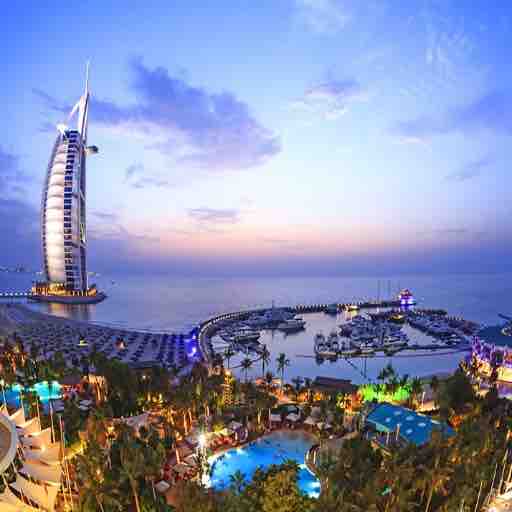}
    \end{minipage}
    }
    \hspace{-1.5mm}
    \subfloat[Style]{
    \begin{minipage}[t]{0.208\textwidth}
    \centering
    \includegraphics[width=\textwidth]{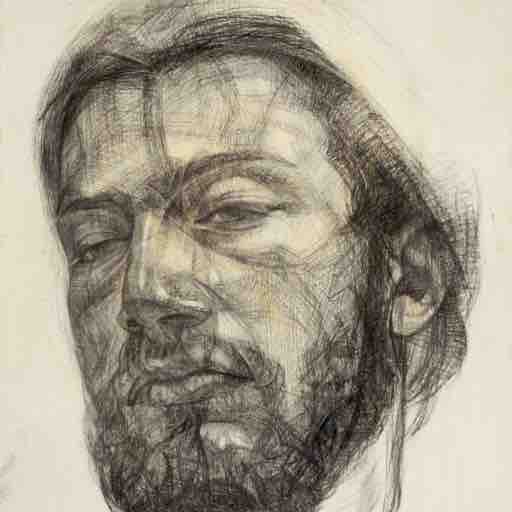}\\[0.5mm]
    \includegraphics[width=\textwidth]{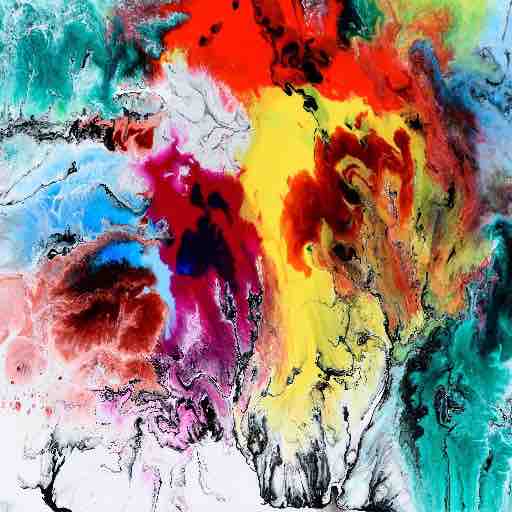}\\[0.5mm]
    \includegraphics[width=\textwidth]{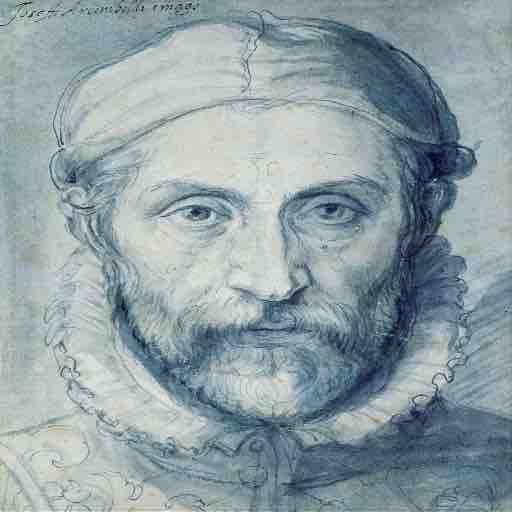}\\[0.5mm]
    \includegraphics[width=\textwidth]{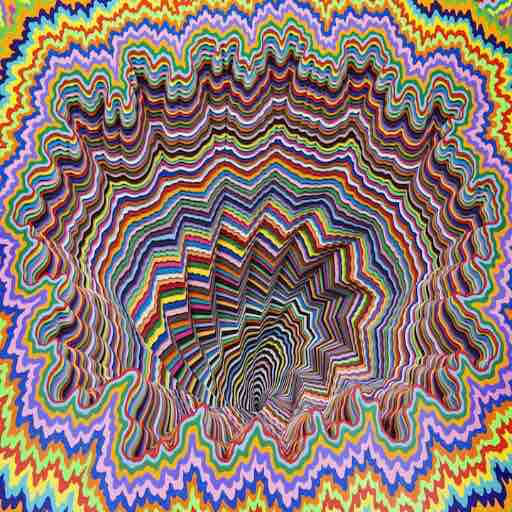}\\[0.5mm]
    \includegraphics[width=\textwidth]{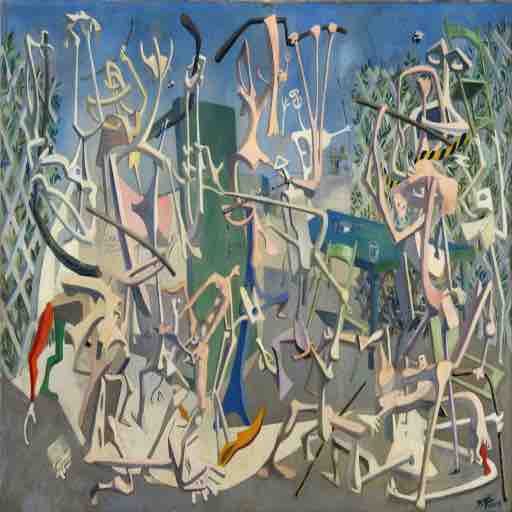}
    \end{minipage}
    }
    \hspace{-1.5mm}
    \subfloat[WCT~\cite{li2017universal}]{
    \begin{minipage}[t]{0.208\textwidth}
    \centering
    \includegraphics[width=\textwidth]{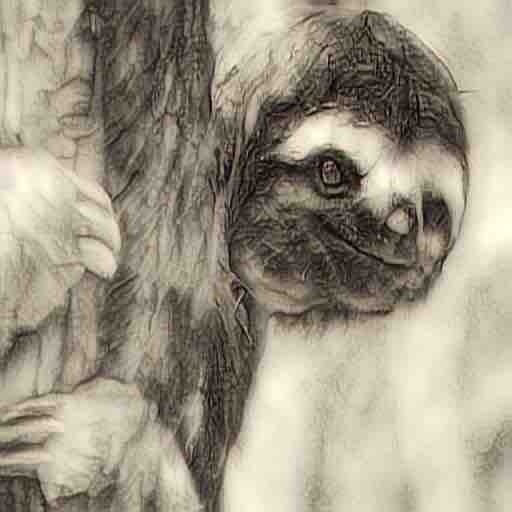}\\[0.5mm]
    \includegraphics[width=\textwidth]{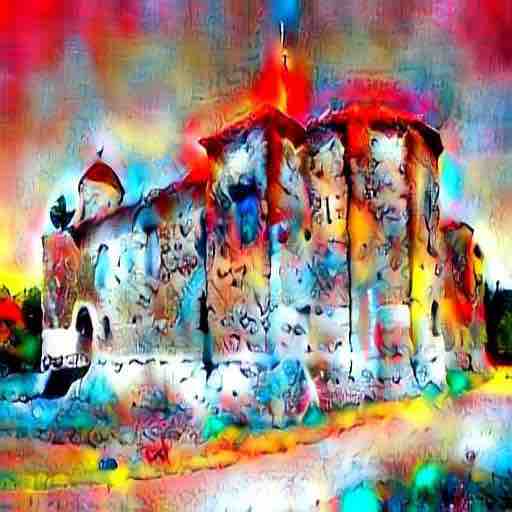}\\[0.5mm]
    \includegraphics[width=\textwidth]{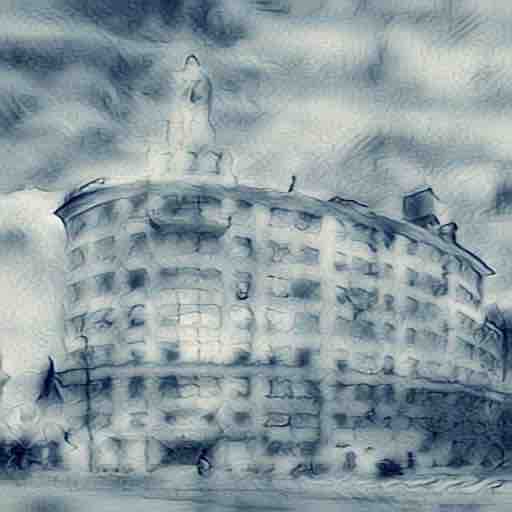}\\[0.5mm]
    \includegraphics[width=\textwidth]{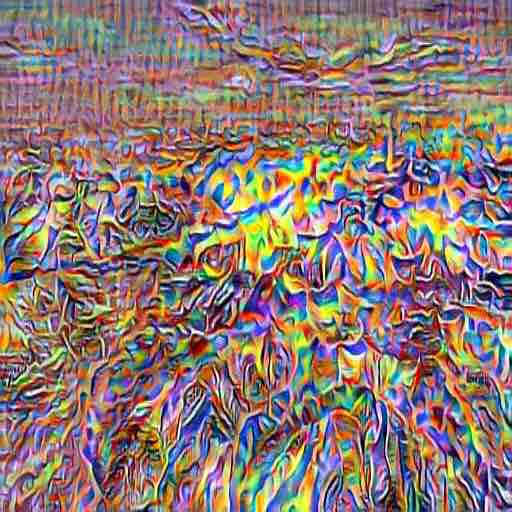}\\[0.5mm]
    \includegraphics[width=\textwidth]{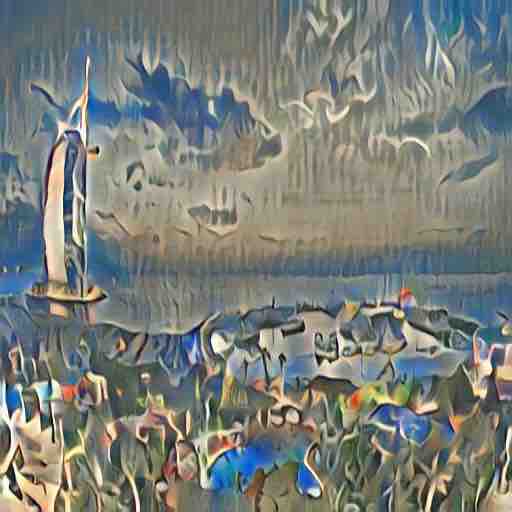}
    \end{minipage}
    }
    \hspace{-1.5mm}
    \subfloat[ArtNet(WCT)]{
    \begin{minipage}[t]{0.208\textwidth}
    \centering
    \includegraphics[width=\textwidth]{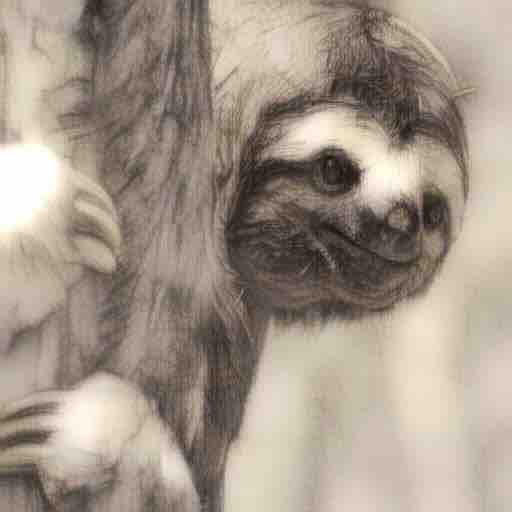}\\[0.5mm]
    \includegraphics[width=\textwidth]{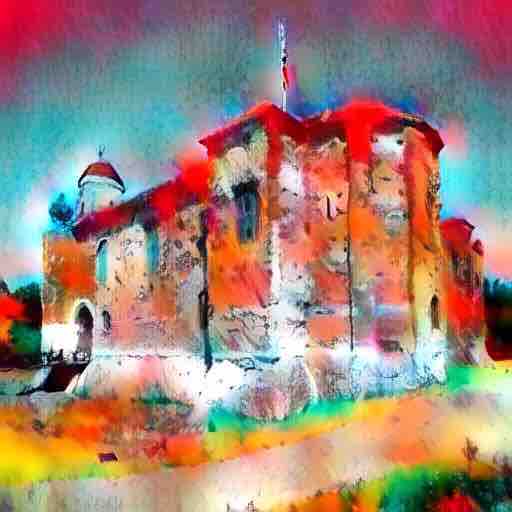}\\[0.5mm]
    \includegraphics[width=\textwidth]{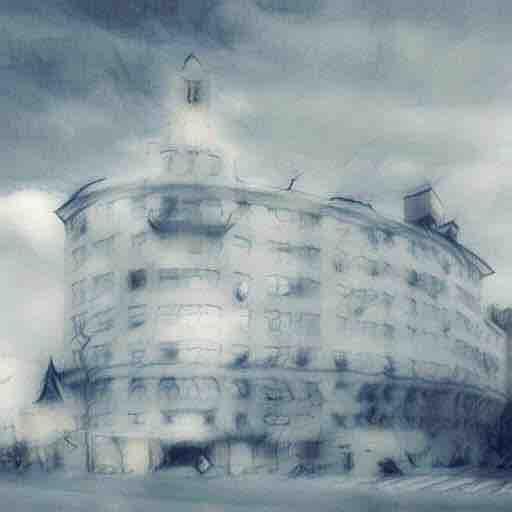}\\[0.5mm]
    \includegraphics[width=\textwidth]{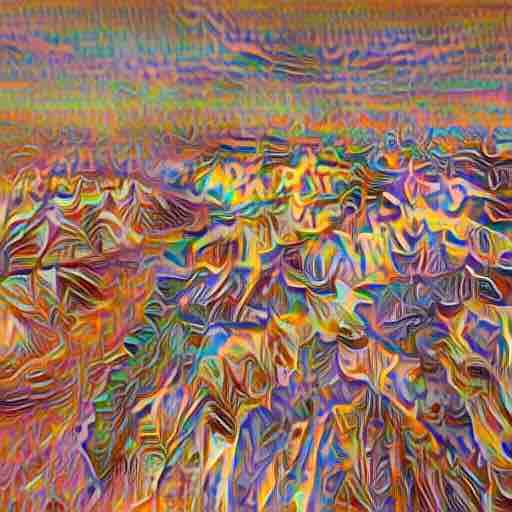}\\[0.5mm]
    \includegraphics[width=\textwidth]{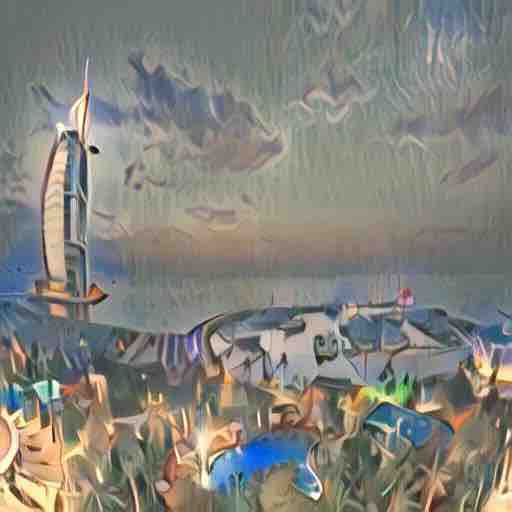}
    \end{minipage}
    }
    \caption{\textbf{Artistic style transfer comparison between the ArtNet(WCT) and WCT~\cite{li2017universal}.}}
    \label{fig:art_wct_3}
\end{figure*}
\begin{figure*}[h]
    \centering
    \subfloat[Content]{
    \begin{minipage}[t]{0.208\textwidth}
    \centering
    \includegraphics[width=\textwidth]{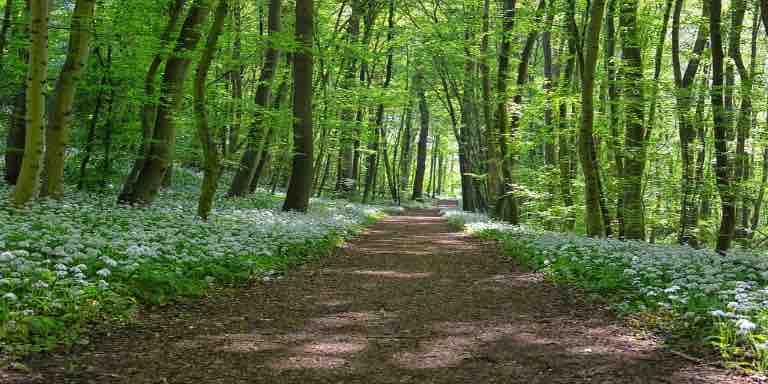}\\[0.5mm]
    \includegraphics[width=\textwidth]{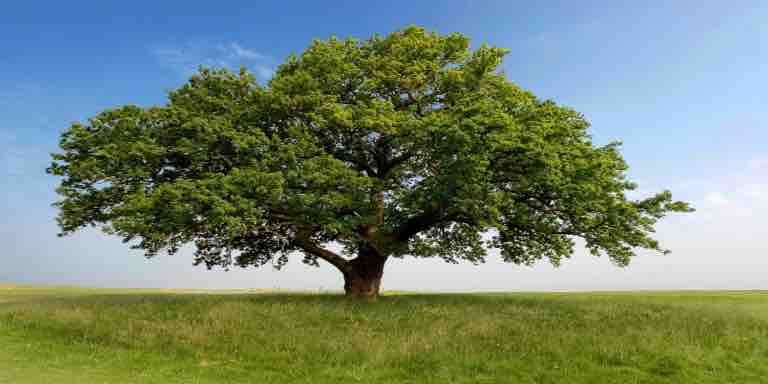}\\[0.5mm]
    \includegraphics[width=\textwidth]{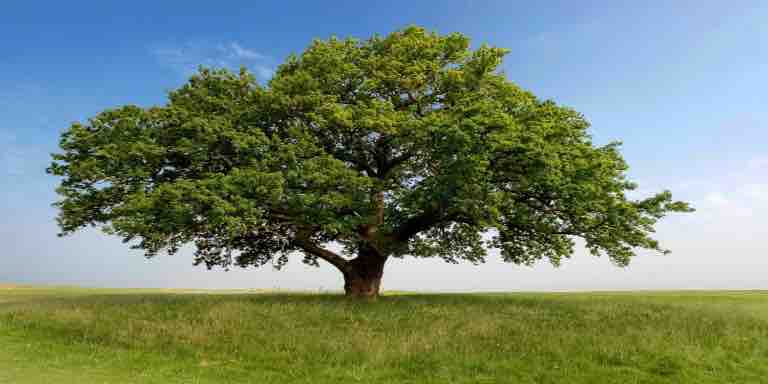}\\[0.5mm]
    \includegraphics[width=\textwidth]{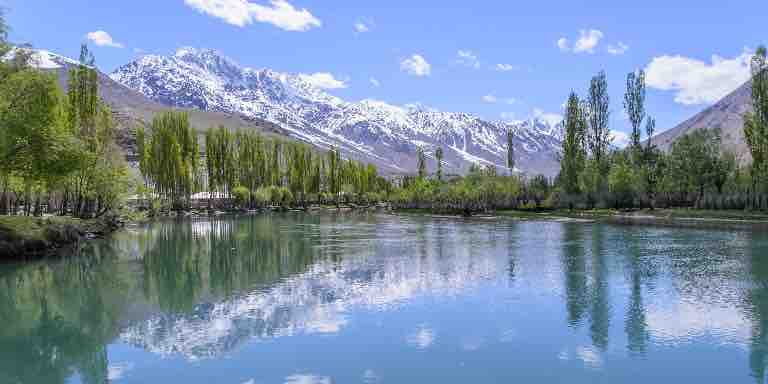}\\[0.5mm]
    \includegraphics[width=\textwidth]{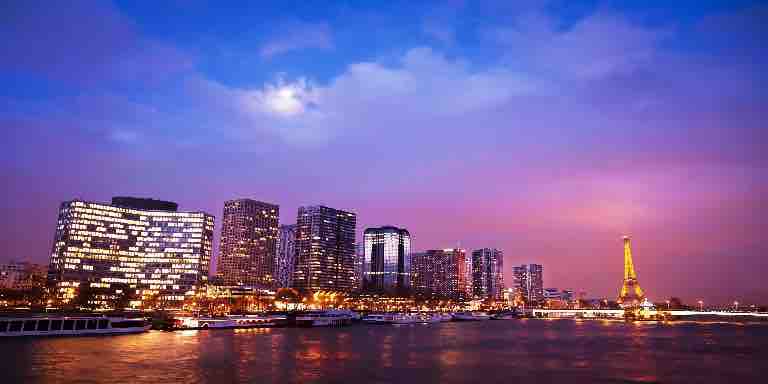}
    \end{minipage}
    }
    \hspace{-1.5mm}
    \subfloat[Style]{
    \begin{minipage}[t]{0.208\textwidth}
    \centering
    \includegraphics[width=\textwidth]{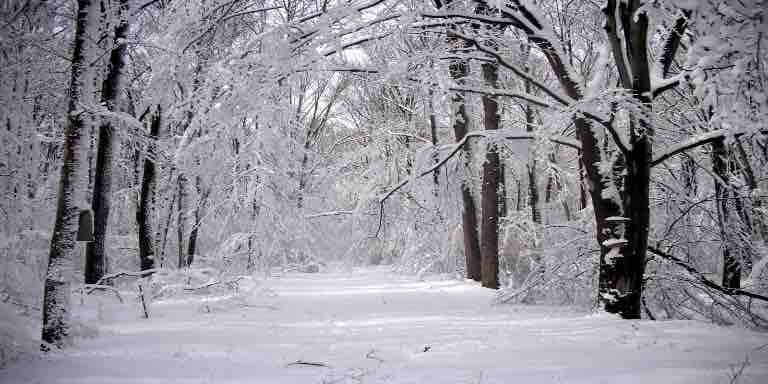}\\[0.5mm]
    \includegraphics[width=\textwidth]{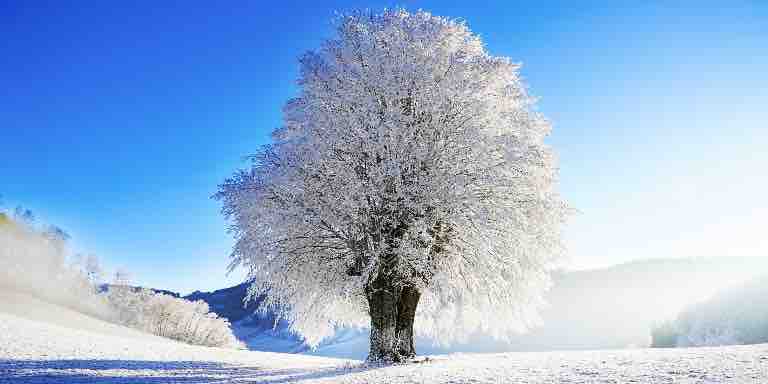}\\[0.5mm]
    \includegraphics[width=\textwidth]{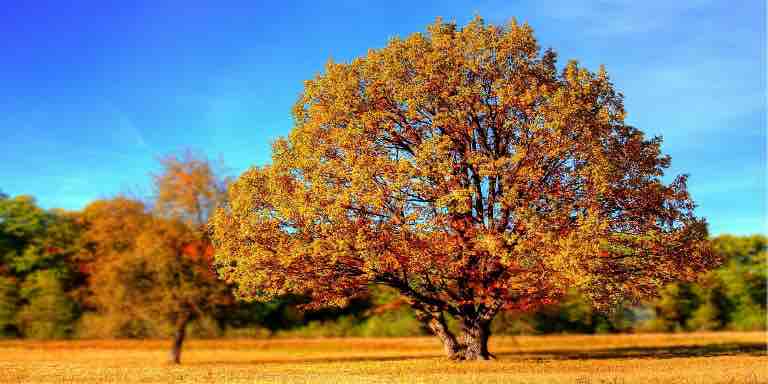}\\[0.5mm]
    \includegraphics[width=\textwidth]{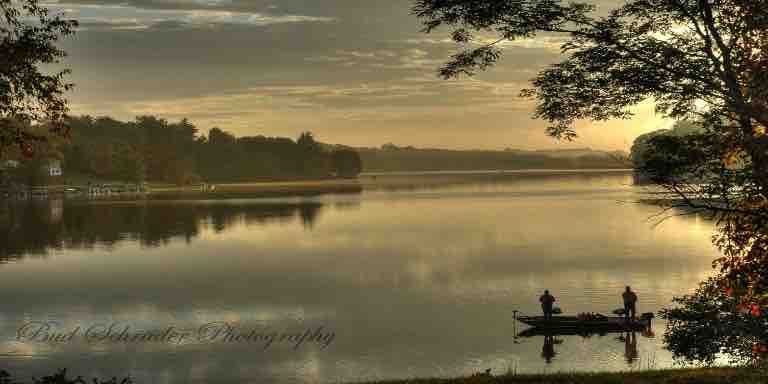}\\[0.5mm]
    \includegraphics[width=\textwidth]{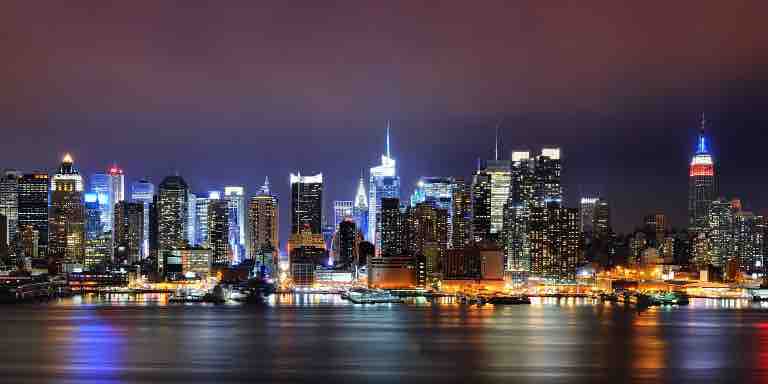}
    \end{minipage}
    }
    \hspace{-1.5mm}
    \subfloat[PhotoWCT~\cite{li2018closed}]{
    \begin{minipage}[t]{0.208\textwidth}
    \centering
    \includegraphics[width=\textwidth]{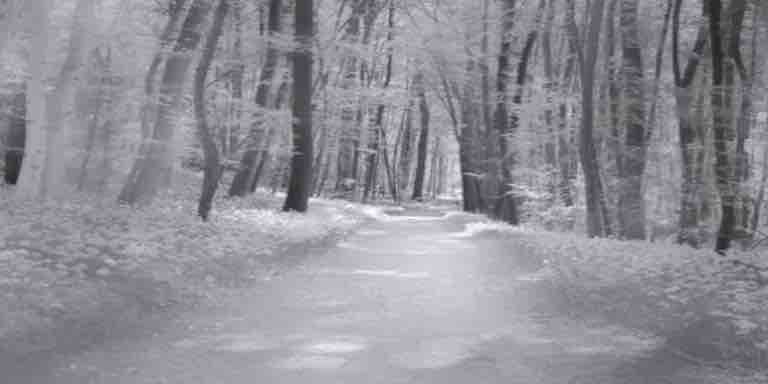}\\[0.5mm]
    \includegraphics[width=\textwidth]{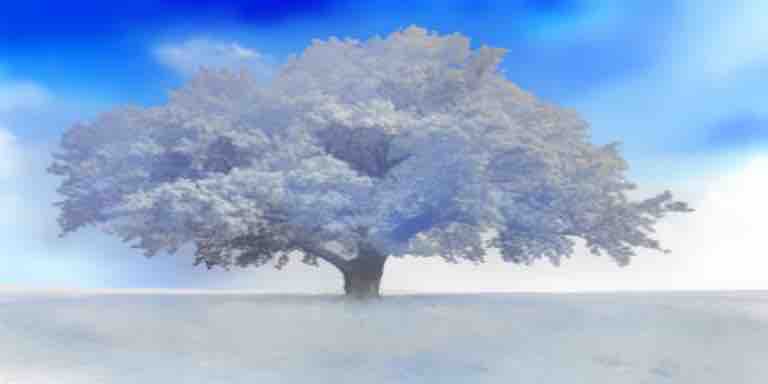}\\[0.5mm]
    \includegraphics[width=\textwidth]{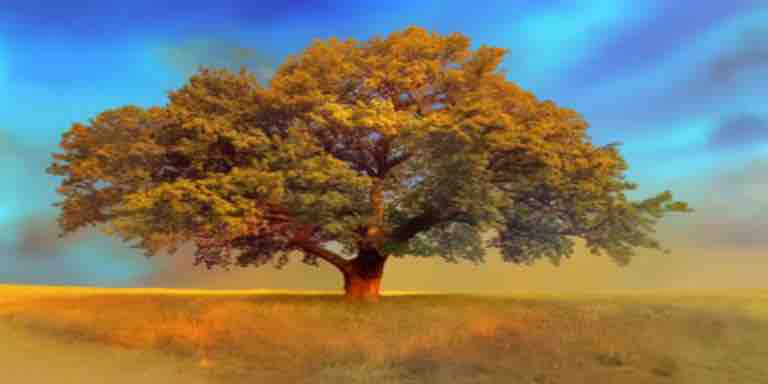}\\[0.5mm]
    \includegraphics[width=\textwidth]{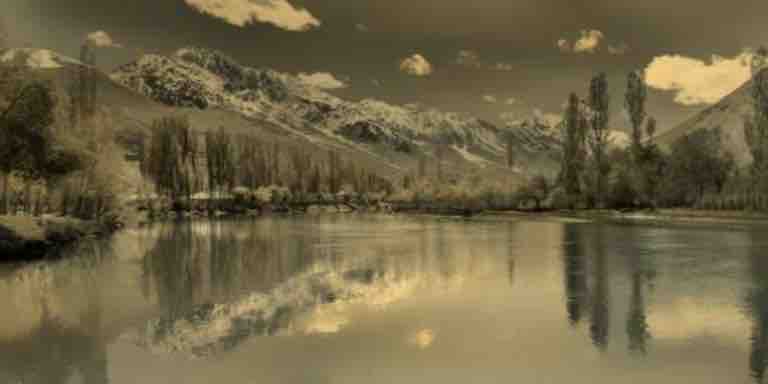}\\[0.5mm]
    \includegraphics[width=\textwidth]{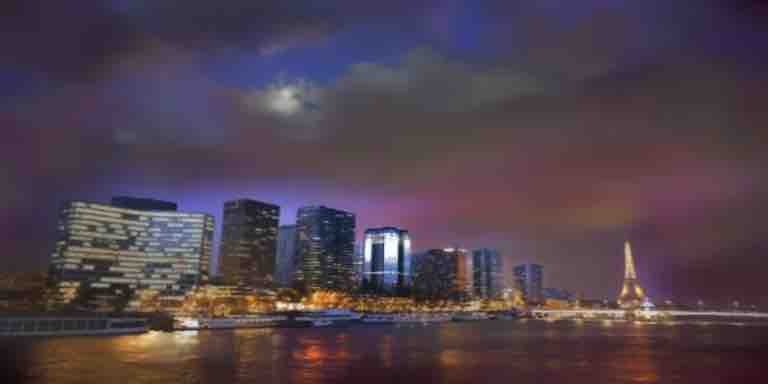}
    \end{minipage}
    }
    \hspace{-1.5mm}
    \subfloat[PhotoNet(WCT)]{
    \begin{minipage}[t]{0.208\textwidth}
    \centering
    \includegraphics[width=\textwidth]{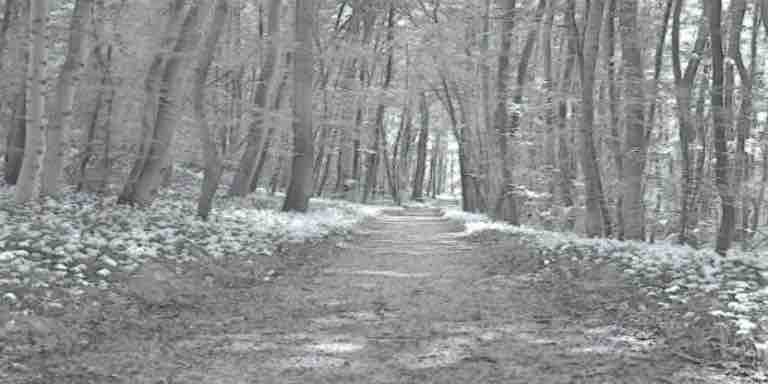}\\[0.5mm]
    \includegraphics[width=\textwidth]{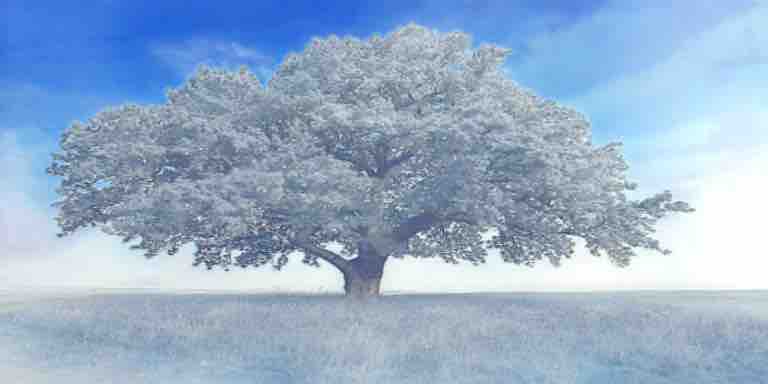}\\[0.5mm]
    \includegraphics[width=\textwidth]{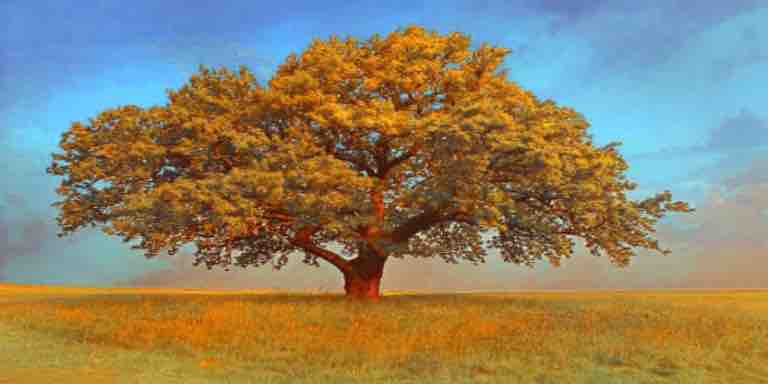}\\[0.5mm]
    \includegraphics[width=\textwidth]{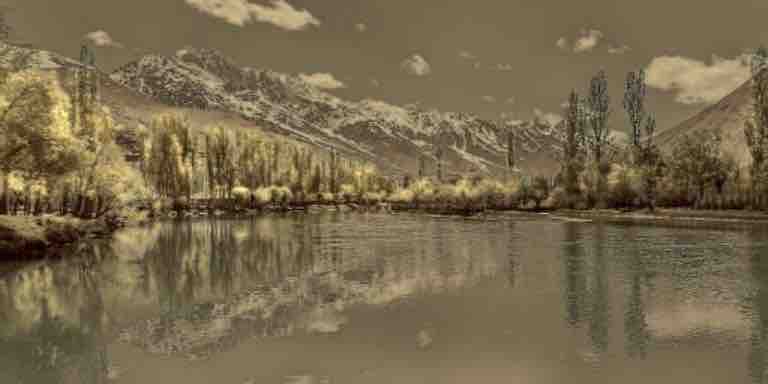}\\[0.5mm]
    \includegraphics[width=\textwidth]{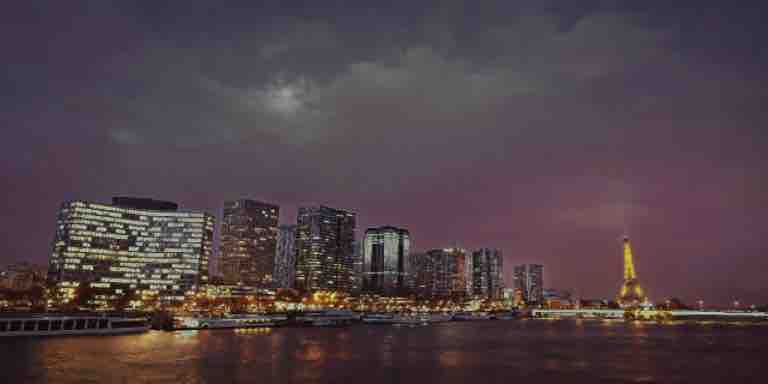}
    \end{minipage}
    }
    \caption{\textbf{Photorealistic style transfer comparison between the PhotoNet(WCT) and PhotoWCT~\cite{li2018closed}.}}
    \label{fig:photo_1}
\end{figure*}
\begin{figure*}[h]
    \centering
    \subfloat[Content]{
    \begin{minipage}[t]{0.208\textwidth}
    \centering
    \includegraphics[width=\textwidth]{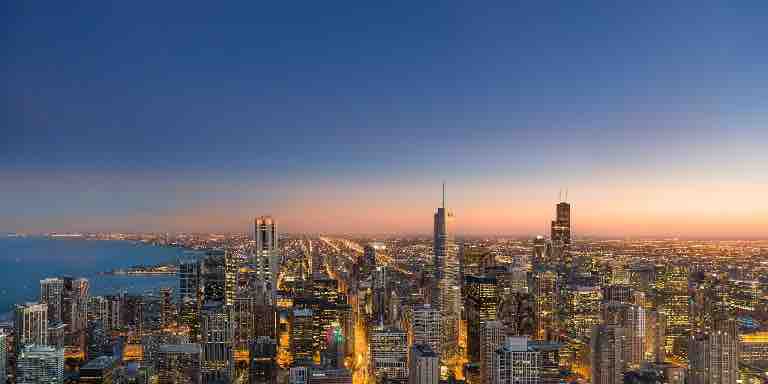}\\[0.5mm]
    \includegraphics[width=\textwidth]{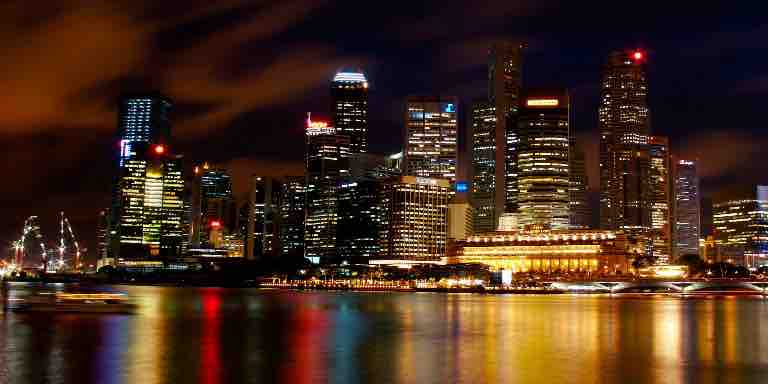}\\[0.5mm]
    \includegraphics[width=\textwidth]{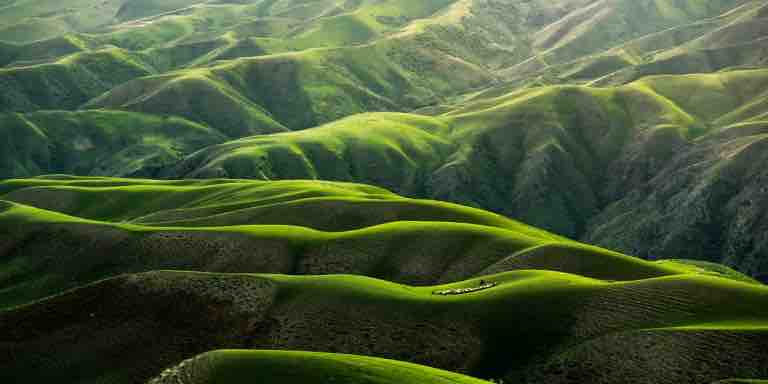}\\[0.5mm]
    \includegraphics[width=\textwidth]{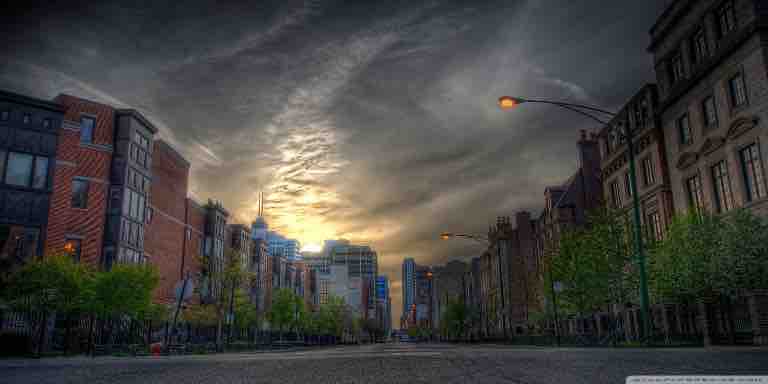}\\[0.5mm]
    \includegraphics[width=\textwidth]{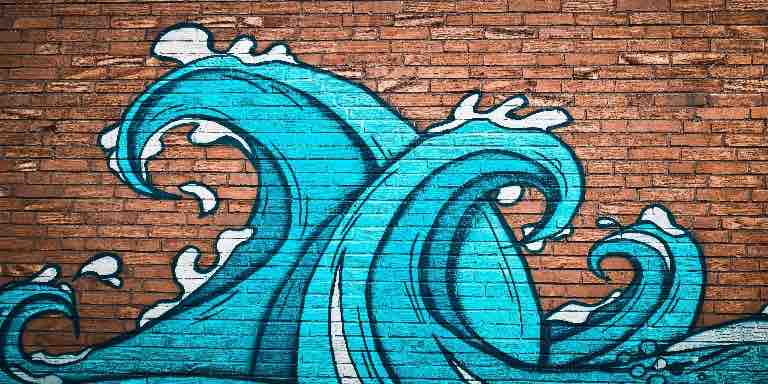}
    \end{minipage}
    }
    \hspace{-1.5mm}
    \subfloat[Style]{
    \begin{minipage}[t]{0.208\textwidth}
    \centering
    \includegraphics[width=\textwidth]{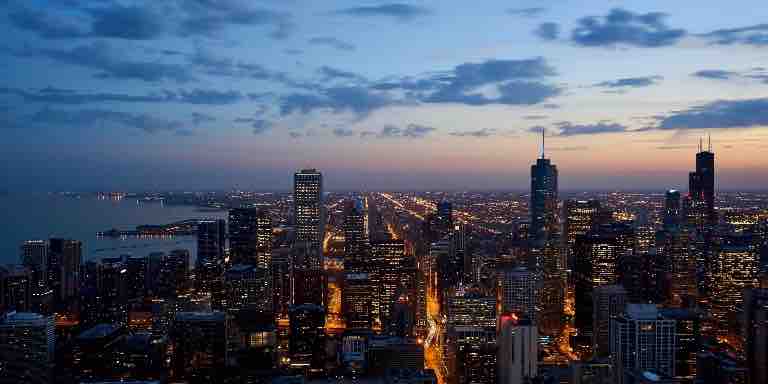}\\[0.5mm]
    \includegraphics[width=\textwidth]{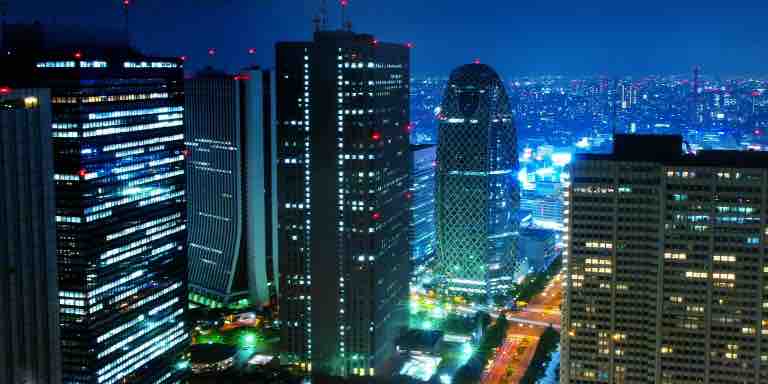}\\[0.5mm]
    \includegraphics[width=\textwidth]{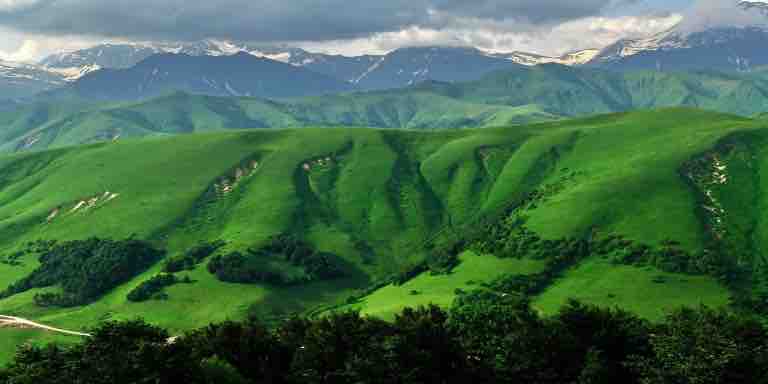}\\[0.5mm]
    \includegraphics[width=\textwidth]{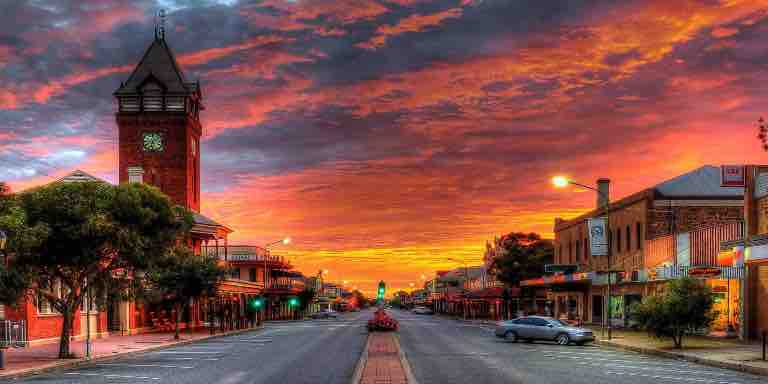}\\[0.5mm]
    \includegraphics[width=\textwidth]{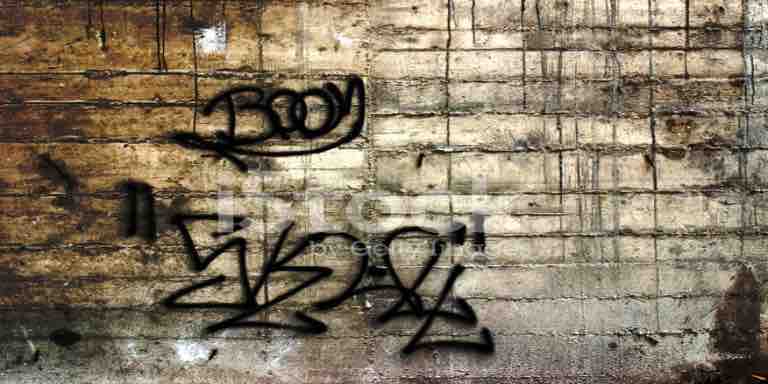}
    \end{minipage}
    }
    \hspace{-1.5mm}
    \subfloat[PhotoWCT~\cite{li2018closed}]{
    \begin{minipage}[t]{0.208\textwidth}
    \centering
    \includegraphics[width=\textwidth]{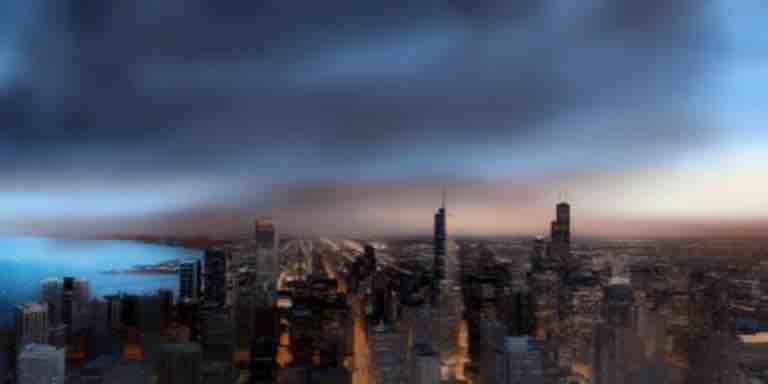}\\[0.5mm]
    \includegraphics[width=\textwidth]{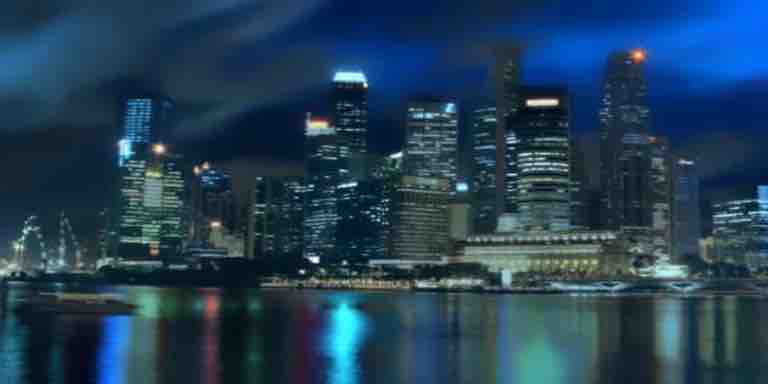}\\[0.5mm]
    \includegraphics[width=\textwidth]{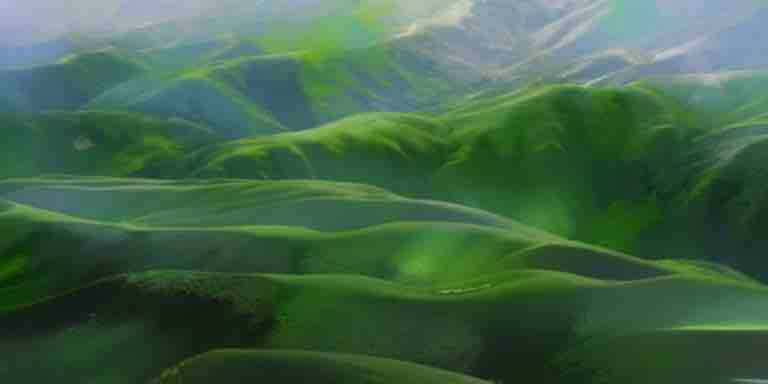}\\[0.5mm]
    \includegraphics[width=\textwidth]{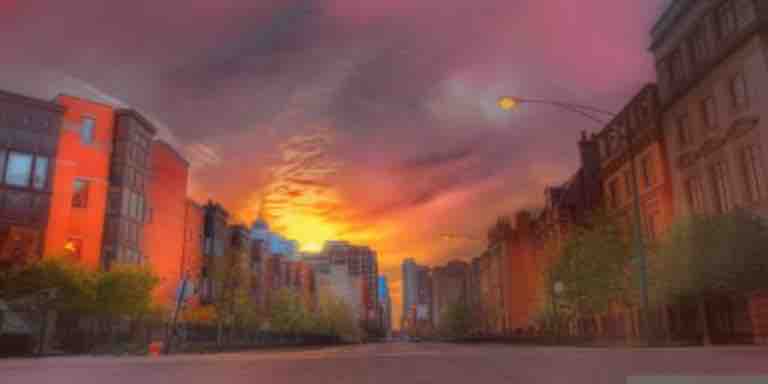}\\[0.5mm]
    \includegraphics[width=\textwidth]{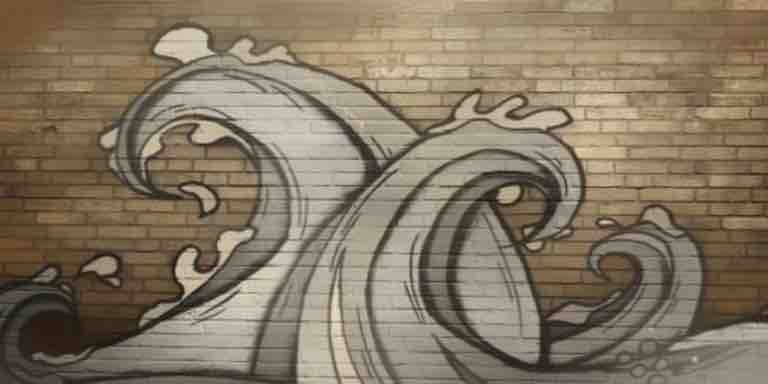}
    \end{minipage}
    }
    \hspace{-1.5mm}
    \subfloat[PhotoNet(WCT)]{
    \begin{minipage}[t]{0.208\textwidth}
    \centering
    \includegraphics[width=\textwidth]{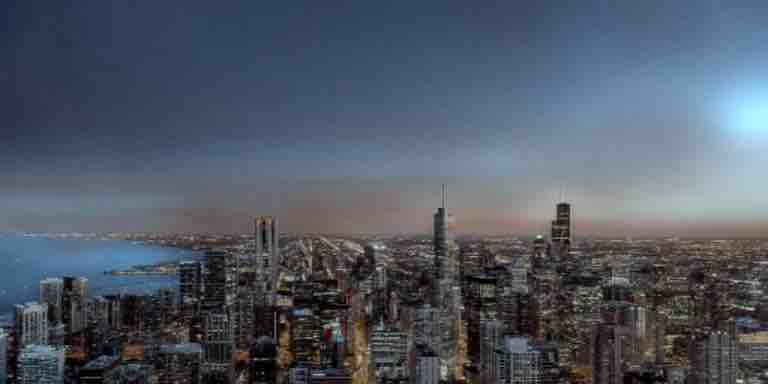}\\[0.5mm]
    \includegraphics[width=\textwidth]{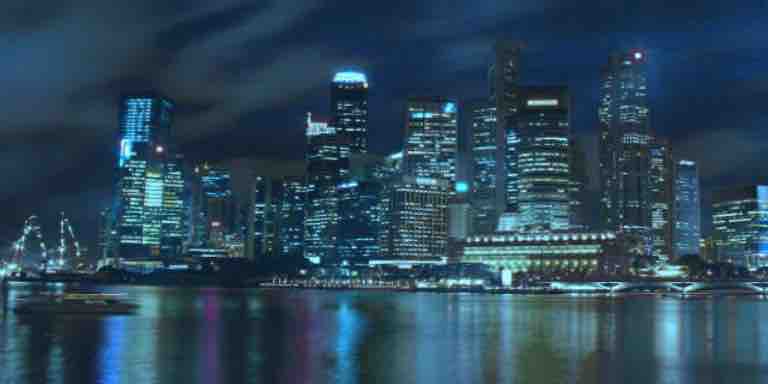}\\[0.5mm]
    \includegraphics[width=\textwidth]{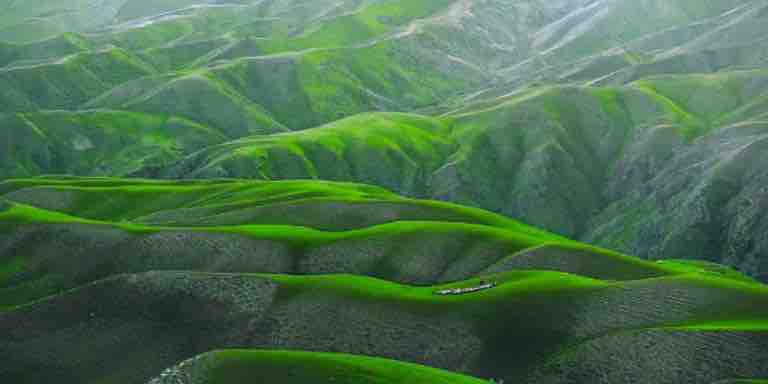}\\[0.5mm]
    \includegraphics[width=\textwidth]{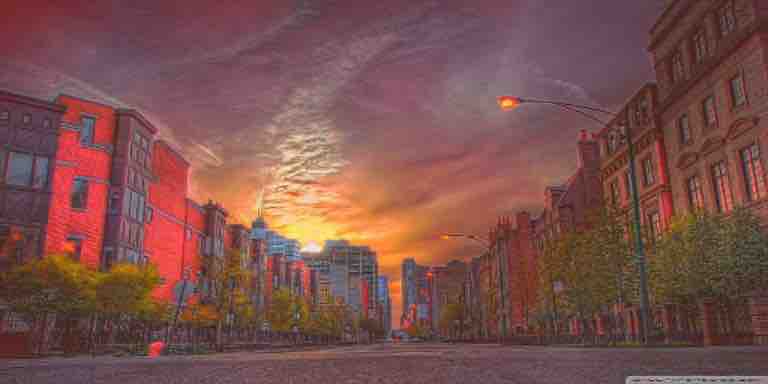}\\[0.5mm]
    \includegraphics[width=\textwidth]{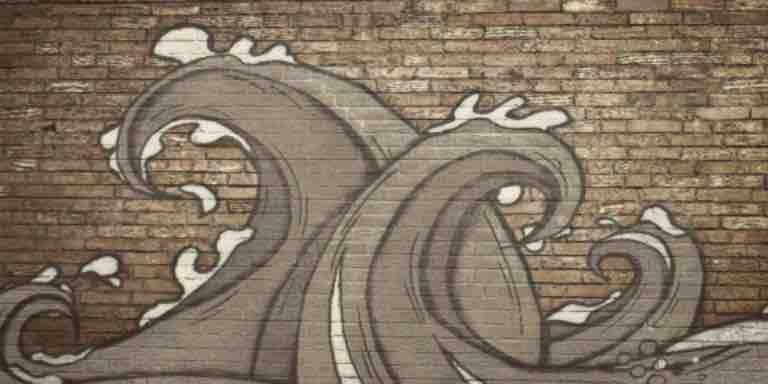}
    \end{minipage}
    }
    \caption{\textbf{Photorealistic style transfer comparison between the PhotoNet(WCT) and PhotoWCT~\cite{li2018closed}.}}
    \label{fig:photo_2}
\end{figure*}
\begin{figure*}[h]
    \centering
    \subfloat[Content]{
    \begin{minipage}[t]{0.208\textwidth}
    \centering
    \includegraphics[width=\textwidth]{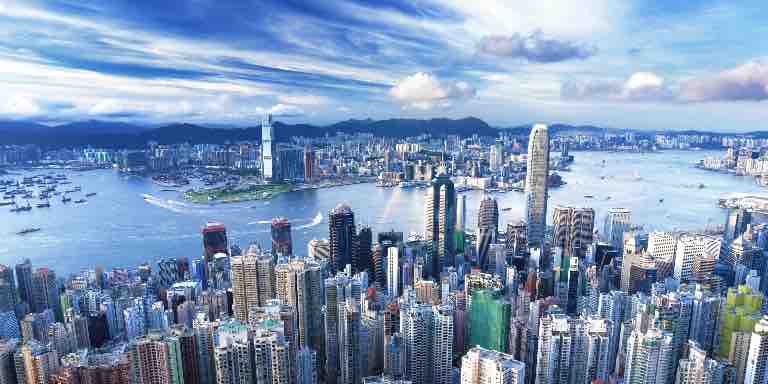}\\[0.5mm]
    \includegraphics[width=\textwidth]{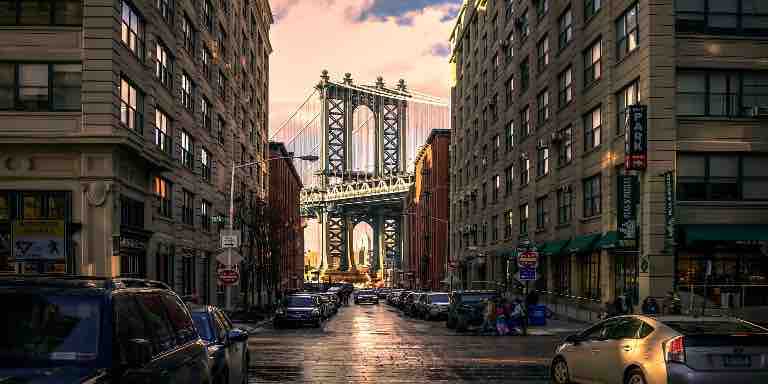}\\[0.5mm]
    \includegraphics[width=\textwidth]{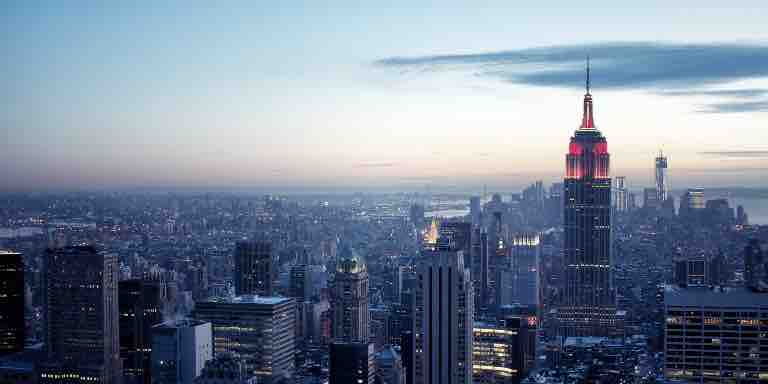}\\[0.5mm]
    \includegraphics[width=\textwidth]{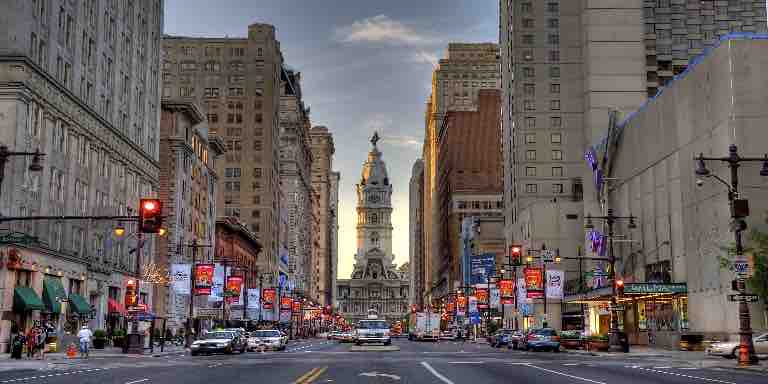}\\[0.5mm]
    \includegraphics[width=\textwidth]{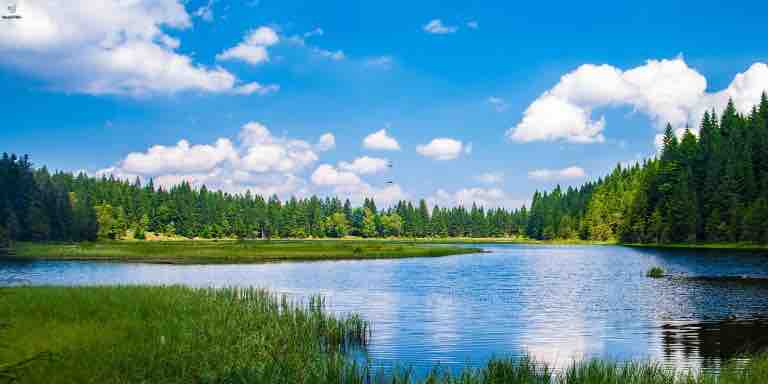}
    \end{minipage}
    }
    \hspace{-1.5mm}
    \subfloat[Style]{
    \begin{minipage}[t]{0.208\textwidth}
    \centering
    \includegraphics[width=\textwidth]{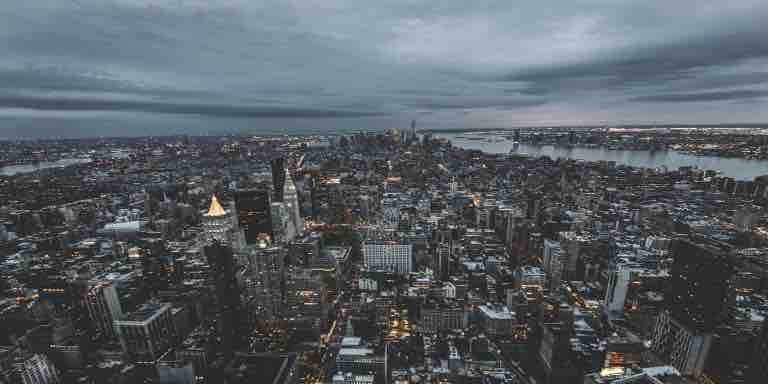}\\[0.5mm]
    \includegraphics[width=\textwidth]{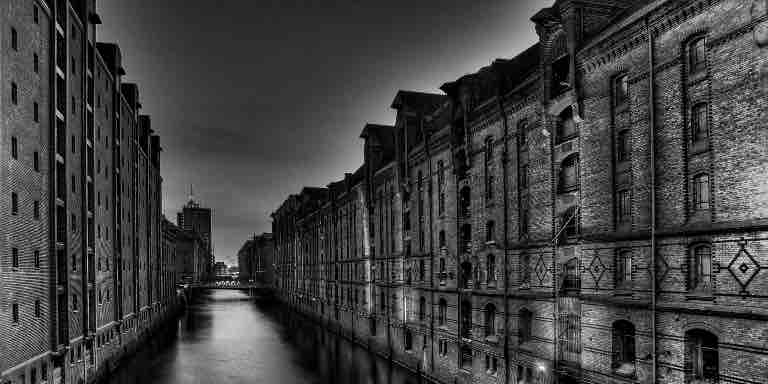}\\[0.5mm]
    \includegraphics[width=\textwidth]{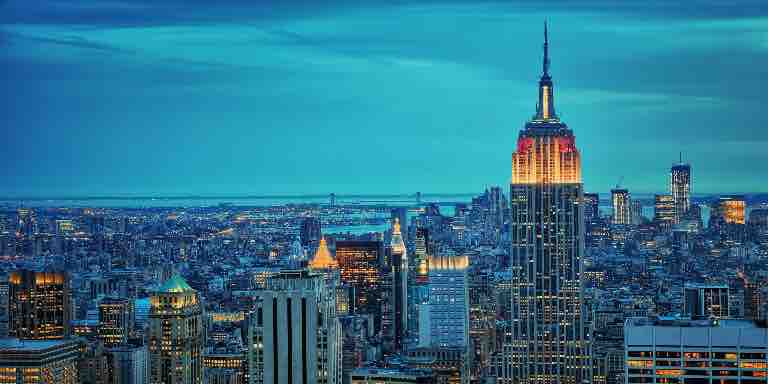}\\[0.5mm]
    \includegraphics[width=\textwidth]{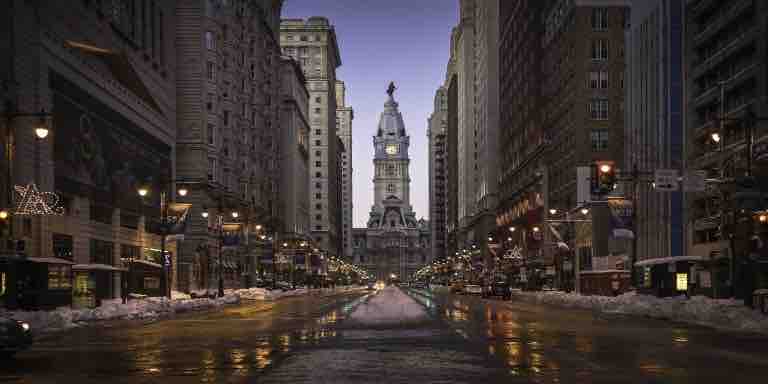}\\[0.5mm]
    \includegraphics[width=\textwidth]{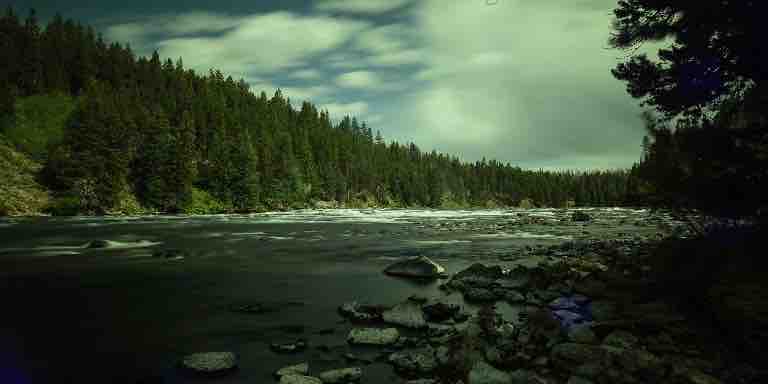}
    \end{minipage}
    }
    \hspace{-1.5mm}
    \subfloat[PhotoWCT~\cite{li2018closed}]{
    \begin{minipage}[t]{0.208\textwidth}
    \centering
    \includegraphics[width=\textwidth]{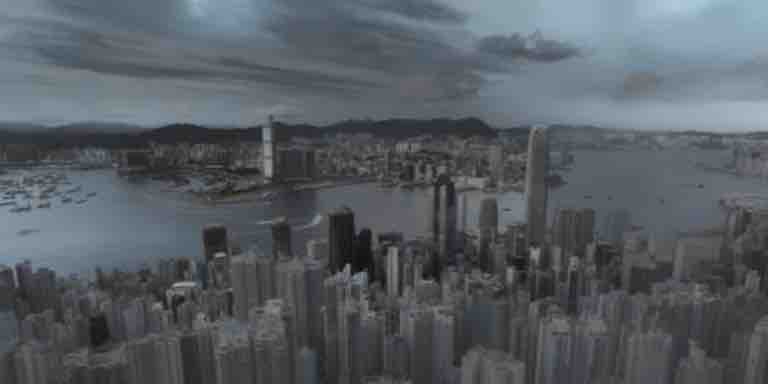}\\[0.5mm]
    \includegraphics[width=\textwidth]{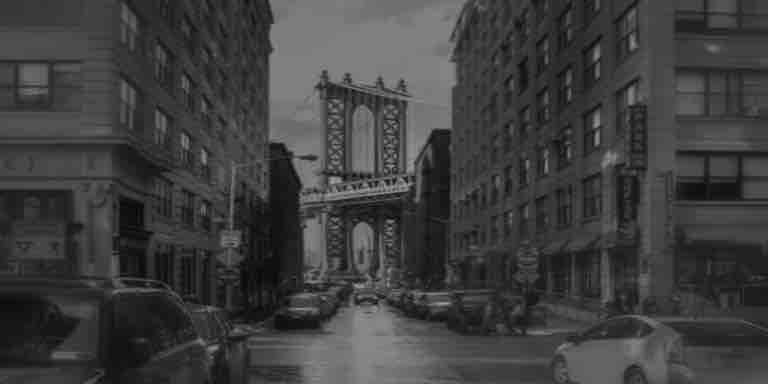}\\[0.5mm]
    \includegraphics[width=\textwidth]{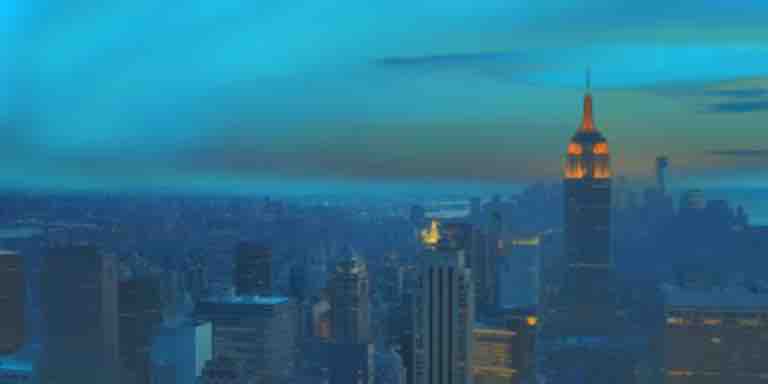}\\[0.5mm]
    \includegraphics[width=\textwidth]{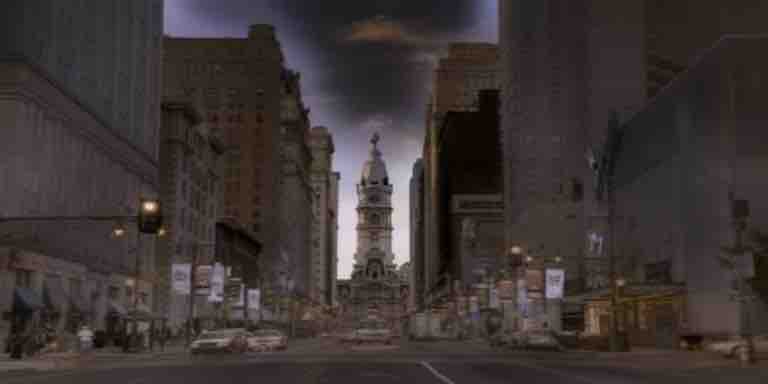}\\[0.5mm]
    \includegraphics[width=\textwidth]{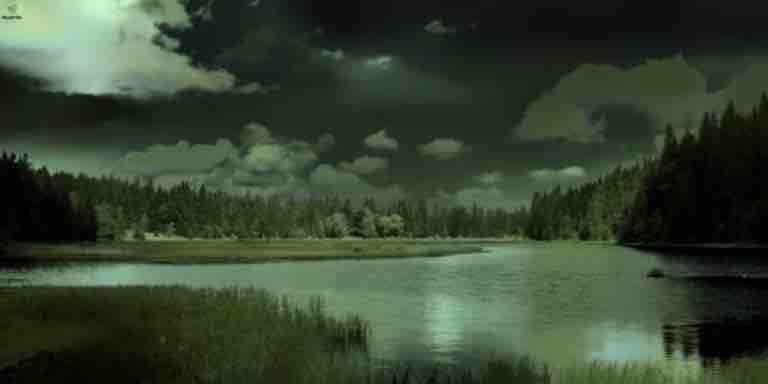}
    \end{minipage}
    }
    \hspace{-1.5mm}
    \subfloat[PhotoNet(WCT)]{
    \begin{minipage}[t]{0.208\textwidth}
    \centering
    \includegraphics[width=\textwidth]{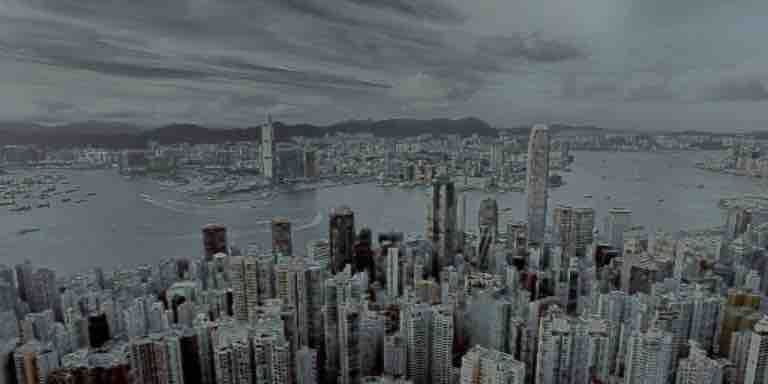}\\[0.5mm]
    \includegraphics[width=\textwidth]{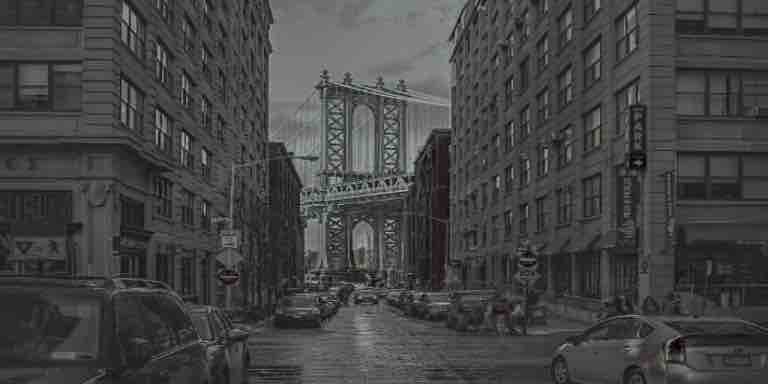}\\[0.5mm]
    \includegraphics[width=\textwidth]{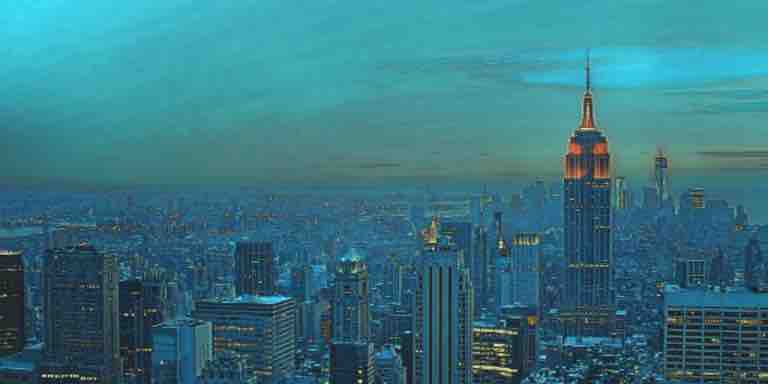}\\[0.5mm]
    \includegraphics[width=\textwidth]{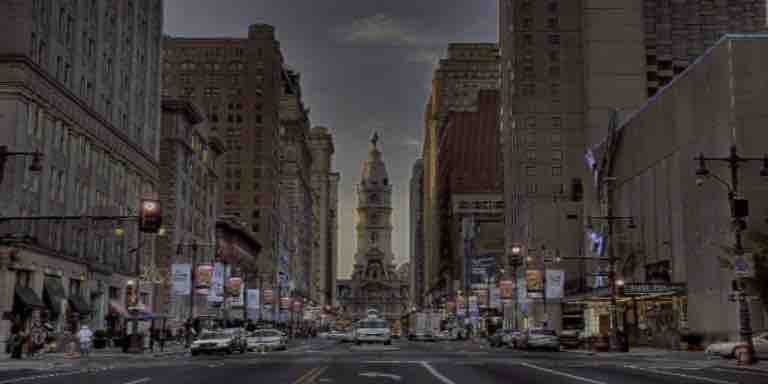}\\[0.5mm]
    \includegraphics[width=\textwidth]{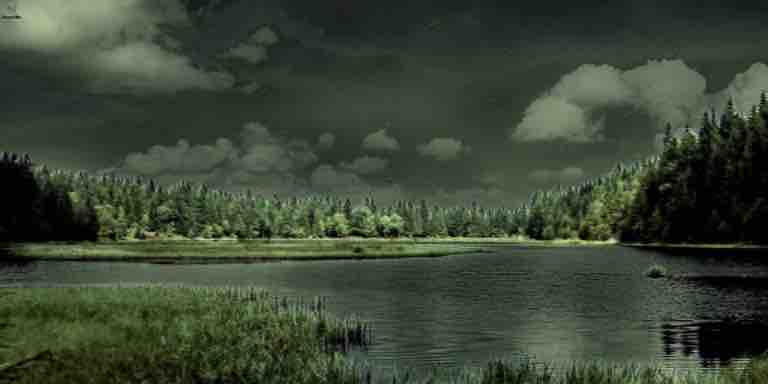}
    \end{minipage}
    }
    \caption{\textbf{Photorealistic style transfer comparison between the PhotoNet(WCT) and PhotoWCT~\cite{li2018closed}.}}
    \label{fig:photo_3}
\end{figure*}


\begin{thebibliography}{10}\itemsep=-1pt

\bibitem{chen2017stylebank}
D.~Chen, L.~Yuan, J.~Liao, N.~Yu, and G.~Hua.
\newblock Stylebank: an explicit representation for neural image style
  transfer.
\newblock In {\em CVPR}, 2017.

\bibitem{chen2016fast}
T.~Q. Chen and M.~Schmidt.
\newblock Fast patch-based style transfer of arbitrary style.
\newblock {\em arXiv preprint arXiv:1612.04337}, 2016.

\bibitem{deng2009imagenet}
J.~Deng, W.~Dong, R.~Socher, L.-J. Li, K.~Li, and L.~Fei-Fei.
\newblock Imagenet: a large-scale hierarchical image database.
\newblock In {\em CVPR}, 2009.

\bibitem{dumoulin2017learned}
V.~Dumoulin, J.~Shlens, and M.~Kudlur.
\newblock A learned representation for artistic style.
\newblock In {\em ICLR}, 2017.

\bibitem{frigo2016split}
O.~Frigo, N.~Sabater, J.~Delon, and P.~Hellier.
\newblock Split and match: example-based adaptive patch sampling for
  unsupervised style transfer.
\newblock In {\em CVPR}, 2016.

\bibitem{gatys2015neural}
L.~A. Gatys, A.~S. Ecker, and M.~Bethge.
\newblock A neural algorithm of artistic style.
\newblock {\em arXiv preprint arXiv:1508.06576}, 2015.

\bibitem{Gatys2016}
L.~A. Gatys, A.~S. Ecker, and M.~Bethge.
\newblock Image style transfer using convolutional neural networks.
\newblock In {\em CVPR}, 2016.

\bibitem{gu2018arbitrary}
S.~Gu, C.~Chen, J.~Liao, and L.~Yuan.
\newblock Arbitrary style transfer with deep feature reshuffle.
\newblock In {\em CVPR}, 2018.

\bibitem{hertzmann1998painterly}
A.~Hertzmann.
\newblock Painterly rendering with curved brush strokes of multiple sizes.
\newblock In {\em SIGGRAPH}, 1998.

\bibitem{hertzmann2001image}
A.~Hertzmann, C.~E. Jacobs, N.~Oliver, B.~Curless, and D.~H. Salesin.
\newblock Image analogies.
\newblock In {\em SIGGRAPH}, 2001.

\bibitem{heusel2017gans}
M.~Heusel, H.~Ramsauer, T.~Unterthiner, B.~Nessler, and S.~Hochreiter.
\newblock Gans trained by a two time-scale update rule converge to a local nash
  equilibrium.
\newblock In {\em NIPS}, 2017.

\bibitem{huang2017arbitrary}
X.~Huang and S.~J. Belongie.
\newblock Arbitrary style transfer in real-time with adaptive instance
  normalization.
\newblock In {\em ICCV}, 2017.

\bibitem{huang2018multimodal}
X.~Huang, M.-Y. Liu, S.~Belongie, and J.~Kautz.
\newblock Multimodal unsupervised image-to-image translation.
\newblock In {\em ECCV}, 2018.

\bibitem{isola2017image}
P.~Isola, J.-Y. Zhu, T.~Zhou, and A.~A. Efros.
\newblock Image-to-image translation with conditional adversarial networks.
\newblock In {\em CVPR}, 2017.

\bibitem{johnson2016perceptual}
J.~Johnson, A.~Alahi, and L.~Fei-Fei.
\newblock Perceptual losses for real-time style transfer and super-resolution.
\newblock In {\em ECCV}, 2016.

\bibitem{kingma2014adam}
D.~P. Kingma and J.~Ba.
\newblock Adam: a method for stochastic optimization.
\newblock {\em arXiv preprint arXiv:1412.6980}, 2014.

\bibitem{li2016combining}
C.~Li and M.~Wand.
\newblock Combining markov random fields and convolutional neural networks for
  image synthesis.
\newblock In {\em CVPR}, 2016.

\bibitem{li2017laplacian}
S.~Li, X.~Xu, L.~Nie, and T.-S. Chua.
\newblock Laplacian-steered neural style transfer.
\newblock In {\em ACM MM}, 2017.

\bibitem{li2017diversified}
Y.~Li, C.~Fang, J.~Yang, Z.~Wang, X.~Lu, and M.-H. Yang.
\newblock Diversified texture synthesis with feed-forward networks.
\newblock In {\em CVPR}, 2017.

\bibitem{li2017universal}
Y.~Li, C.~Fang, J.~Yang, Z.~Wang, X.~Lu, and M.-H. Yang.
\newblock Universal style transfer via feature transforms.
\newblock In {\em NIPS}, 2017.

\bibitem{li2018closed}
Y.~Li, M.-Y. Liu, X.~Li, M.-H. Yang, and J.~Kautz.
\newblock A closed-form solution to photorealistic image stylization.
\newblock In {\em ECCV}, 2018.

\bibitem{liao2017visual}
J.~Liao, Y.~Yao, L.~Yuan, G.~Hua, and S.~B. Kang.
\newblock Visual attribute transfer through deep image analogy.
\newblock {\em arXiv preprint arXiv:1705.01088}, 2017.

\bibitem{liu2017unsupervised}
M.-Y. Liu, T.~Breuel, and J.~Kautz.
\newblock Unsupervised image-to-image translation networks.
\newblock In {\em NIPS}, 2017.

\bibitem{liu2016coupled}
M.-Y. Liu and O.~Tuzel.
\newblock Coupled generative adversarial networks.
\newblock In {\em NIPS}, 2016.

\bibitem{luan2017deep}
F.~Luan, S.~Paris, E.~Shechtman, and K.~Bala.
\newblock Deep photo style transfer.
\newblock In {\em CVPR}, 2017.

\bibitem{pitie2005n}
F.~Pitie, A.~C. Kokaram, and R.~Dahyot.
\newblock N-dimensional probability density function transfer and its
  application to color transfer.
\newblock In {\em ICCV}, 2005.

\bibitem{risser2017stable}
E.~Risser, P.~Wilmot, and C.~Barnes.
\newblock Stable and controllable neural texture synthesis and style transfer
  using histogram losses.
\newblock {\em arXiv preprint arXiv:1701.08893}, 2017.

\bibitem{ronneberger2015u}
O.~Ronneberger, P.~Fischer, and T.~Brox.
\newblock U-net: convolutional networks for biomedical image segmentation.
\newblock In {\em International Conference on Medical Image Computing and
  Computer-assisted Intervention}, 2015.

\bibitem{rudin1992nonlinear}
L.~I. Rudin, S.~Osher, and E.~Fatemi.
\newblock Nonlinear total variation based noise removal algorithms.
\newblock {\em Physica D: nonlinear phenomena}, 60(1-4):259--268, 1992.

\bibitem{sheng2018avatar}
L.~Sheng, Z.~Lin, J.~Shao, and X.~Wang.
\newblock Avatar-net: multi-scale zero-shot style transfer by feature
  decoration.
\newblock In {\em CVPR}, 2018.

\bibitem{shih2014style}
Y.~Shih, S.~Paris, C.~Barnes, W.~T. Freeman, and F.~Durand.
\newblock Style transfer for headshot portraits.
\newblock {\em ACM Transactions on Graphics}, 33(4):148, 2014.

\bibitem{shih2013data}
Y.~Shih, S.~Paris, F.~Durand, and W.~T. Freeman.
\newblock Data-driven hallucination of different times of day from a single
  outdoor photo.
\newblock {\em ACM Transactions on Graphics}, 32(6):200, 2013.

\bibitem{shrivastava2017learning}
A.~Shrivastava, T.~Pfister, O.~Tuzel, J.~Susskind, W.~Wang, and R.~Webb.
\newblock Learning from simulated and unsupervised images through adversarial
  training.
\newblock In {\em CVPR}, 2017.

\bibitem{simonyan2014very}
K.~Simonyan and A.~Zisserman.
\newblock Very deep convolutional networks for large-scale image recognition.
\newblock {\em arXiv preprint arXiv:1409.1556}, 2014.

\bibitem{taigman2016unsupervised}
Y.~Taigman, A.~Polyak, and L.~Wolf.
\newblock Unsupervised cross-domain image generation.
\newblock In {\em ICLR}, 2017.

\bibitem{ulyanov2016texture}
D.~Ulyanov, V.~Lebedev, A.~Vedaldi, and V.~S. Lempitsky.
\newblock Texture networks: feed-forward synthesis of textures and stylized
  images.
\newblock In {\em ICML}, 2016.

\bibitem{ulyanov1607instance}
D.~Ulyanov, A.~Vedaldi, and V.~Lempitsky.
\newblock Instance normalization: the missing ingredient for fast stylization.
\newblock {\em arXiv preprint arXiv:1607.08022}, 2016.

\bibitem{ulyanov2017improved}
D.~Ulyanov, A.~Vedaldi, and V.~S. Lempitsky.
\newblock Improved texture networks: maximizing quality and diversity in
  feed-forward stylization and texture synthesis.
\newblock In {\em CVPR}, 2017.

\bibitem{wang2018high}
T.-C. Wang, M.-Y. Liu, J.-Y. Zhu, A.~Tao, J.~Kautz, and B.~Catanzaro.
\newblock High-resolution image synthesis and semantic manipulation with
  conditional gans.
\newblock In {\em CVPR}, 2018.

\bibitem{wang2017multimodal}
X.~Wang, G.~Oxholm, D.~Zhang, and Y.-F. Wang.
\newblock Multimodal transfer: a hierarchical deep convolutional neural network
  for fast artistic style transfer.
\newblock In {\em CVPR}, 2017.

\bibitem{winnemoller2006real}
H.~Winnem\"{o}ller, S.~C. Olsen, and B.~Gooch.
\newblock Real-time video abstraction.
\newblock {\em ACM Transactions on Graphics}, 25(3):1221--1226, 2006.

\bibitem{yu2018deep}
F.~Yu, D.~Wang, E.~Shelhamer, and T.~Darrell.
\newblock Deep layer aggregation.
\newblock In {\em CVPR}, 2018.

\bibitem{zhao2017pspnet}
H.~Zhao, J.~Shi, X.~Qi, X.~Wang, and J.~Jia.
\newblock Pyramid scene parsing network.
\newblock In {\em CVPR}, 2017.

\bibitem{zhu2017unpaired}
J.-Y. Zhu, T.~Park, P.~Isola, and A.~A. Efros.
\newblock Unpaired image-to-image translation using cycle-consistent
  adversarial networks.
\newblock In {\em ICCV}, 2017.

\end{thebibliography}
\end{document}